\documentclass[10pt,journal]{IEEEtran}
\usepackage{fontenc}
\usepackage{graphicx}
\usepackage[cmex10]{amsmath}
\usepackage{array}
\usepackage{amssymb}
\usepackage{amsthm}
\usepackage{xcolor}
\usepackage[FIGBOTCAP]{subfigure}
\usepackage{parskip}
\usepackage{mathrsfs}
\usepackage{bm}
\usepackage[noadjust]{cite}
\usepackage{amsmath,graphicx}
\usepackage[bookmarks=false]{hyperref}
\usepackage[font=small,labelfont=bf]{caption}
\hypersetup{
    colorlinks=true,
    linkcolor=blue,
    filecolor=magenta,
    urlcolor=blue,
}
\usepackage{bm}
\usepackage{epstopdf}
\usepackage{multirow}
\usepackage{algorithm}
\usepackage{algpseudocode}
\usepackage{amssymb}
\usepackage{amsmath}
\usepackage{fancyhdr}
\usepackage{graphicx}
\usepackage{float}
\usepackage{subfigure}
\usepackage{comment}

\providecommand{\ve}[1]{\boldsymbol{\mathrm{#1}}}

\theoremstyle{definition}
\newtheorem{definition}{Definition}[section]
\newtheorem{theorem}{Theorem}[section]
\newtheorem{corollary}{Corollary}[theorem]

\begin{document}
\title{Concept Drift Detection and Adaptation with Hierarchical Hypothesis Testing}
\author{Shujian~Yu, Zubin~Abraham, Heng~Wang, \\ Mohak~Shah, Yantao~Wei, and Jos\'{e} C. Pr\'{i}ncipe%
\thanks{Shujian Yu and Jos\'{e} C. Pr\'{i}ncipe are with the Department of Electrical and Computer Engineering, University of Florida, Gainesville, FL 32611,
  USA (email: yusjlcy9011@ufl.edu; principe@cnel.ufl.edu).}%
\thanks{Zubin~Abraham is with the Robert Bosch Research and Technology Center, Sunnyvale, CA 94085,
  USA (email: zubin.abraham@us.bosch.com).}%
\thanks{Mohak~Shah is with the Robert Bosch Research and Technology Center, Sunnyvale, CA 94085, and the University of Illinois at Chicago, Chicago, IL
60637, USA (email: Mohak.Shah@us.bosch.com).}%
\thanks{Heng~Wang is with the MZ Inc. (formerly Machine Zone), Research, Palo Alto, CA 94304,
  USA (email:hengwang@mz.com).}
\thanks{Yantao Wei is with the School of Educational Information Technology, Central China Normal University, Wuhan 430079, China (e-mail: yantaowei@mail.ccnu.edu.cn).}}%

\IEEEtitleabstractindextext{%
	\begin{abstract}
		A fundamental issue for statistical classification models in a streaming environment is that the joint distribution between predictor and response variables changes over time (a phenomenon also known as concept drifts), such that their classification performance deteriorates dramatically. In this paper, we first present a hierarchical hypothesis testing (HHT) framework that can detect and also adapt to various concept drift types (e.g., recurrent or irregular, gradual or abrupt), even in the presence of imbalanced data labels. A novel concept drift detector, namely Hierarchical Linear Four Rates (HLFR), is implemented under the HHT framework thereafter. By substituting a widely-acknowledged retraining scheme with an adaptive training strategy, we further demonstrate that the concept drift adaptation capability of HLFR can be significantly boosted. The theoretical analysis on the Type-I and Type-II errors of HLFR is also performed. Experiments on both simulated and real-world datasets illustrate that our methods outperform state-of-the-art methods in terms of detection precision, detection delay as well as the adaptability across different concept drift types.		
	\end{abstract}
	
	}

\maketitle
\IEEEdisplaynontitleabstractindextext
\IEEEpeerreviewmaketitle
%\long\def\/*#1*/{}
\graphicspath{{figures/}}

% for peerreview papers, inserts a page break and creates the second title.
% Will be ignored for other modes.
%\IEEEpeerreviewmaketitle

\section{Introduction}\label{sec:introduction}
With the exponential growth of data, it becomes increasingly challenging to design and implement effective techniques for analyzing and detecting changes in a streaming environment~\cite{slavakis2014stochastic,hu2014toward}. As a result, early approaches for detecting statistical changes in a time series (such as change point detectors), have had to be extended for online detection of changes in multivariate data streams \cite{basseville1993detection}. Some of these techniques for detecting intrinsic changes in the relationship of the incoming data have been successfully applied to various real-world applications, such as email filtering, network traffic analysis and user preference prediction~\cite{gama2014survey,wang2017systematic}.

Online classification is another common task performed on multivariate streaming data that takes advantage of these statistical relationships to predict a class label at each time index \cite{ross2012exponentially}. If the underlying source (or joint data distribution) that generates the data is not stationary, the optimal decision rule for the classifier would change over time - a phenomena known as concept drift \cite{widmer1996learning}. Given the impact of concept drift on the predictive performance of an online classifier, there is often a need to detect these concept drifts as early as possible. The inability of change point detectors to detect these concept drifts, has motivated the need for concept drift detectors that not only monitor the join distribution of a multivariate data stream but also changes in its relationship to the class labels of the streaming data.

There are two different approaches to address concept drifts in streaming data~\cite{ross2012exponentially}. The first, automatically adapts the parameters of a statistical model in an incremental fashion \cite{klinkenberg2004learning,bifet2007learning,du2014detecting} or employs an ensemble of classifiers, trained on different windows over the stream, to give the optimal decision \cite{street2001streaming,katakis2010tracking,katakis2008ensemble,elwell2011incremental}. There is no explicit detection of drifts in these methods, but retraining of new classifiers. The second approach integrates a statistical model and a concept drift detector, whose purpose is to signal the need for updating the statistical model once a concept drift is detected. Existing methods in this category monitor the error rate or an error-driven statistics and make a decision based on the statistical learning theory \cite{gama2004learning,wang2013concept,wang2015concept,antwi2012perfsim}. Unlike the first approach that only mitigates deteriorating classification performance over time, the second approach enables identification of the time instant related to concept drift occurrences. The promptness of the alert, i.e. the time that mediates the start of drift until its detection, is crucially important in applications like malware detection or network monitoring \cite{gama2014survey,wang2017systematic}. %Identifying when a concept drift has occurred is important in instances such as credit card fraud detection application, where it may be necessary to gather data pertaining to a concept drift triggered by a particular fraudsters \cite{pavlidis2012adaptive}.% This paper focuses on approaches for detecting concept drifts . %However, we also present a potential .... to intelligently adapt model parameters.

In this paper, we use the second approach and present a novel hierarchical hypothesis testing (HHT) framework for concept drift detection and adaptation. This framework is inspired by the hierarchical architecture that was recently proposed for change point detection \cite{alippi2011hierarchical,alippi2016hierarchical} (see section \ref{section1_1} for more discussion on the difference between concept drift detection and change point detection). The presented work intends to bring new perspectives to the field of concept drift detection and adaptation with the recent advances in hierarchical mechanism (e.g., \cite{alippi2013just,alippi2016hierarchical}) and provides the following contributions. First, we present Hierarchical Linear Four Rates (HLFR) detector~\cite{yu2017concept}, a novel HHT-based concept drift detection method, which is applicable to different types of concept drifts (e.g., recurrent or irregular, gradual or abrupt). A detailed analysis on the Type-I and Type-II errors of the proposed HLFR is also performed. Second, we present an adaptive training approach instead of the commonly used retraining strategy, once a drift is confirmed. The motivation is to leverage knowledge from the historical concept (rather than discard this information as in the retraining strategy), to enhance the classification performance in the new concept. We term this improvement adaptive HLFR (A-HLFR). Admittedly, leveraging previous knowledge to boost classification performance is not novel in the streaming classification scenario. However, to the best of our knowledge, previous work either uses the first approach that do not explicitly identify timestamps or the types of drifts (e.g., \cite{minku2010impact,minku2012ddd,sun2018concept}) or relies heavily on previous restored samples (e.g., \cite{alippi2013just}) which contradicts the \textbf{single pass} criterion\footnote{\textbf{Single pass} criterion: a sample from the data stream should be discarded rather than stored in the memory, once it has been processed~\cite{ross2012exponentially,domingos2003general}.} \cite{ross2012exponentially,domingos2003general}. From this perspective, we are among the first to investigate feasible solutions to perform ``knowledge transfer" without losing intrinsic drift detection capability and the utilization of previous samples. Third, we carry out comprehensive experiments to investigate the benefits of HLFR (in detection) and A-HLFR (in detection and adaptation), and validate the advantage of adaptive training strategy.

\begin{comment}
\textcolor{red}{
This paper extends \cite{yu2017concept} and presents three original contributions. The first contribution is that we provide a detailed error analysis on the Type-I and Type-II errors of proposed HLFR, as well as a comparison with a baseline ensemble detecting approach. Second, we developed an enhancement to HLFR, that is capable of adaptive learning with the aid of novel classifier update mechanism. Third, we carry out extensive experiments to investigate the benefits of HLFR (in detection) and A-HLFR (in detection and adaptation), and validate the rationale of adaptive training strategy.}
\end{comment}

The rest of the paper is organized as follows. In Section \ref{section1}, we give the problem formulation of concept drift and also briefly review related work. In section~\ref{section2}, we present HLFR and elaborate on the layer-I and layer-II tests employed. This section also includes the derivation of the detailed values of Type-I and Type-II errors associated with HLFR. Additionally, we present A-HLFR, that not only detects drifts but also adapts the classifier to handle concept drifts. In section \ref{experiments}, experiments are presented and discussed. Finally, we present the conclusion in section \ref{conclusions}.

\section{Previous approaches} \label{section1}

\subsection{Problem Formulation} \label{section1_1}

Given a continuous stream of labeled samples $\{\mathbf{X}_t, y_t\}$, $t=1,2,...$, a classifier $\hat{f}$ can be learned so that $\hat{f}(\bm{X}_t)\mapsto y_t$. Here, $\mathbf{X}_t$ is a $d$-dimensional feature vector in a predefined vector space $\mathcal{X}=\mathbb{R}^d$ and $y_t\in\{0,1\}$\footnote{This paper only considers binary classification.}. At every time instant $t$, we split the samples into sets $S_A$ (containing $n_A$ recent samples) and set $S_B$ (containing $n_B$  examples that appeared prior to those in $S_A$). A concept drift refers to the joint distribution $P_t(\bm{X},y)$ that generates samples in $S_A$ differs from that in $S_B$ \cite{widmer1996learning,gama2014survey,krawczyk2017ensemble}. From a Bayesian perspective, concept drifts can manifest two fundamental forms of changes~\cite{kelly1999impact}: 1) a change in the posterior probability $P_t(y|\bm{X})$; and 2) a change in the marginal probability $P_t(\bm{X})$ or $P_t(y)$. Existing studies tend to prioritize detecting posterior distribution change \cite{wang2017systematic}, also known as real concept drift \cite{widmer1993effective}, because it clearly indicates the optimal decision rule.

\begin{figure}[!htbp]
\centering
\begin{tabular}{ccc}
\subfigure[Data distribution in concept I] {\includegraphics[width=.23\textwidth]{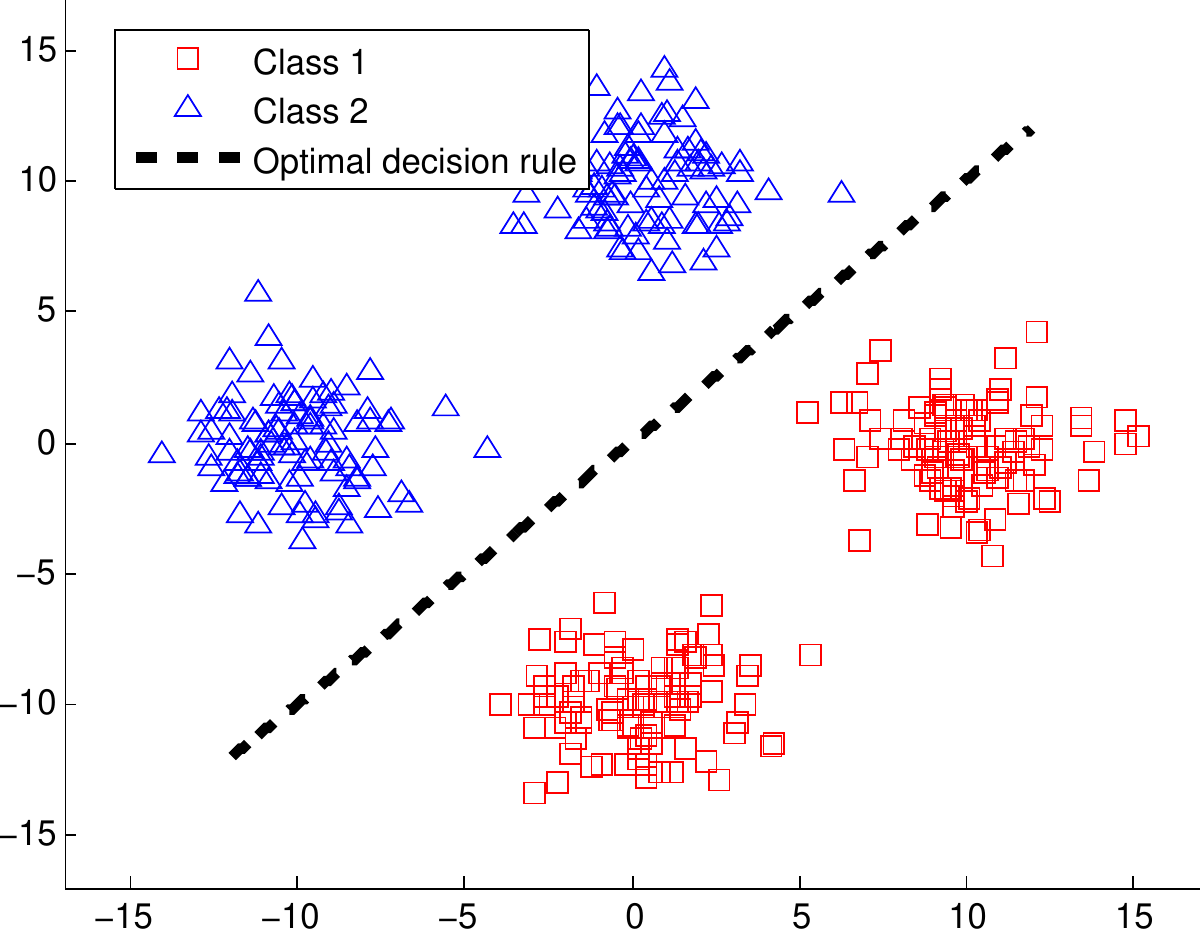}}
\subfigure[Data distribution in concept II] {\includegraphics[width=.23\textwidth]{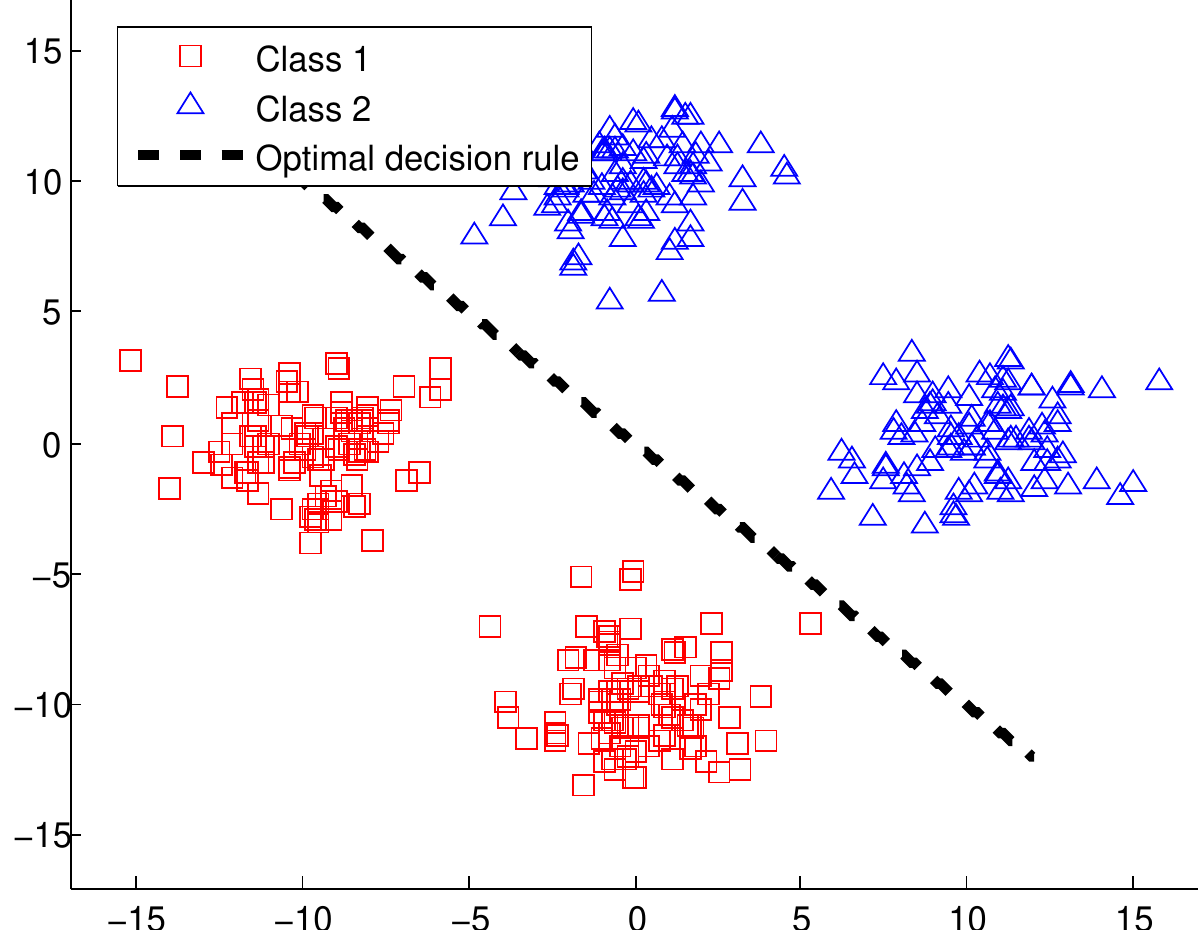}}
\end{tabular}
\caption{The limitations of change point detector on concept drift detection. (a) and (b) demonstrate the feature (i.e., $\ve{X}$) distribution in 2-D plane in two consecutive concepts (selected from the Two Classes Rotating (2CR) dataset \cite{souza2015data}), where the ``red rectangle" denotes class $1$ and the ``blue triangle" represents class $2$. There is no distribution change on $P_t(\ve{X})$ and $P_t(y)$ (because the labels are balanced). The only factor that evolves over time is $P_t(y|\ve{X})$, the optimal decision rule (see the black dashed line).\vspace{-0.0cm}}
\label{fig:demo_motivation}
\end{figure}

A closely related problem to concept drift detection is the classical change point detection that has been well studied theoretically and practically before. Unlike concept drift detectors, change point detectors are targeted at detecting changes in the generating distribution of the streaming data (i.e., $P(\mathbf{X}_t)$) \cite{harel2014concept}. The standard change point detection methods are typically based on statistical decision theory, some reference books include \cite{basseville1993detection,sandberg2001nonlinear,brodsky2013nonparametric,chen2011parametric}. %In general, these methods usually compute a statistic from the available data that is sensitive to changes between the two sets of examples. The measured values of the statistic are then compared to the expected value under the null hypothesis that both samples are from the same distribution. The resulting $p$-value can be seen as a measure of the strength of the drift. A good statistic must be sensitive to detecting changes in data characteristics of samples observed from different distributions.
Although a change point detector may benefit the performance of concept drift detector, purely modeling $P_t(\mathbf{X})$ is insufficient to solve the problem of concept drift detection \cite{sethi2017reliable}. An intuitive example is shown in Fig. \ref{fig:demo_motivation}, in which $P_t(\bm{X})$ remains unchanged, while the class labels change. On the other hand, it still remains a big challenge to detect any type of distributional changes, especially for multivariate or high-dimensional data \cite{harel2014concept,wang2015concept}. For these reasons, instead of selecting the intermediate solution of change point detection, we solve the problem by monitoring the ``significant" drift in the prediction risk of the underlying predictor based on the risk minimization principle~\cite{vapnik1991principles}. %This is motivated by the fact that any drift of $P(\hat{y}_t, y_t)$ would imply a drift in $P(\mathbf{X}_t, y_t)$ with probability $1$ \cite{billingsley2008probability}, where $\hat{y}_t$ is the predicted class label using the current classifier $\hat{f}$ (i.e., $\hat{y}_t=\hat{f}(\bm{X}_t)$).

\subsection{Benchmarking concept drift detection approaches} \label{section1_2}
An extensive review on learning under concept drifts is beyond the scope of this paper, and we refer interested readers to some recently published surveys \cite{gama2014survey,wang2017systematic,krawczyk2017ensemble} for some classical methods and recent progresses. In this section, we only review previous work of most relevance to the presented method, i.e., concept drift detection approaches.

The method that renewed attention to this problem was the Drift Detection Method (DDM) \cite{gama2004learning}. DDM monitors the sum of overall classification error ($\hat{P}^{(t)}_{error}$) and its empirical standard deviation ($\hat{S}^{(t)}_{error}=\sqrt{\hat{P}^{(t)}_{error}(1-\hat{P}^{(t)}_{error})/t}$).
Despite its simplicity, DDM always fails to detect real drift points unless the sum of the Type-I and Type-II errors changes.
Early Drift Detection Method (EDDM) \cite{baena2006early}, on the other hand, suggests monitoring the distance between two consecutive classification errors. EDDM performs better than DDM, especially in the scenario of slow gradual changes.
However, it requires waiting for a minimum of $30$ classification errors before calculating the monitoring statistic at each time instant, an impractical condition for imbalanced data. A third error based method, i.e., STEPD \cite{nishida2007detecting}, applies a test of equal proportion to compare the classification accuracy in a recent window with the historical classification accuracy excluding this recent window.

Following the early work, a few new methods have been proposed to improve DDM from different perspectives. Drift Detection Method for Online Class Imbalance (DDM-OCI) \cite{wang2013concept} deals with imbalanced data. Unfortunately, DDM-OCI is prone to trigger lots of false positives due to an inherent weakness in the model: the test statistic used by DDM-OCI $\hat{R}^{(t)}_{tpr}$ is not approximately distributed as $\mathcal{N}(P^{(t)}_{tpr},\frac{P^{(t)}_{tpr}(1-P^{(t)}_{tpr})}{t})$ under the null hypothesis\footnote{$\hat{R}^{(t)}_{tpr}$ is a modified estimator of $P^{(t)}_{tpr}$, which satisfies $\hat{R}^{(t)}_{tpr}=\eta\hat{R}^{(t-1)}_{tpr}+(1-\eta)1_{y_t=\hat{y}_t}$ where $\eta$ denotes a time decaying factor \cite{wang2013concept}.}~\cite{wang2015concept}.
%Hence, the $tpr$ rationale of constructing confidence levels specified in \cite{gama2004learning} is not suitable with the null distribution of $\hat{R}^{(t)}_{tpr}$.
PerfSim \cite{antwi2012perfsim} also deals with imbalanced data. Different from DDM-OCI, PerfSim tracks the cosine similarity of four entries associated with confusion matrix to determine an occurrence of concept drift. However, the threshold used to distinguish concept drift was user-specified.  Moreover, PerfSim assumes the data comes in batch-incremental manner \cite{read2012batch} which makes it impractical in real applications, especially when the decisions are required to be made instantly. Other related work includes the Exponentially Weighted Moving Average (EWMA) for concept drift detection (ECDD) \cite{ross2012exponentially} and the Drift Detection Method based on the Hoeffding's inequality (HDDM) \cite{frias2015online}. An experimental comparative study is available in \cite{gonccalves2014comparative}.

%On the other hand, Exponentially weighted moving average (EWMA) charts \cite{roberts1959control} was adapted for concept drift detection in ECDD \cite{ross2012exponentially} to control the computational complexity and false positive rate. The test statistic ECDD monitors is the Bernoulli parameter $p$ estimated from the overall classification error, and a significant increase or decrease of $p$ indicates a concept drift.

%Linear Four Rates (LFR) \cite{wang2015concept} was recently proposed to address the limitation of single test statistic by monitoring the four rates associated with the confusion matrix of the data stream. Although, LFR outperforms aforementioned methods, it still triggered a number of false alarms. Thus, a novel architecture (or framework), which can take advantage of LFR but also control the number of false alarms, becomes an essential task for concept drift detection.

\subsection{Hierarchical architecture on change-point or concept drift detection} \label{section1_3}

Hierarchical architectures have been extensively studied in the machine learning community in the last decades. One of the most recent examples is the Deep Predictive Coding Networks (DPCN) \cite{principe2014cognitive}, a neural-inspired hierarchical generative model which is effective on modeling sensory data.

However, the hierarchical architectures for change point (or concept drift) detection were seldom investigated. The first hierarchical change point test (HCDT) was proposed in \cite{alippi2011hierarchical} based on the Intersection of Confidence Intervals (ICI) rule \cite{alippi2010change}. It has later been extended in a higher perspective by incorporating a general methodology to design HCDT \cite{alippi2016hierarchical}. However, as a change point detector, HCDT has its intrinsic limitations as emphasized in section \ref{section1_1}. Although it can be modified for concept drift detection by tracking the classification error with a Bernoulli distribution assumption, a univariate indicator (or statistic) is insufficient to provide accurate concept drift detection \cite{wang2015concept}, especially when the classifier becomes unstable. Moreover, we already proved that the derived statistics (in Layer-I) are geometrically weighted sum of Bernoulli random variables~\cite{wang2015concept}, rather than simply following the Bernoulli distribution in the common sense.

%Moreover, the indicators or statistics mentioned in \cite{alippi2016hierarchical} are typically computed in a subsequence window. A key assumption behind window-based strategy is that the data comes in batch-incremental manner \cite{read2012batch} or the system needs to wait for a windowed time period before making a decision. This is not well suited for real applications, as the data typically comes in a fully online manner, i.e., instance-by-instance.

%Following \cite{alippi2011hierarchical}, the same authors also developed a hierarchical architecture to handle recurrent concept drifts \cite{alippi2013just}, a very special type of concept drifts \cite{gama2014survey}. This work provides a systematic and flexible solution towards learning under recurrent concept drifts. However, from our perspective, it concentrates more on boosting streaming classification performance, as it does not provide any comparison with benchmarking concept drift detectors, like DDM \cite{gama2004learning}, etc. Besides, the Layer-II test is difficult to be extended for other types of concept drift. Moreover, the concept drift adaptation strategy in \cite{alippi2013just} (i.e., retraining a classifier using all the samples from the same concept) contradicts the \textbf{single pass} criterion \cite{ross2012exponentially,domingos2003general} that is widely acknowledged in streaming data mining algorithms design.

This work is motivated by \cite{alippi2016hierarchical}. However, in order to make the designed algorithm well suited for broader classes of concept drift detection (rather than change point detection) without losing accuracy and proper classifier adaptation, we proposed HLFR, a novel hierarchical architecture (together with two novel testing methods in each layer) for concept drift detection that is applicable to different concept drift types and data stream distributions (e.g., balanced or imbalanced labels). Moreover, we present an adaptive training approach instead of the retraining scheme commonly employed, once a drift is confirmed. The proposed adaptation approach is not limited to a single concept drift type and strictly follows the \textbf{single pass} criterion that does not need any historical data. Results show that the proposed approach captures more information from the data than previous work.

\section{Hierarchical Linear Four Rates (HLFR)} \label{section2}

\begin{figure}
\centering
\includegraphics[height=3.5cm,width=8.5cm]{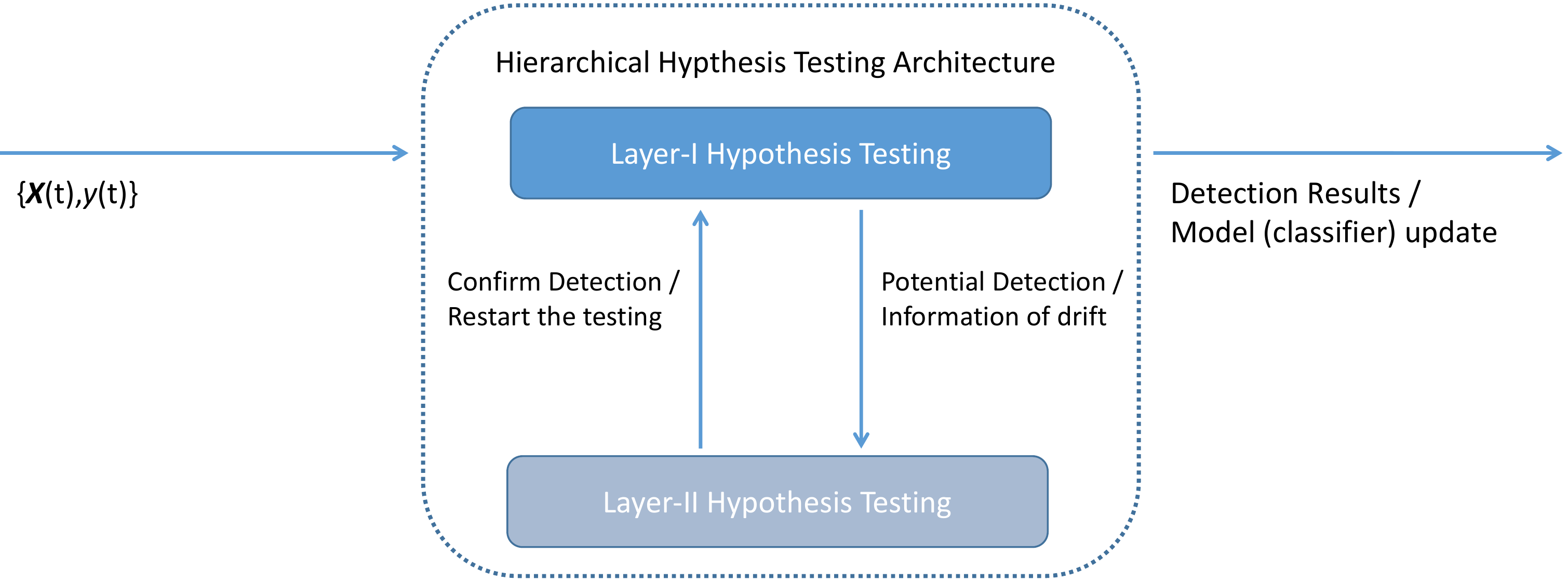}
\caption{The proposed hierarchical hypothesis testing (HHT) framework for concept drift detection and adaptation.\vspace{-0.5cm}}
\label{fig:hierarchical architecture}
\end{figure}

This section presents a novel hierarchical hypothesis testing (HHT) framework for concept drift detection and adaptation.
As shown in Fig.~\ref{fig:hierarchical architecture}, HHT features two layers of hypothesis tests. The Layer-I test is executed online. Once it detects a potential drift, the Layer-II test is activated to confirm (or deny) the validity of the suspected drift. Depending on the decision results of Layer-II test, the HHT reconfigures or restarts the Layer-I test correspondingly. A new concept drift detector, namely Hierarchical Linear Four Rates (HLFR), is developed under the HHT framework. HLFR implements a sequential hypothesis testing \cite{helstrom1968statistical,siegmund1985sequential}, and the two layers cooperate closely to improve online classification capability jointly. HLFR, is summarized in Algorithm \ref{HierarchicalAlg}.

%unlike concept drift detectors that are designed to only work with linear discriminant classifiers \cite{kuncheva2008adaptive}, support vector machines (SVM) \cite{klinkenberg2000detecting}, etc.
%This paper selects soft margin SVM as the baseline classifier due to its accuracy and robustness \cite{bousquet2002stability}. We also test the drift detection performance using other classifiers in experiments.

HLFR does not make use of any intrinsic property or impose any assumption on the underlying classifier. This modular property enables HLFR to be easily deployed with any classifier (support vector machine (SVM), $k$-nearest neighbors (KNN), etc.). It is worth noting that ensemble of detectors~\cite{Du2015A,maciel2015lightweight} may appear to share similarities with the proposed HHT framework in this paper. However, the two architectures are significantly different in the way to organize different hypothesis tests. For HHT, the Layer-II test is only activated when the Layer-I test detects a suspected drift points (i.e., the Layer-II is an auxiliary and validation module to Layer-I in the hierarchical architecture), whereas the ensemble of detectors conducts different tests in a parallel manner (i.e., each test is performed independently and synchronously with no priority, and the final decision is made by a voting scheme). To further illustrate the differences, a rigorous investigation of the Type-I and Type-II errors analysis concerning our HHT framework and the ensemble of detectors are illustrated in Section~\ref{sec:error_analysis}.

\begin{algorithm}[htb]
\caption{Hierarchical Linear Four Rates (HLFR)}
\label{HierarchicalAlg}
\begin{algorithmic}[1]
\Require
Data $\{\mathbf{X}_t, y_t\}_{t=1}^\infty$ where $\mathbf{X}_t\in \mathbb{R}^d$, $y_t\in \{0,1\}$; Initially trained classifier $\hat{f}(\cdot)$.
\Ensure
Concept drift time points $\{T_{cd}\}$.
\For {$t = 1$ to $\infty$}
\State Perform Layer-I hypothesis test.
\If {(Layer-I detects potential drift point $T_{pot}$)}
\State Perform Layer-II hypothesis test on $T_{pot}$
\If {(Layer-II confirms the potentiality of $T_{pot}$)}
\State $\{T_{cd}\}\leftarrow T_{pot}$; Update $\hat{f}(\cdot)$.
\Else
\State Discard $T_{pot}$; Restart Layer-I test.
\EndIf
\EndIf
\EndFor
\end{algorithmic}
\end{algorithm}

\subsection{Layer-I Hypothesis Test}

HLFR selects our recently developed Linear Four Rates (LFR)~\cite{wang2015concept} in its Layer-I test. According to the results shown in~\cite{wang2015concept}, LFR always exhibits promising performances in terms of shorter detection delay and higher detection precision, compared with other prevalent concept drift detectors.
This is not surprising, as LFR monitors four rates (or statistics) associated with a confusion matrix (i.e., the true positive rate ($P_{tpr}$), the true negative rate ($P_{tnr}$), the positive predictive value ($P_{ppv}$) and the negative predictive value ($P_{npv}$)) simultaneously, thus it can sufficiently and precisely make use of the error information.

The key idea for LFR is straightforward: {$P_{tpr}$, $P_{tnr}$, $P_{ppv}$, $P_{npv}$} should remain the same in a stable or stationary concept. Therefore, a significant change of any $P_\star$ ($\star\in{tpr,tnr,ppv,npv}$) may imply a change in the underlying joint distribution $P_t(\mathbf{X},y)$ or concept. Specifically, at each time instant $t$, LFR conducts four independent tests with the following null and alternative hypotheses:
\begin{eqnarray*}
H_0: \forall_{\star}, P(\hat{P}_{\star}^{(t-1)})=P(\hat{P}_{\star}^{(t)})\\
H_A: \exists_{\star}, P(\hat{P}_{\star}^{(t-1)})\neq P(\hat{P}_{\star}^{(t)})\\
\star\in\{tpr,tnr,ppv,npv\}
\end{eqnarray*}
The concept is stable if $H_0$ hypothesis holds and is considered to have a potential drift if $H_0$ hypothesis is rejected. Intuitively, LFR should be more sensitive to any type of drift, as it keeps track of four rates simultaneously. By contrast, almost all previous methods use a single specific statistic that can only capture partial of the distributional information: DDM, ECDD and HDDM use the overall error rate, EDDM relies on the average distance between adjacent classification errors, DDM-OCI deals with the minority class recall, STEPD monitors a ratio of recent accuracy and overall accuracy, whereas PerfSim considers the cosine similarity coefficient of four entries in confusion matrix.

The LFR is summarized in Algorithm \ref{FourRatesAlg}. During implementation, LFR modifies $P_\star^{(t)}$ with $R_\star^{(t)}$ as employed in \cite{wang2013concept,wang2013learning} (see also footnote $3$). $R_\star^{(t)}$ is essentially a weighted linear combination of the classifier's current and previous performances. In \cite{wang2015concept}, we have proved that $R_\star^{(t)}$ follows a weighted independent and identically distributed ($i.i.d.$) Bernoulli distribution. Given this property, we are able to obtain the ``BoundTable" by conducting Monte-Carlo simulations. Based upon these bound values, LFR considers that a concept drift is likely to occur when any $R_\star^{(t)}$ succeeds the warning bound (warn.bd), and sets the warning signal ($warn.time\leftarrow t$). If any $R_\star^{(t)}$ reaches the corresponding detection bound (detect.bd), the concept drift is affirmed at ($detect.time\leftarrow t$). Interested readers can refer to \cite{wang2015concept} for more details.

\begin{comment}
Note that, HLFR includes two modifications to LFR. First, we update the time decaying factor $\eta_\star$ with\footnote{Note that, this time-varying representation of $\eta_\star^{(t)}$ is hyperparameter free and self-bounded between $(2\eta_\star^{(t-1)}-1,1)$ with a ``sigmoid" like curve.}:
$$\small{\eta_\star^{(t)}=
\begin{cases}
(\eta_\star^{(t-1)}-1)e^{-(R_\star^{(t)}-R_\star^{(t-1)})}+1& R_\star^{(t)}\geq R_\star^{(t-1)}\\
(1-\eta_\star^{(t-1)})e^{R_\star^{(t)}-R_\star^{(t-1)}}+(2\eta_\star^{(t-1)}-1)& R_\star^{(t)}<R_\star^{(t-1)}
\end{cases}}$$
This adaptation is motivated from the adaptive signal processing domain \cite{haykin2008adaptive}. The key idea is that a larger time decaying factor shall be used when $R_\star$ is increasing, which indicates the classifier tends to perform better with most recent data. Moreover, we applied polynomial curve fittings to BoundTable so that a closed-form function can improve calculations of warning/detection bounds in terms of both memory usage and resolution. Unless otherwise specified, the LFR mentioned in this paper refers to the modified one.
\end{comment}

\begin{algorithm}[htb]
\caption{Linear Four Rates (Layer-I)}
\label{FourRatesAlg}
\begin{algorithmic}[1]
\Require
Data $\{(\mathbf{X}_t, y_t)\}_{t=1}^{\infty}$ where $\mathbf{X}_t \in \mathbb{R}^d$ and $y_t \in \{0,1\}$;
Binary classifier  $\hat{f}(\cdot)$;
Time decaying factors $\eta_{\star}$;
warn significance level $\delta_{\star}$;
detect significance level $\epsilon_{\star}$.
\Ensure
Potential concept drift time points $\{T_{pot}\}$.
\State $R^{(0)}_{\star} \leftarrow 0.5, $
%	\[(R^{(0)}_{tpr}, R^{(0)}_{tnr}, R^{(0)}_{ppv}, R^{(0)}_{npv}) \leftarrow (0.5, 0.5, 0.5, 0.5),\]
$\hat{P}^{(0)}_{\star} \leftarrow 0.5$ where $\star \in \{tpr, tnr, ppv, npv\}$
%	\[(P^{(0)}_{tpr}, P^{(0)}_{tnr}, P^{(0)}_{ppv}, P^{(0)}_{npv}) \leftarrow (0.5, 0.5, 0.5, 0.5)\]
and confusion matrix
$C^{(0)} \leftarrow   \left( \begin{array}{cc}
1 & 1 \\
1 & 1 \end{array} \right)$;
\For  {$t=1$ to $\infty$}
\State $\hat{y}_t \leftarrow \hat{f}(\mathbf{X}_t)$
%\Comment {\%Classify $\mathbf{X}_t$ using $\hat{f}(\star)$ \%}
\State $C^{(t)}[\hat{y}_t][y_t] \leftarrow C^{(t-1)}[\hat{y}_t][y_t] + 1$

\While {($\star \in \{tpr, tnr, ppv, npv\}$)}
\If{($\star$ is influenced by $(y_t, \hat{y}_t)$)}    %\Comment{ \% There are two rates out of four are invovled in current instance where one from $\{tpr, tnr\}$ and the other from $\{ppv, npv\}$\%}
\State $R^{(t)}_{\star} \leftarrow \eta_{\star}R^{(t-1)}_{\star} + (1-\eta_{\star}) \mathbf{1}_{\{y_t = \hat{y}_t\}}$ %\Comment{\%Update the influenced rate\%}
\Else
\State $R^{(t)}_{\star} \leftarrow R^{(t-1)}_{\star}$;
\EndIf

\If {( $\star \in \{tpr, tnr\}$)}
\State $N_\star \leftarrow C^{(t)}[ 0, \mathbf{1}_{\{\star = tpr\}}] + C^{(t)}[1, \mathbf{1}_{\{\star = tpr\}}]$
\State $\hat{P}^{(t)}_\star \leftarrow \dfrac{C^{(t)}[ \mathbf{1}_{\{\star = tpr\}},  \mathbf{1}_{\{\star = tpr\}}]}{N_\star}$
\Else
\State $N_\star \leftarrow C^{(t)}[ \mathbf{1}_{\{\star = ppv\}}, 0] + C^{(t)}[ \mathbf{1}_{\{\star = ppv\}}, 1]$
\State $\hat{P}^{(t)}_\star \leftarrow \dfrac{C^{(t)}[ \mathbf{1}_{\{\star = ppv\}},  \mathbf{1}_{\{\star = ppv\}}]}{N_\star}$
\EndIf
\State $\text{warn.bd}_{\star} \leftarrow \text{BoundTable}(\hat{P}^{(t)}_{\star}, \eta_{\star}, \delta_{\star}, N_\star)$
\State $\text{detect.bd}_{\star} \leftarrow \text{BoundTable}(\hat{P}^{(t)}_{\star}, \eta_{\star}, \epsilon_{\star}, N_\star)$
\EndWhile
\If{ ( any $R^{(t)}_\star$ exceeds $\text{warn.bd}_\star$ \& warn.time is NULL) }
\State $\text{warn.time} \leftarrow t$
\ElsIf {(no $R^{(t)}_\star$ exceeds $\text{warn.bd}_\star$ \& warn.time is not NULL)} %\Comment {\%$R^{(t)}_\star$ back to normality, warning removed\%}
\State $\text{warn.time} \leftarrow$ NULL
\EndIf
\If{ ( any $R^{(t)}_\star$ exceeds $\text{detect.bd}_\star$ ) }
	%    \State $\text{previous.detection} \leftarrow t$
\State detect.time $\leftarrow t$;
\State relearn $\hat{f}(\cdot)$ by $\{(\mathbf{X}_t, y_t)\}_{t = \text{warn.time}}^{\text{detect.time}}$ or wait for sufficient instances;
\State reset $R^{(t)}_{\star}, \hat{P}^{(t)}_{\star}, C^{(t)}$ as step 1;
\State $\{T_{pot}\} \leftarrow t$.
\EndIf
\EndFor
\end{algorithmic}
\end{algorithm}

%\begin{algorithm}[htb]
%\label{BoundTable}
%\caption{BoundTable Generation in LFR}
%\begin{algorithmic}[1]
%\Require
%Estimate of underlying rate $\hat{P}$;
%Time decaying factor $\eta$;
%significance level $\alpha$;
%number of time steps $N_\star$;
%number of Monte-Carlo simulations $num.of.MC$;
%\Ensure
%a numeric bound value corresponding to required significance level.
%\For {$j = 1$ to $num.of.MC$}
%\State Generate $N_\star$ independent Bernoulli random variables $\{I_1,I_2,\dots, I_{N_\star}\}$ where %$I_i \stackrel{iid}{\sim} Bernoulli(\hat{P})$
%\State $R[j] \leftarrow (1-\eta)\sum_{i=1}^{N_\star}\eta^{N_\star - i}I_i$
%\EndFor
%\State $\{R[j]\}_{j=1}^{num.of.MC}$ forms an empirical distribution $\hat{F}(R)$,
%find the $\alpha-$level quantile as the lower bound $lb \leftarrow quantile(\hat{F}(R), \alpha)$,
%find the $(1-\alpha)-$level quantile as the upper bound $ub \leftarrow quantile(\hat{F}(R), 1-\alpha)$.
%\end{algorithmic}
%\end{algorithm}

\subsection{Layer-II Hypothesis Test}
The four rates are more sensitive metrics that enable LFR to be able to promptly detect any types of concept drifts. However, the sensitivity of four rates also makes LFR is more likely to trigger ``false positive" detections. The Layer-II test serves to validate detections raised by Layer-I test, thus significantly remove these ``false positive" detections. In HLFR, we use a permutation test (see Algorithm \ref{PermutationAlg}) in its Layer-II test. Permutation test has been well studied theoretically and practically before, it does not require apriori information regarding the monitored process or the nature of the data \cite{good2013permutation}.

Specifically, we partition the streaming observations into two consecutive segments based on the suspected drift instant $T_{pot}$ provided by the Layer-I test, and employ a new statistical hypothesis test to compare the inherent properties of these two segments to assess a possible variations in the joint distribution $P_t(\mathbf{X}, y)$.
Then, the general idea behind our designed permutation test is to test whether the prediction average risk (evaluated over the second segment using a classifier trained on the first segment) is significantly different from its sampling distribution under the null hypothesis (i.e., no drift occurs). Here, we measure the prediction average risk with zero-one loss. Zero-one loss contains partial information of the four rates.
Intuitively, if no concept drift has occurred, the zero-one loss on the ordered train-test split (i.e., $\hat{E}_{ord}$ in line $4$) should not deviate too much from that of the shuffled splits (i.e., $\hat{E}_i$, $i=1,2,\cdots,P$, in line $8$), a realization of its sampling distribution under the null hypothesis~\cite{harel2014concept}.

%We want to emphasise that the selection of the test statistic used at Layer-II should be strictly related to that used at Layer-I. To this end, we choose zero-one loss over the ordered train-test split, as the test statistic in the Layer-II test.

\begin{algorithm}[htb]
\caption{Permutation Test (Layer-II)}
\label{PermutationAlg}
\begin{algorithmic}[1]
\Require
Potential drift time $T_{pot}$;
Permutation window size $W$;
Permutation number $P$;
Classification algorithm $\mathbf{A}$;
Significant rate $\eta$.
\Ensure
\emph{decision} ($True$ $positive$ or $False$ $positive$?).
\State $S_{ord}\leftarrow$ streaming segment before $T_{pot}$ of length $W$.
\State $S'_{ord}\leftarrow$ streaming segment after $T_{pot}$ of length $W$.
\State Train classifier $f_{ord}$ on $S_{ord}$ using $\mathbf{A}$.
\State Test classifier $f_{ord}$ on $S'_{ord}$ to get the zero-one loss $\hat{E}_{ord}$.
\For {$t = 1$ to $P$}
\State $(S_i, S'_i)\leftarrow$ random split of $S_{org}\bigcup S'_{org}$.
\State Train classifier $f_i$ on $S_i$ using $\mathbf{A}$.
\State Test classifier $f_i$ on $S'_i$ to get the zero-one loss $\hat{E}_i$.
\EndFor
\If {$\frac{1+\sum\nolimits_{i=1}^P\mathbf{1}[\hat{E}_{ord}\leq\hat{E}_i]}{1+P}\leq\eta$}
\State \emph{decision}$\leftarrow$$T_{pot}$ is $True$ $positive$.
\Else
\State \emph{decision}$\leftarrow$$T_{pot}$ is $False$ $positive$.
\EndIf\\
\Return \emph{decision}
\end{algorithmic}
\end{algorithm}

\subsection{Error analysis on Hierarchical Hypothesis Testing} \label{sec:error_analysis}

To further give credence to the success of HHT framework in practical applications, we present a theoretical analysis to its associated Type-I and Type-II errors.

In the problem of concept drift detection, the Type-I error (also known as a ``false positive" rate) refers to the incorrect rejection of a true null hypothesis $H_0$ (i.e., no drift occurs). By contrast, the Type-II error (also known as a ``false negative" rate) is incorrectly retaining a false null hypothesis when the alternative hypothesis $H_A$ is true. On the other hand, for any (single-layer) hypothesis test, the Type-I error $\alpha$ is exactly the selected significance level, whereas the Type-II error (denoted with $\beta$) is determined by the power of the test and the power is exactly $(1-\beta)$.

Let us denote by $\alpha_1$ and $\beta_1$ the Type-I and Type-II errors of Layer-I test, and $\alpha_2$ and $\beta_2$ the Type-I and Type-II errors of Layer-II test. Also, we denote by $\alpha$ and $\beta$ the overall Type-I and Type-II errors of HHT framework.

%\subsubsection{The $\alpha$ and $\beta$ of HHT}
By definition, the Type-I error $\alpha$ of HHT is given by:
\begin{equation} \label{eq1}
\begin{split}
\alpha & = P(\text{HHT rejects}~H_0|H_0) \\
 & = P(\{\text{Layer-I rejects}~H_0\}\&\{\text{Layer-II rejects}~H_0\}|H_0) \\
 & = P(\text{Layer-I rejects}~H_0|H_0)\times P(\text{Layer-II rejects}~H_0|H_0) \\
 & = \alpha_1\alpha_2
\end{split}
\end{equation}
where ``$\&$" denotes \textbf{AND} logic operator.

Eq.~(\ref{eq1}) assumes that the performance of Layer-I test and Layer-II test is independent, i.e., the detection results of Layer-I and Layer-II tests will not be mutually influenced when they are being tested independently. Given that the test statistics and manners are totally different in Layer-I and Layer-II tests of HLFR, this assumption makes sense. In fact, even though the performance of Layer-I and Layer-II tests are related to each other, $\alpha$ still satisfies $\alpha\leq\max(\alpha_1,\alpha_2)$, which suggests that the HHT framework will not increase the Type-I error even in the worst case.

%\footnote{For example, if we DDM~\cite{gama2004learning} in Layer-I and HDDM~\cite{frias2015online} in Layer-II, it is far-fetched to claim that the performance of Layer-I test is independent of Layer-II test. This is because both these two tests monitor the moving average of the overall classification error (i.e., the same test statistic), the only difference is that DDM is a parameter test whereas HDDM is non-parametric.}, $\alpha$ still satisfies $\alpha\leq\max(\alpha_1,\alpha_2)$, which suggests that the HHT framework will not increase the Type-I error even in the worst case.

Similarly, the overall Type-II error $\beta$ is given by:
\begin{equation} \label{eq2}
\begin{split}
\beta & = P(\text{HHT fails to reject}~H_0|H_A) \\
 & = P(\text{Layer-I fails to reject}~H_0|H_A) \\
 &\quad +P(\{\text{Layer-I rejects}~H_0\}\&\{\text{Layer-II fails to reject}~H_0\}|H_A) \\
 & = P(\text{Layer-I fails to reject}~H_0|H_A) \\
 &\quad +P(\text{Layer-I rejects}~H_0|H_A)\times P(\text{Layer-II fails to reject}~H_0|H_A) \\
 & = \beta_1+(1-\beta_1)\beta_2
\end{split}
\end{equation}

Again, we assume the performance independence of Layer-I and Layer-II tests. However, even though this condition is not met, we still have $\beta_1\leq\beta\leq\beta_1+\max(1-\beta_1,\beta_2)$. This is an unfortunate fact, as it suggests a fundamental limitation of the HHT framework: it may increase the Type-II error. Given the fact that majority of the current concept drift detectors have high detection power (i.e., $\beta$ is small) yet suffer from a relatively high ``false positive" rate, the cost is acceptable.

As emphasized earlier, a similar architecture to the proposed HHT framework is the ensemble of detectors~\cite{Du2015A,maciel2015lightweight}. The most widely used decision rule for ensemble of detectors is that, given a pool of candidate detectors, the system determines a drift if any one of the detectors finds a drift. This way, suppose there are $K$ candidate detectors, the Type-I and Type-II errors of the ensemble of detectors are given by (assuming pairwise performance independence~\cite{wozniak2016ensembles}):
\begin{equation} \label{eq3}
\begin{split}
\alpha & = P(\text{at least one of the ensemble detectors rejects}~H_0|H_0) \\
 & = 1-P(\text{all detectors do not reject}~H_0|H_0) \\
 & = 1-P(\{1\text{st detector does not reject}~H_0\}\&\ \\
 &\quad \cdots\&\{K\text{-th detector does not reject}~H_0\}|H_0) \\
 & = 1-P(1\text{st detector does not reject}~H_0|H_0)\times \\
 &\quad \cdots\times P(K\text{-th detector does not reject}~H_0|H_0)\\
 & = 1-(1-\alpha_1)(1-\alpha_2)\cdots(1-\alpha_K)
\end{split}
\end{equation}
\begin{equation} \label{eq4}
\begin{split}
\beta & = P(\text{all detectors fails to reject}~H_0|H_A) \\
 & = P(1\text{st detector fails to reject}~H_0|H_A)\times \\
 &\quad \cdots\times P(K\text{-th detector fails to reject}~H_0|H_A) \\
 & = \beta_1\beta_2\cdots\beta_K
\end{split}
\end{equation}

By referring to Eqs.~(\ref{eq1})-(\ref{eq4}), it is easy to find that, although the architecture of HHT and the ensemble of detectors look similar, their functionalities and mechanisms are totally different. HHT attempts to remove ``false positive" detections as much as possible, thus significantly decreases the Type-I error. However, HHT may increase the Type-II error at the same time. The ensemble of detectors, on the other hand, aim to further improve detection power (thus decrease the Type-II error) at the cost of increased Type-I error\footnote{\scriptsize{$1-(1-\alpha_1)(1-\alpha_2)\cdots(1-\alpha_K)\geq1-(1-\alpha_i)=\alpha_i$, ($i=1,2,\cdots,K$).}}. Given that the prevalent concept drift detectors always have high detection power (e.g., LFR and HDDM) yet suffer from lots of ``false positive" detections, it may not be necessary to naively combine different detectors in an ensemble manner. This is also the reason why the ensemble of detectors do not demonstrate any performance gain over single-layer-based drift detectors in a recent experimental survey paper~\cite{wozniak2016ensembles}.

%Having illustrated the analytical expressions for $\alpha$ and $\beta$ of both HHT and the ensemble of detector, as well as their intrinsic connections, we specify the detailed values of $\{\alpha_1,\beta_1\}$ in Layer-I test and $\{\alpha_2,\beta_2\}$ in Layer-II test of HLFR in the supplementary material for completeness.

%Given (1) and (2), to have an analytical expression for $\alpha$ and $\beta$, we need to specify $\beta_1$ and $\beta_2$.

Having illustrated the analytical expressions for the overall Type-I and Type-II errors of the HHT framework (i.e.,  $\{\alpha,\beta\}$), we now specify the detailed values of Type-I and Type-II errors in Layer-I test (i.e., $\{\alpha_1,\beta_1\}$) as well as the Type-I and Type-II errors in Layer-II test (i.e., $\{\alpha_2,\beta_2\}$) of our proposed HLFR algorithm for completeness. We have $\alpha=\alpha_1\alpha_2$ and $\beta=\beta_1+(1-\beta_1)\beta_2$.

\subsubsection{The $\alpha_1$ and $\beta_1$ of Layer-I test}

The Type-I error $\alpha_1$ of Layer-I test is upper bounded by its detection significance level (i.e., $\epsilon_{\star}$ in Algorithm~\textcolor{blue}{2} of manuscript). On the other hand, although the test statistics $R_{\star}$ ($\star\in\{tpr,tnr,ppv,npv\}$) are geometrically weighted sum of Bernoulli random variables under a stable concept (i.e., $H_0$ hypothesis) up to time $T$, i.e., $R_\star^{(T)}=(1-\eta_\star)\sum_{i=1}^{N_\star}\eta_\star^{N_\star-i}I_i$, where $\{I_i\}_{i=1}^{N_\star}\stackrel{i.i.d.}{\sim}Bernoulli(P_\star)$ and $P_\star$ is the underlying rate, two reasons make it impossible to get a close-form expression or upper bound for Type-II error $\beta_1$ of Layer-I test.

1) It is hard to obtain the closed-form distribution function of $R_\star^{(T)}$ under $H_0$. Although \cite{bhati2011distribution} investigated the closed-form distribution function of $R_\star^{(T)}$ under $H_0$ for the special case $P_\star=0.5$, it still remains a question for other values of $P_\star$.

2) The closed-form distribution function of $R_\star^{(T)}$ under $H_A$ is unattainable. This is because $R_\star^{(T)}$ could have arbitrary (or unconstrained) distributions when the concept changes.

\begin{figure}[!htbp]
\centering
\includegraphics[height=7.5cm,width=8.5cm]{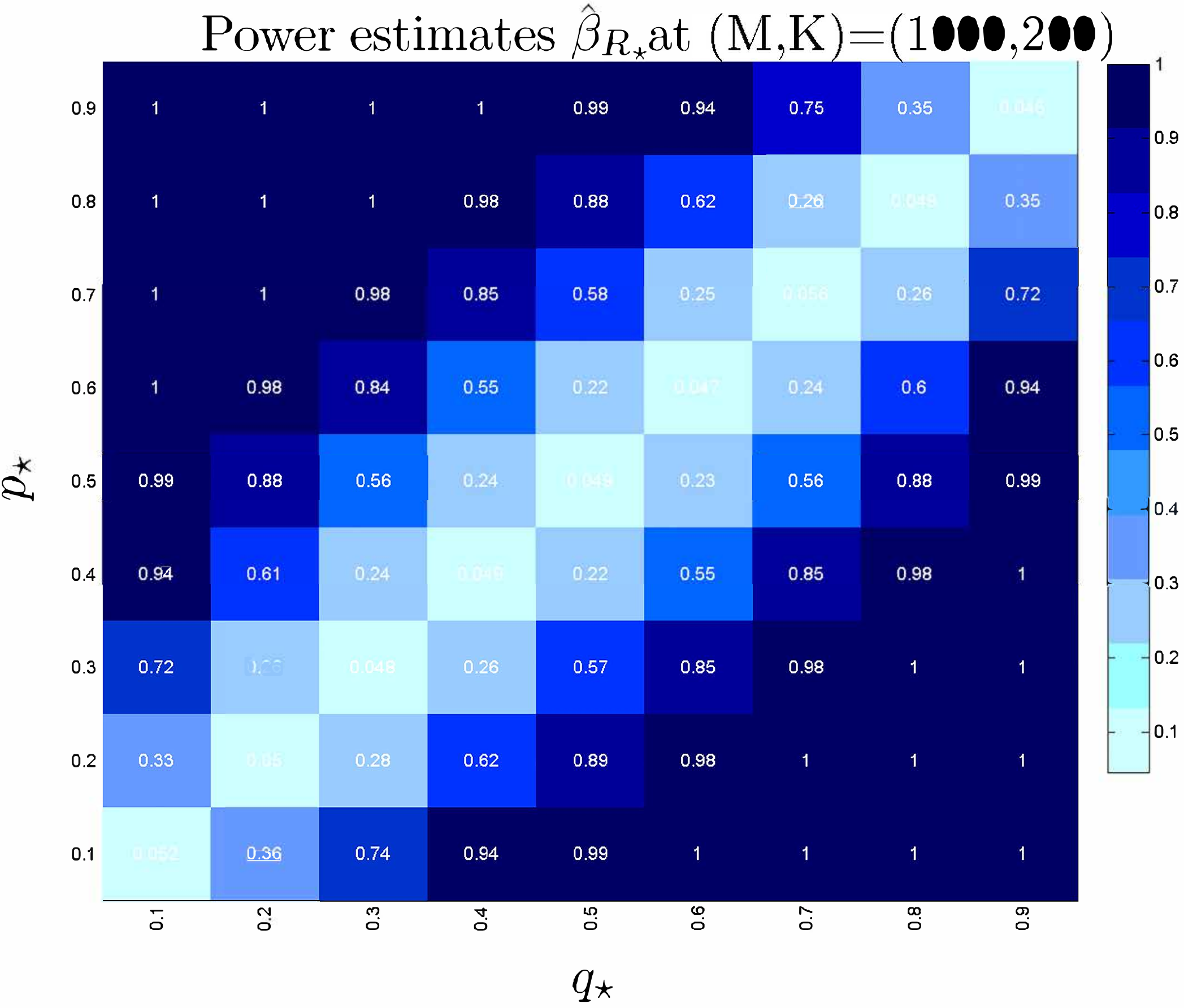}
\caption{Heatmap of power estimate $\hat{\beta}_{R_*}$.\vspace{-0.2cm}}
\label{fig:power_test}
\end{figure}

Therefore, this section only empirically investigates the power of Layer-I test using synthetic data to illustrate and reveal its properties. We denote by $\hat{\beta}_{R_*}$ the power estimate of $R_*^{(t)}$. Suppose the null distribution is at $t=M$ and alternative distribution is at $t=M+k$, $1\leq k\leq K$, where $K$ is the maximal detection time delay. Then suppose the underlying rate is drifted from $p_*$ (the first concept) to $q_*$ (the second concept). Fig.~\ref{fig:power_test} is a heatmap of limiting power estimates on all $(p_*,q_*)$ pairs using $M=1000,K=200$. We can see that $\hat{\beta}_{R_*}$ is already close to $1$, when $p_*$ and $q_*$ are significantly different. In this case, the Type-II error $\beta_1$ reduces to $0$, because $\beta_{R_*}=1-\beta_1$.

\subsubsection{The $\alpha_2$ and $\beta_2$ of Layer-II test}
Same as the Layer-I test, the Type-I error $\alpha_2$ of the Layer-II test is upper bounded by its selected significance level (i.e., $\eta$ in Algorithm~\textcolor{blue}{3} of manuscript). Thus, we focus our analysis on its power. Before that, we give the following two definitions.

\begin{definition}{\cite{bousquet2002stability}}
An algorithm $\mathbf{A}$ has error stability $\gamma_n$ with respect to loss function $\mathcal{L}$ if:\\
\begin{equation}
\forall Z_n\in\mathcal{Z}^n,~\forall i\in\{1,2,\cdots,n\},~|R_D(\mathbf{A}_{Z_n})-R_D(\mathbf{A}_{Z_n^{-i}})|\leq\gamma_n,\\
\end{equation}
where $\mathbf{A}_S$ refers to a predictor obtained using $\mathbf{A}$ trained on set $S$ with cardinality $n$, $Z_n^{-i}$ is the set $Z_n$ with the sample $i$ removed, and $\gamma_n$ decreases with $n$. $R_D(h)=\mathbb{E}_{z\sim D}[\mathcal{L}(h,z)]$ is the risk of $h\in \mathcal{H}$ with respect to distribution $D$, and $\mathbb{E}$ denotes expectation.
\end{definition}

\begin{definition}{\cite{harel2014concept}}
A stream segment $[t_1,t_2]$ is said to have $\varepsilon$-permitted variations, if for some $\varepsilon>0$, with respect to $h\in \mathcal{H}$, if:\\
\begin{equation}
\max\limits_{i,j\in[t_1,t_2]}|R_{D_i}(h)-R_{D_j}(h)|\leq\varepsilon.\\
\end{equation}
\end{definition}

Given two subsequences with equal length $W$, the Layer-II test in our HLFR method aims to determine whether the average prediction risk on ordered train-test split deviates too much from that of the shuffled splits by testing the following hypothesis:
\begin{eqnarray*}
H_0: |R_{ord}-R_{perm}| &\leq& \triangle\\
H_A: |R_{ord}-R_{perm}| &>& \triangle+g(\varepsilon)\\
\end{eqnarray*}
where $R_{ord}=R_{S'_{ord}}(\mathbf{A}_{S_{ord}})$ denotes the risk on ordered train-test split (i.e., $S_{ord}$ and $S'_{ord}$ in lines $1$ and $2$ of Algorithm~\textcolor{blue}{3}), whereas $R_{perm}=\mathbb{E}_{S\sim U_{2W}}[R_{S'}(\mathbf{A}_{S})]$ denotes the risk on shuffled splits (i.e., $S$ and $S'$ in line $6$ of Algorithm~\textcolor{blue}{3}) and $U_{2W}$ refers to the uniform distribution over all possible~$\binom {2W}{W}$ training sets of size $W$ from the two segments of samples, $\triangle$ is a parameter that controls the maximum allowable change rate and $g(\varepsilon)$ is a $\varepsilon$-related function that will be elaborated in the following theorem.

Having illustrated the essence of Layer-II test, given \textbf{Definition 1} and \textbf{Definition 2}, the following corollary upper bounded its Type-II error $\beta_2$.
\begin{corollary}
For an algorithm with stability $\gamma_n=\mathcal{O}(\frac{1}{n})$ and any $\delta\in(0,1)$, we have that under $H_A$, the probability of obtaining a ``false negative" detection is bounded as follows:
\begin{equation}
P\big[|\hat{E}_{ord}-\hat{E}_{perm}|\leq \Theta\big]\leq\eta.
\end{equation}
\end{corollary}
Here $\Theta=6W\gamma_{W}+\sqrt{4\log{\big(\frac{4}{\eta}\big)}/W}+\triangle+\varepsilon$, in which $W$ and $\eta$ are the permutation window size and the significance rate in Algorithm~\textcolor{blue}{3} of manuscript, $\varepsilon$ refers to the small variation in \textbf{Definition 2} and $\triangle$ denotes the maximum allowable change rate. $\hat{E}_{ord}$ and $\hat{E}_{perm}$ denote the estimated zero-one loss of ordered train-test split and shuffled splits (see Algorithm~\textcolor{blue}{3} for more details). For simplicity, we set $\triangle=0$ in our Algorithm~\textcolor{blue}{3} to avoid introducing extra hyperparameters. Note that, the above corollary is a special example of Theorem $3$ in~\cite{harel2014concept}. Interested readers can refer to the supplementary material of~\cite{harel2014concept} for complete proof.

\subsection{Adaptive Hierarchical Linear Four Rates (A-HLFR)}
Although HLFR can be used for streaming data classification with concept drifts (just like its DDM \cite{gama2004learning}, EDDM \cite{baena2006early} and STEPD \cite{nishida2007detecting} counterparts), naively retraining a new classifier after each concept drift detection severely deteriorates its classification performance. This stems from the fact that once a drift is confirmed, it discards all the (relevant) information from previous experience and uses only limited samples from current concept to retrain a classifier. A promising solution to avoid such circumstance is to first extract such kind of relevant knowledge from past experience and then ``transfer" this knowledge to the new classifier~\cite{gama2014survey,sun2018concept,ditzler2013incremental}. To this end, Adaptive Hierarchical Linear Four Rates (A-HLFR) is an integral part of the proposed solution. A-HLFR makes a simple yet strategic modification to HLFR: replacing re-training scheme in HLFR framework with an adaptive learning strategy. Specifically, we substitute SVM (this paper selects soft margin SVM as the base classifier due to its accuracy and robustness~\cite{bousquet2002stability}) with adaptive SVM (A-SVM) \cite{yang2007cross} once a concept drift is confirmed. The pseudocode of A-HLFR is the same as Algorithm \ref{HierarchicalAlg}. The only exception comes from the layer-I test, where the re-training scheme with standard SVM (line $28$ in Algorithm \ref{FourRatesAlg}) is substituted with A-SVM. %We refer interested readers to~\cite{yang2007cross,yang2007adapting} for more details of A-SVM, such as its formulation and optimization.} %A brief introduction to A-SVM is presented in the supplementary material for completeness.}

\subsubsection{Adaptive SVM - Motivations and Formulations}
A fundamental difficulty for learning supervised models once a concept drift is confirmed, is that the training samples from new and previous concepts are drawn from different distributions. A short detection delay (especially for state-of-the-art concept drift detection methods) results in extremely limited training samples from the new concept. These limited training samples from the new concept, coupled with the fact that it may be likely that consecutive concepts are closely related or relevant, inspires the idea of adapting the previous models with samples from the new concept to boost the concept drift adaptation capability. %Note that, although sample weighting scheme \cite{klinkenberg2004learning} provides a plausible solution, especially for ensembles of classifiers \cite{barua2014mwmote}, the effectiveness is not salient through vast of simulations when using single classifier.

Recall the earlier mentioned problem formulation, we are required to classify samples in the new concept, where only a limited number of labeled samples (i.e., a newly observed \textbf{primary dataset} $D=\{(\mathbf{X}_i,y_i)\}_{i=1}^N$) are available for updating a classifier. To circumvent the drawbacks of limited training samples, the
\textbf{auxiliary classifier} $\hat{f}^a$ training on previously observed fully-labeled \textbf{auxiliary dataset} $D^a=\{(\mathbf{X}_i^a,y_i^a)\}_{i=1}^{N^a}$ should also be considered. This is because the dataset $D$ is sampling from a joint distribution $P(\mathbf{X},y)$ that is related to, yet different from, the joint distribution $P^a(\mathbf{X},y)$ of dataset $D^a$ in an unknown manner. If we apply the auxiliary classifier $\hat{f}^a$ on the \textbf{primary dataset} $D$, the performance is poor since $\hat{f}^a$ is biased to $P(\mathbf{X},y)$. On the other hand, although we can retrain a classifier using samples in $D$ such that the new classifier is unbiased to $P(\mathbf{X},y)$, the classification accuracy may suffers from high variance due to limited training samples.

%Consequently, to classify samples in $D$, the auxiliary classifier $\hat{f}^a$ cannot perform well since it is biased to $P(\mathbf{X},y)$. On the other hand, a new classifier trained only using examples in $D$, although unbiased, may suffers from high variance due to limited training samples.

In order to achieve an improved bias-variance tradeoff, we employ adaptive SVM (A-SVM), initiated in \cite{yang2007cross}, to adapt $\hat{f}^a$ to $D$. Intuitively, the key idea of A-SVM is to learn an adaptive classifier $\hat{f}$ from $\hat{f}^a$ by regularizing the distance between $\hat{f}$ and $\hat{f}^a$, which can be formulated as:
\begin{eqnarray}
\begin{aligned}
\label{eq:A_SVM}
 & \min\limits_{w}\frac{1}{2}\|w-w^a\|^2 +C\sum\limits_{i=1}^{N}\xi_i \\
 & \text{s.t.}\quad \xi_i\geq0 \\
 & \qquad y_iw^T\phi{(\mathbf{X}_i)}\geq1-\xi_i,\forall(\mathbf{X}_i,y_i)\in D
\end{aligned}
\end{eqnarray}
where $\phi$ represents a feature mapping to project sample $\mathbf{X}$ into a high-dimensional space or reproducing kernel Hilbert space (for linear SVM, $\phi(\mathbf{X})=\mathbf{X}$), $w^a$ denotes the classifier parameters estimated from $D^a$. Eq. (\ref{eq:A_SVM}) jointly optimizes the $\ell_2$ distance between $w$ and $w^a$ as well as the classification error. The optimization to A-SVM is presented in~\cite{yang2007cross,yang2007adapting}.

\section{Experiments} \label{experiments}

This section presents three sets of experiments to demonstrate the superiority of HLFR and A-HLFR over the prevalent baseline methods, in terms of concept drifts detection and adaptation. Section \ref{section5.2} validates the benefits and advantages of HLFR on concept drift detection, using both quantitative metrics and visual evaluation.
Section \ref{section5.4} uses two real-world examples (one for email filtering, another for weather prediction) to illustrate the effectiveness and potency of using an adaptive training method to improve the capability of concept drift adaptation. In Section~\ref{section5.5}, we empirically demonstrate that 1) the benefits of adaptive training are not limited to HLFR, i.e., it provides a general solution to classifier adaptation for all concept drift detectors, like DDM, EDDM, etc.; and 2) the concept drift detection capability will not be impacted by the adaptive training strategy, i.e., HLFR and A-HLFR can achieve almost the same concept drift detection precision. Finally, we give a brief analysis to the computational complexity of all competing methods in section \ref{section5.6}. All the experiments mentioned in this work were conducted in MATLAB $2013$a on an Intel i5-3337 $1.80$GHz PC with $6$GB RAM.

\subsection{Concept Drift Detection with HLFR} \label{section5.2}
We first compare the performance of HLFR against five state-of-the-art concept drift detection methods: DDM \cite{gama2004learning}, EDDM \cite{baena2006early}, DDM-OCI \cite{wang2013concept}, STEPD \cite{nishida2007detecting}, as well as the recently proposed LFR \cite{wang2015concept}.
The parameters used in these methods were taken as recommended by their authors: the warning and detection thresholds of DDM (EDDM) are $\alpha=3$ ($\alpha=0.95$) and $\beta=2$ ($\beta=0.90$) respectively; the warning and detection significance levels of LFR (STEPD) are $\delta_{\star}=0.01$ ($w=0.05$) and $\epsilon_{\star}=0.0001$ ($d=0.01$) respectively; whereas the parameters of DDM-OCI vary across different data under testing. For our proposed HLFR, the significant rate $\eta$ in Layer-II test is set to $0.05$, and $P=1000$ permutations were used throughout this paper.

%In the following drift detection tasks,

Four benchmark data streams are selected for evaluation, namely ``SEA" \cite{gama2004learning}, ``Checkerboard" \cite{elwell2011incremental}, ``Rotating hyperplane", and USENET1 \cite{katakis2008ensemble}. These datasets include both synthetic and real-world data. A comprehensive description to these datasets is introduced in~\cite{yu2017concept}. Drifts are synthesized in the data, thus controlling ground truth concept drift locations and enabling precise quantitative analysis. Table \ref{Tab:summary_dataset} summarized drift types and the data properties for each stream. Obviously, the selected datasets span the gamut of concept drift types.

\begin{table}[!hbpt]
\small
\setlength{\tabcolsep}{2pt}
\begin{center}
\caption{Summary of properties of selected datasets}\label{Tab:summary_dataset}
\begin{tabular}{cccccc}\hline
Data property & SEA& Checkerboard & Hyperplane & USENST1 \\ \hline
gradual &  $no$ & $no$ &  $yes$ & $no$ \\
abrupt &  $yes$ & $yes$ &  $yes$ & $yes$ \\
recurrent &  $yes$  & $no$ & $yes$ & $yes$ \\
imbalance &  $no$ & $no$ & $yes$ & $no$ \\
high dimensional &  $no$ & $no$ &  $no$ & $yes$ \\ \hline
\end{tabular}
\end{center}
\end{table}

\begin{figure*}[!htbp]
%\centering
\begin{tabular}{ccc}
\subfigure[SEA dataset] {\includegraphics[width=.47\textwidth]{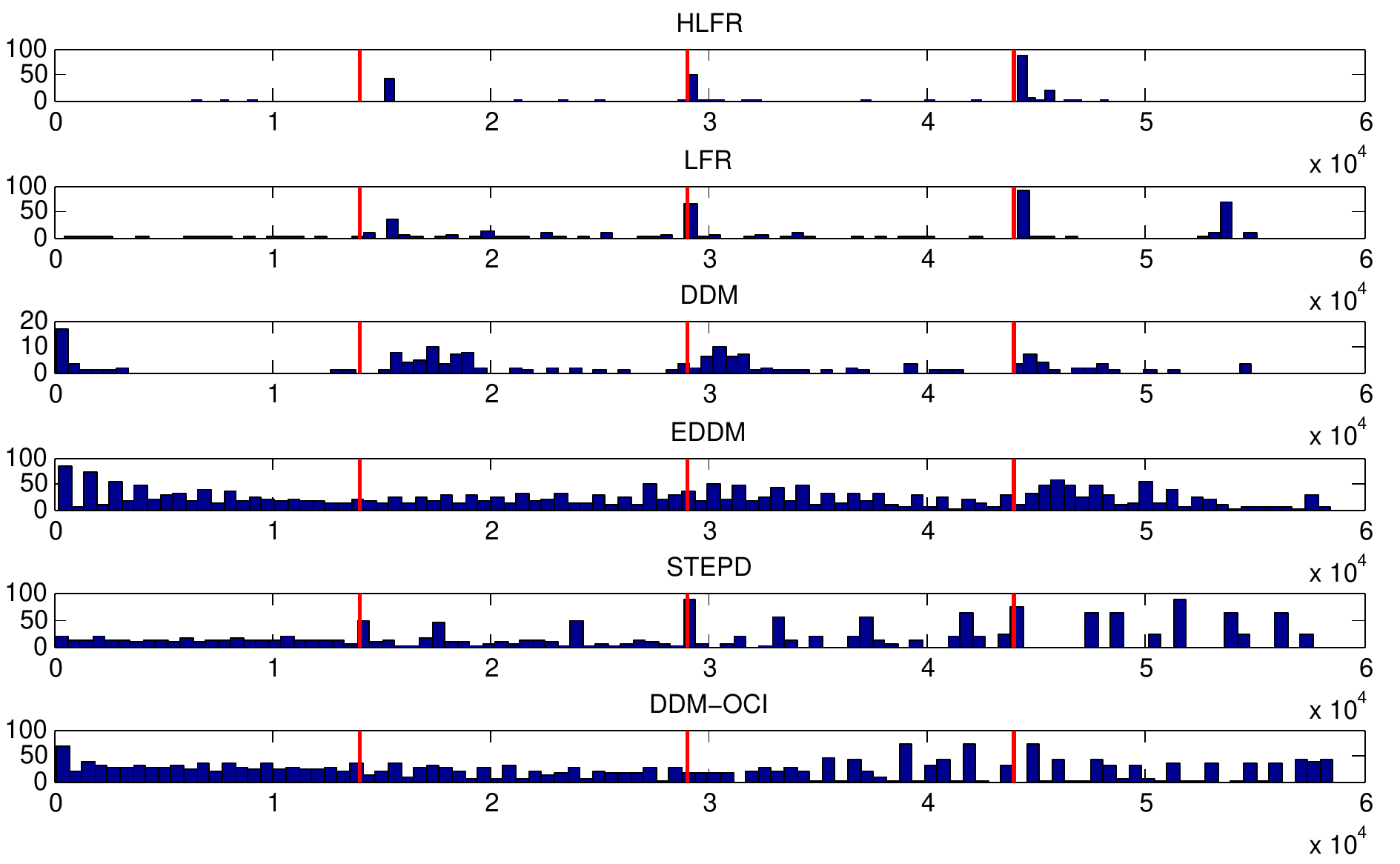}}
\subfigure[Checkerboard dataset] {\includegraphics[width=.47\textwidth]{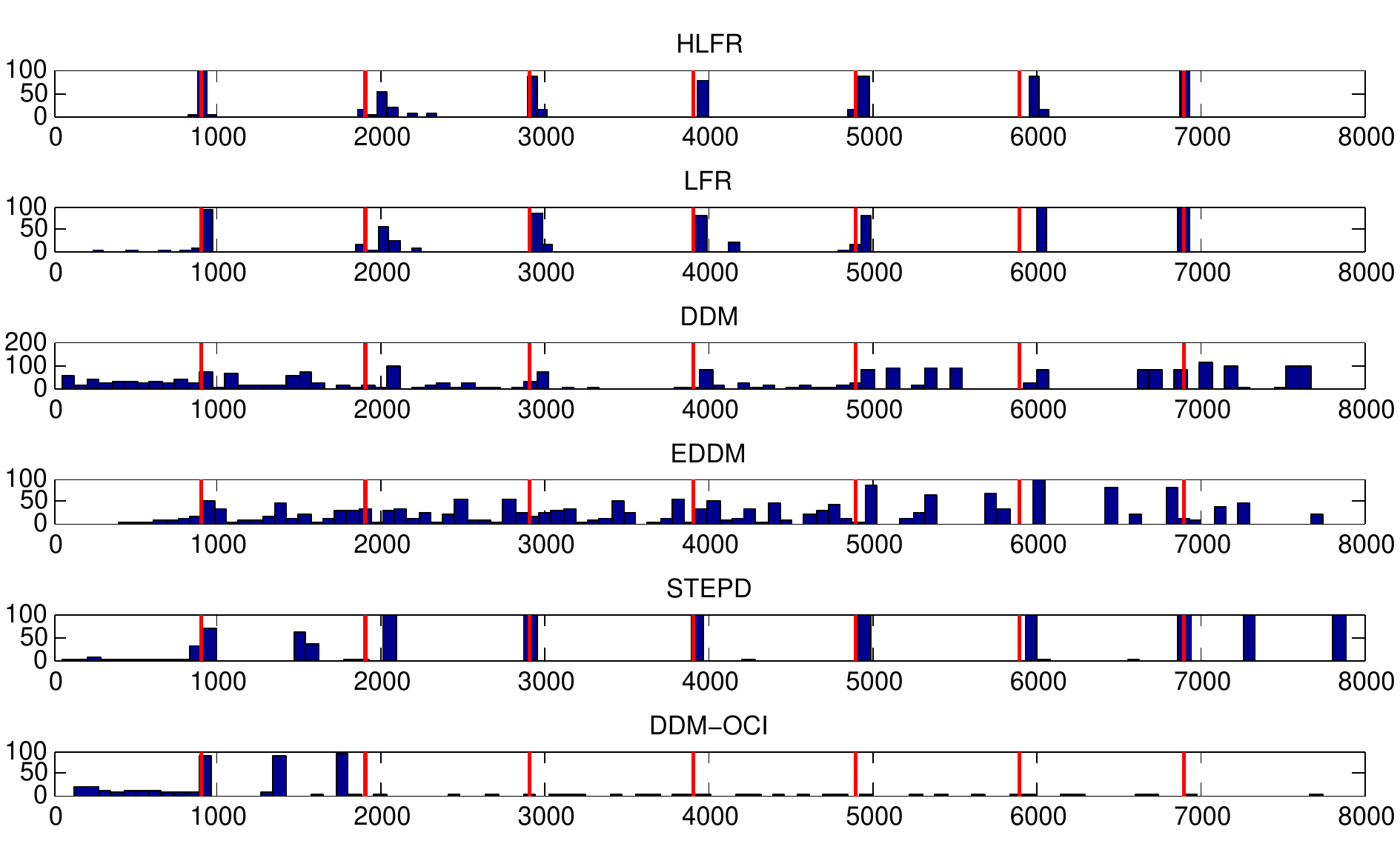}} & \\
\subfigure[Rotating hyperplane dataset] {\includegraphics[width=.47\textwidth]{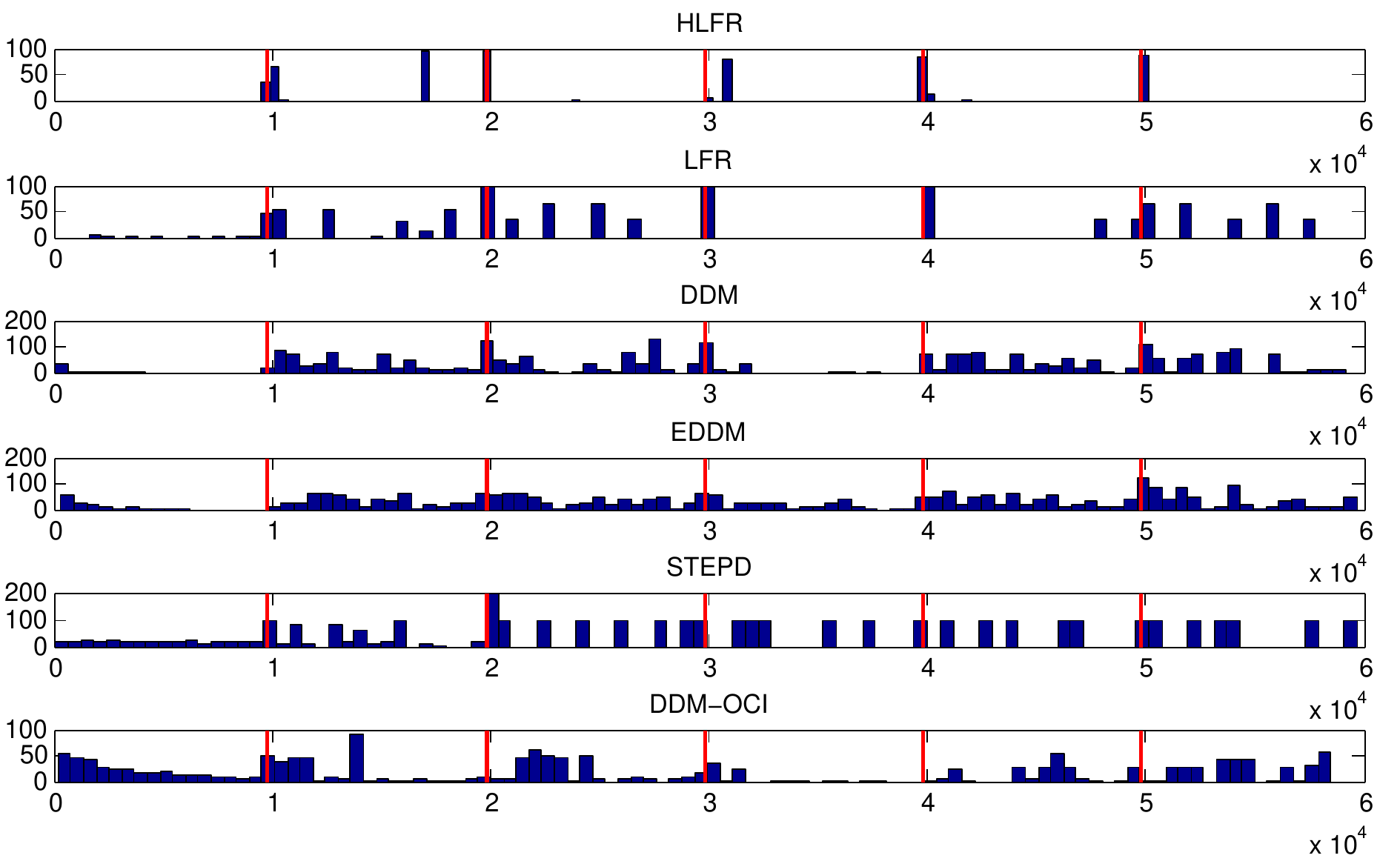}}
\subfigure[USENET1 dataset] {\includegraphics[width=.47\textwidth]{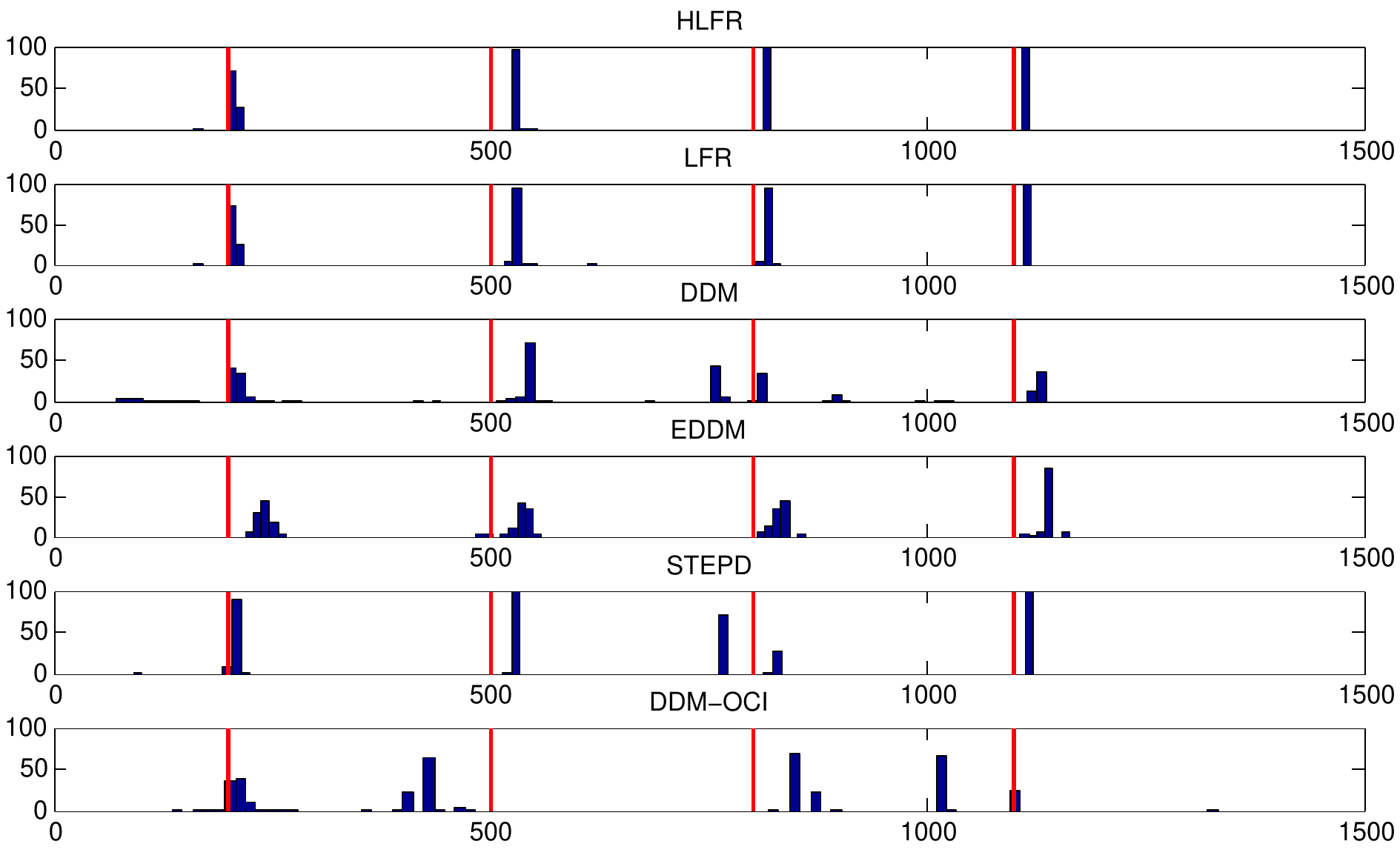}}\\
\end{tabular}
\caption{The histograms of detected concept drift points, generated using different methods, over (a) SEA; (b) Checkerboard; (c) Rotating hyperplane and (d) USENET1 datasets. In each row, the red bars denote the ground truth locations of concept drift points, whereas the blue bars are the histogram of detected points summarized over $100$ independent trails.\vspace{-0.2cm}}
\label{fig:detection_comparison}
\end{figure*}

Each stream was generated and tested independently for $100$ times.
The base classifier used for all competing methods in all streams is a (soft margin) linear SVM with regularization parameter $C=1$. The only exception comes from USENET1, in which a radial basis function (RBF) kernel SVM with kernel width $1$ is selected.
Fig. \ref{fig:detection_comparison} demonstrates the detection results of different methods averaged over these $100$ trails. As can be seen, HLFR and LFR significantly outperform their competitors in terms of promptly detecting concept drifts with fewer missed or false detections, regardless of drift types or data properties. By integrating the Layer-II test, HLFR further improves on LFR by effectively reducing even the few false positives triggered by LFR.

\begin{figure*}[!htbp]
\centering
\begin{tabular}{ccc}
\subfigure[Precision over SEA dataset] {\includegraphics[width=.23\textwidth]{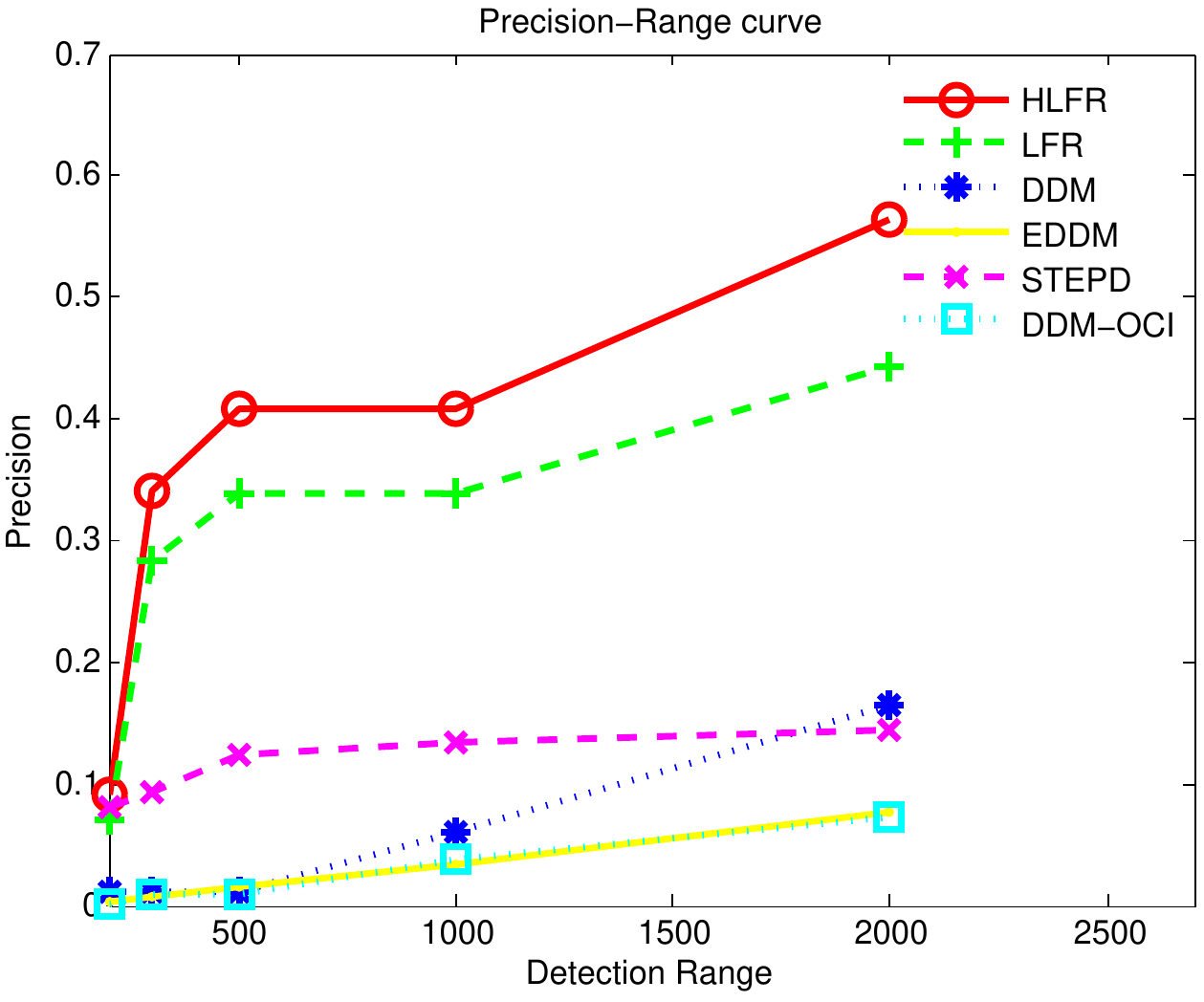}}
\subfigure[Precision over Checkerboard dataset] {\includegraphics[width=.23\textwidth]{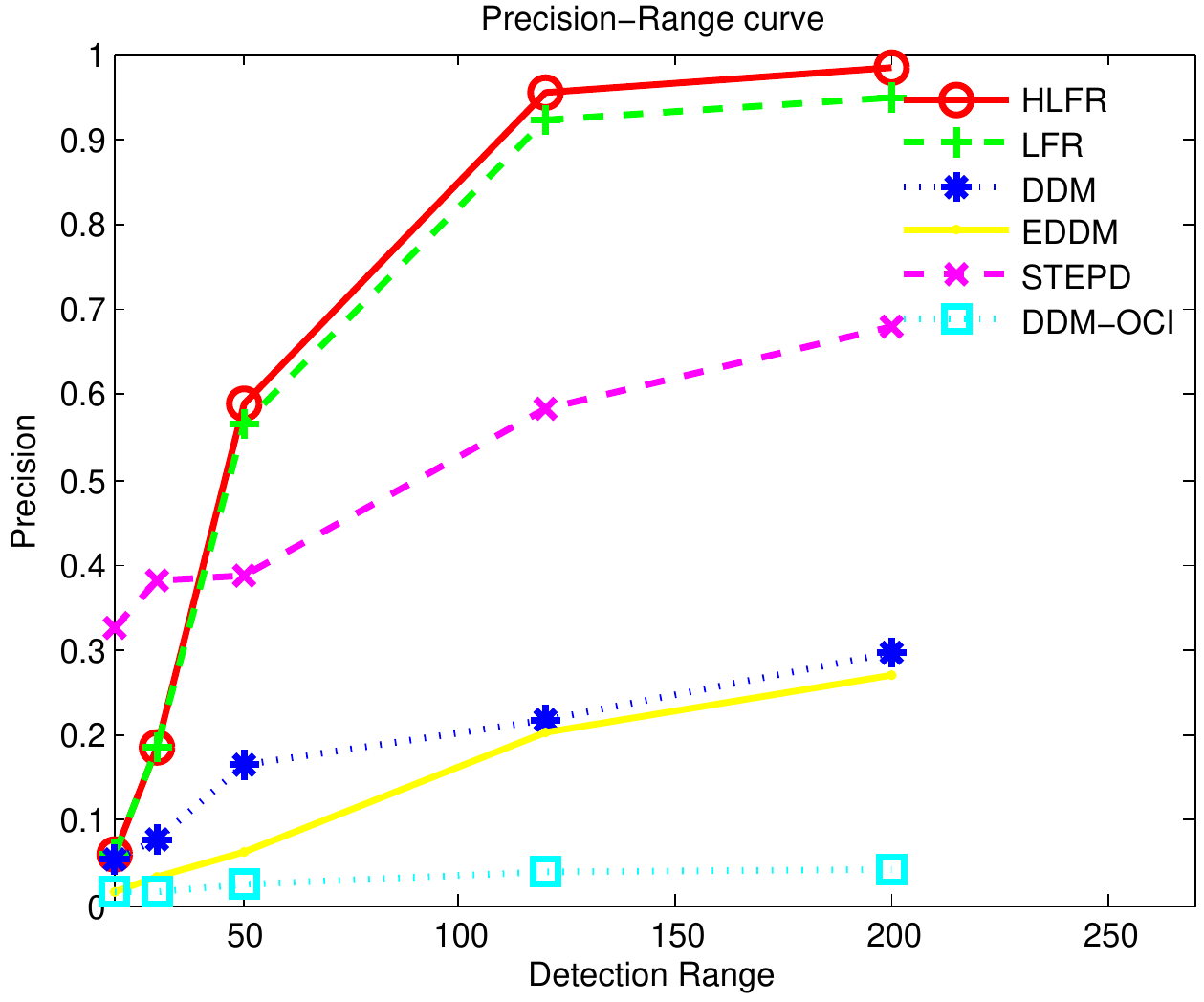}}
\subfigure[Precision over Hyperplane dataset] {\includegraphics[width=.23\textwidth]{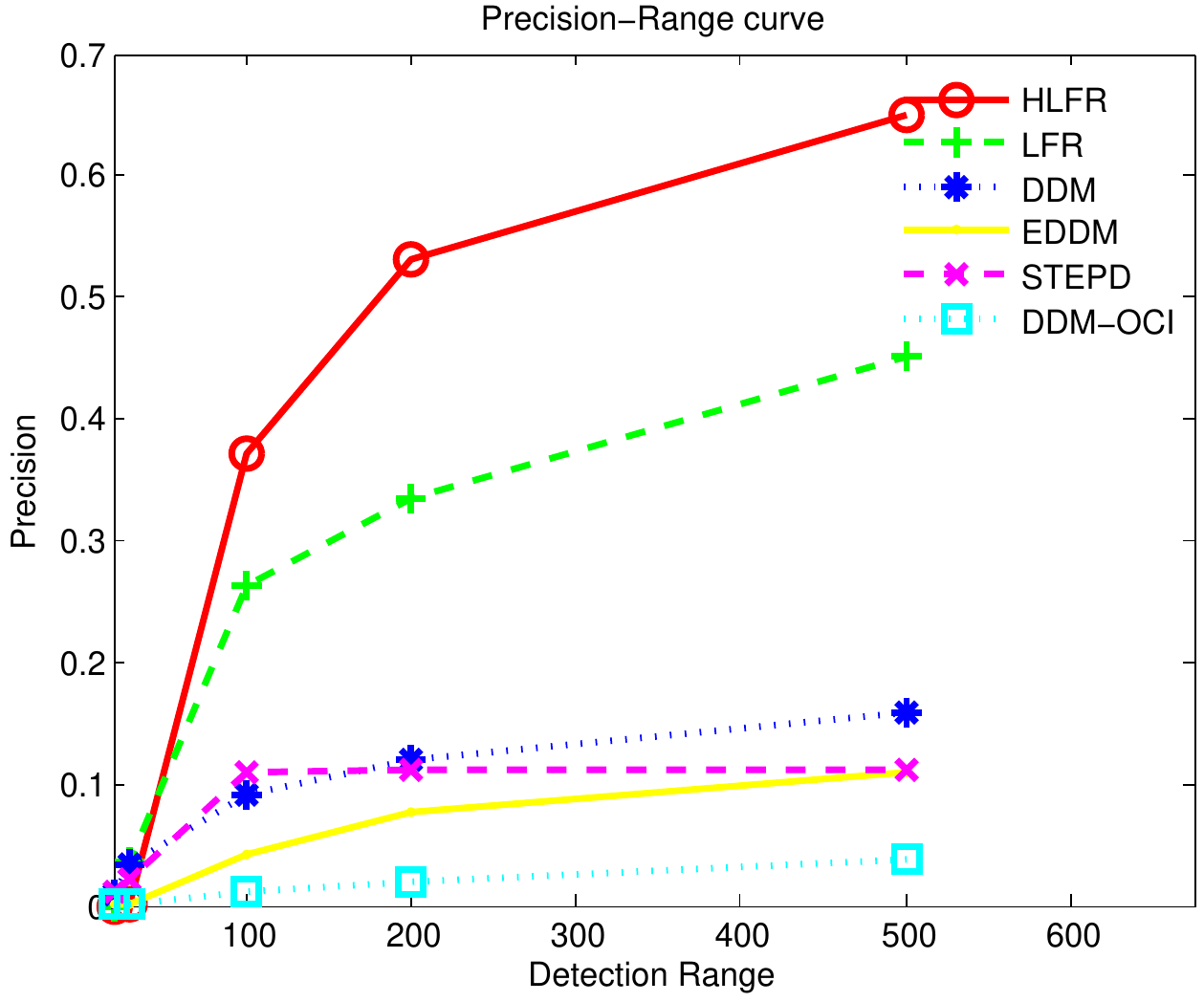}}
\subfigure[Precision over USENET1 dataset] {\includegraphics[width=.23\textwidth]{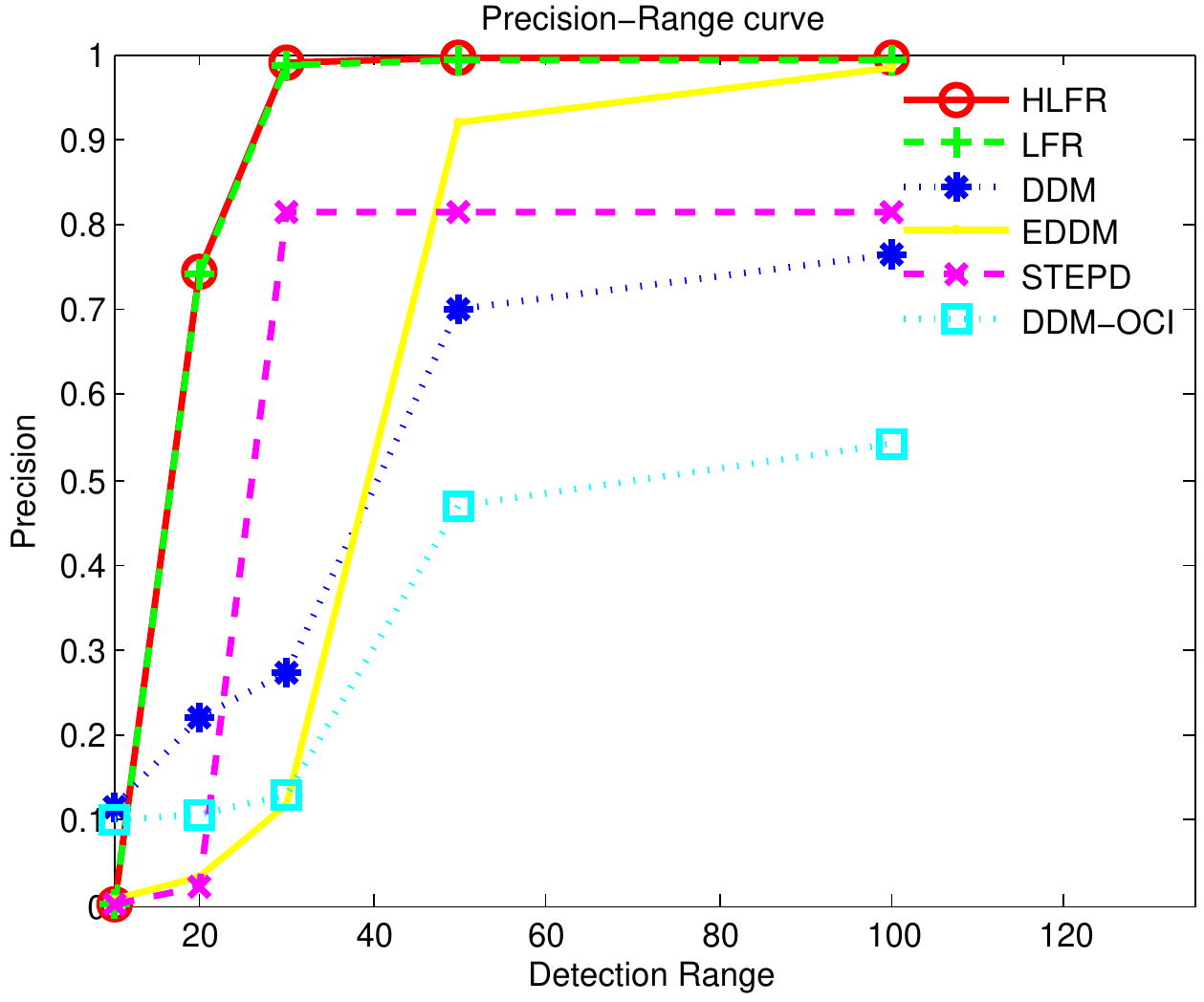}} & \\
\subfigure[Recall over SEA dataset] {\includegraphics[width=.23\textwidth]{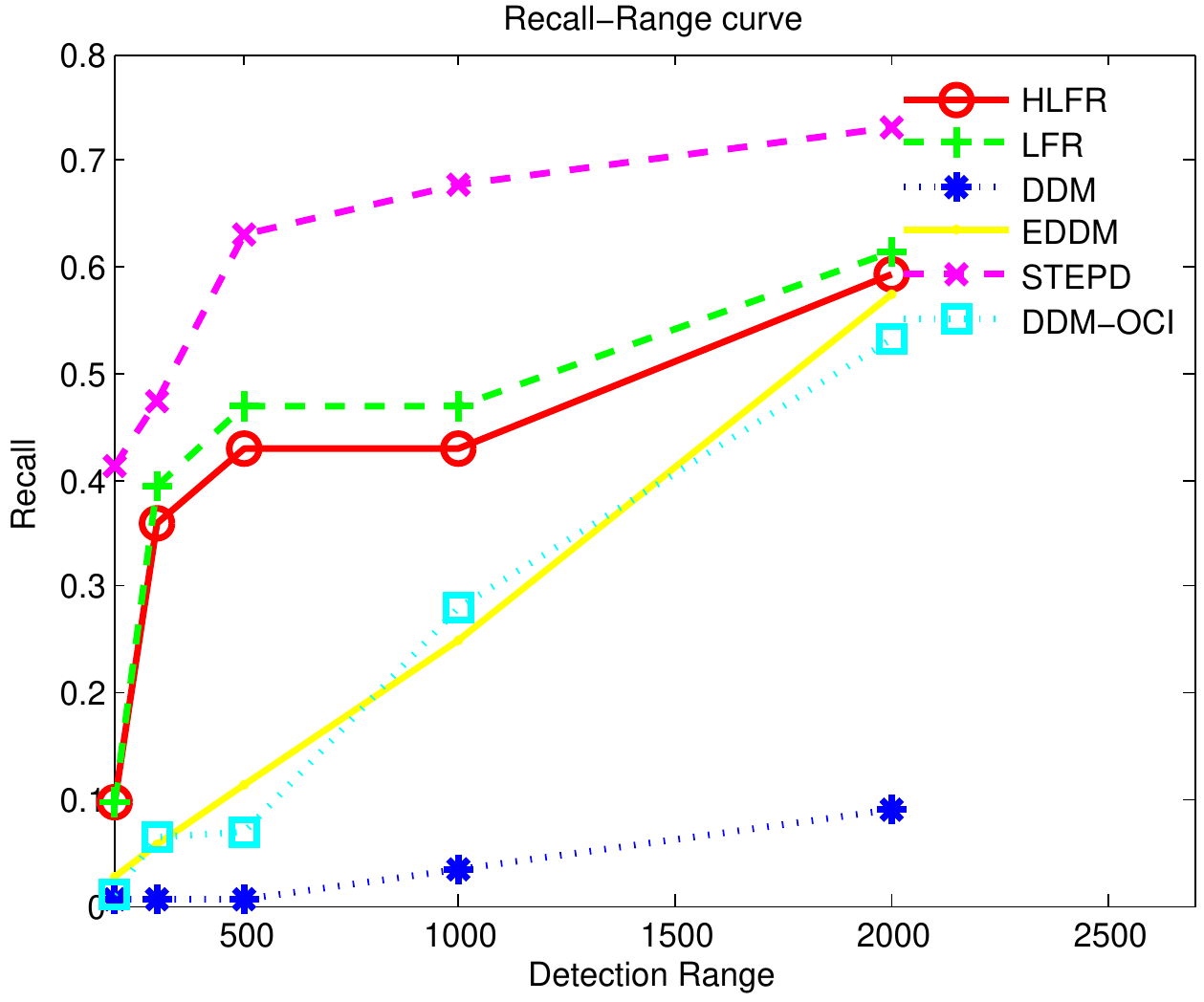}}
\subfigure[Recall over Checkerboard dataset] {\includegraphics[width=.23\textwidth]{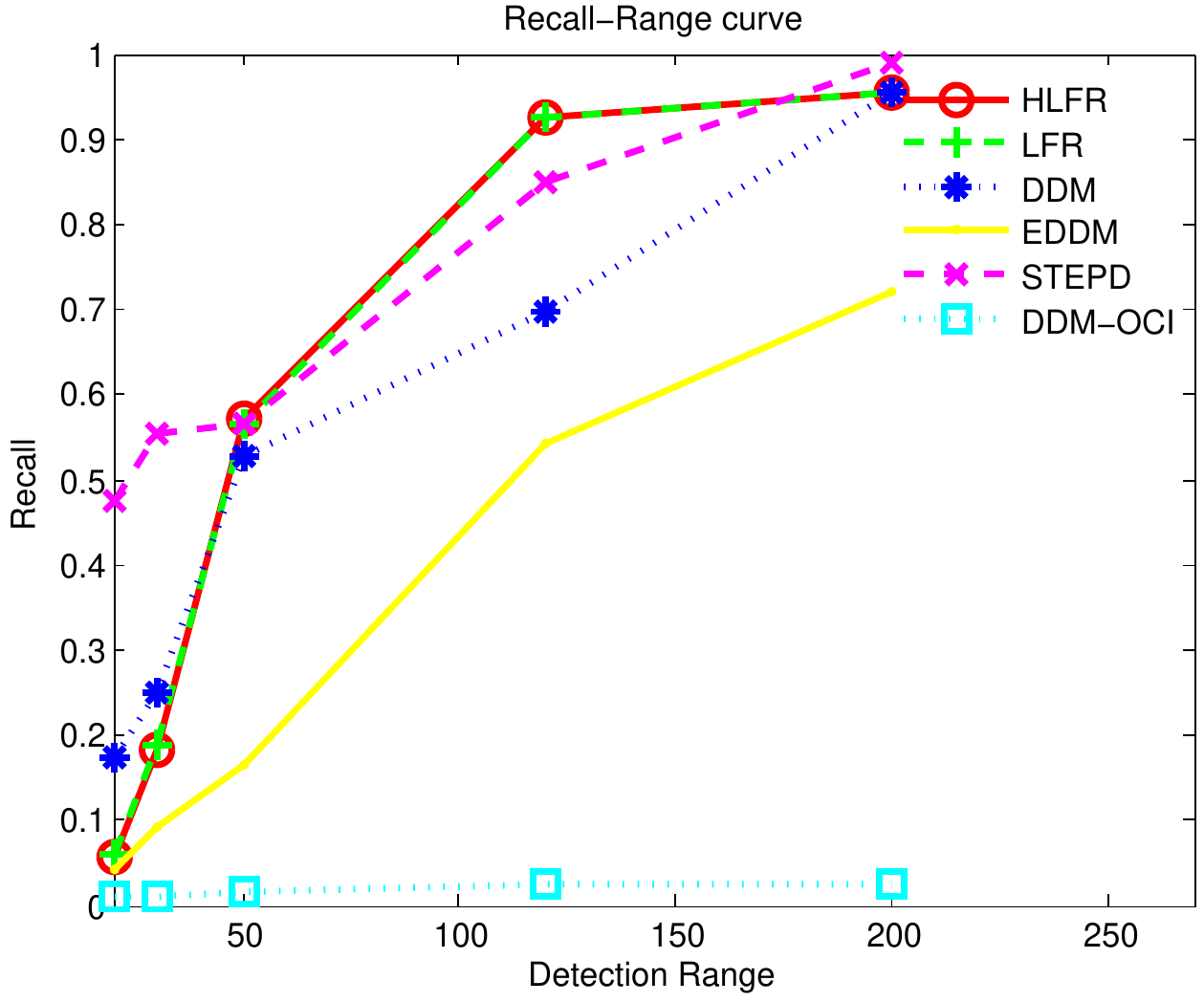}}
\subfigure[Recall over Hyperplane dataset] {\includegraphics[width=.23\textwidth]{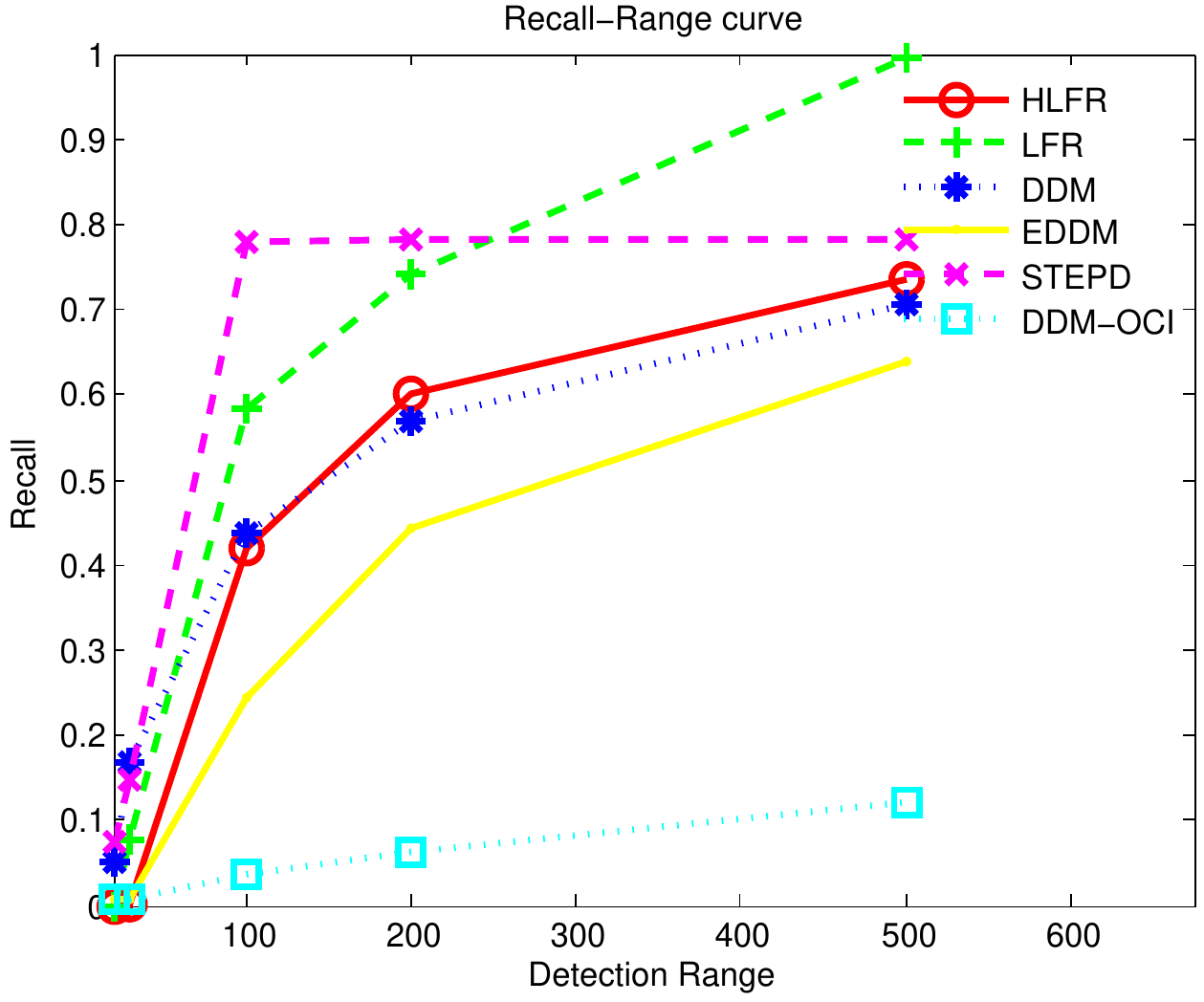}}
\subfigure[Recall over USENET1 dataset] {\includegraphics[width=.23\textwidth]{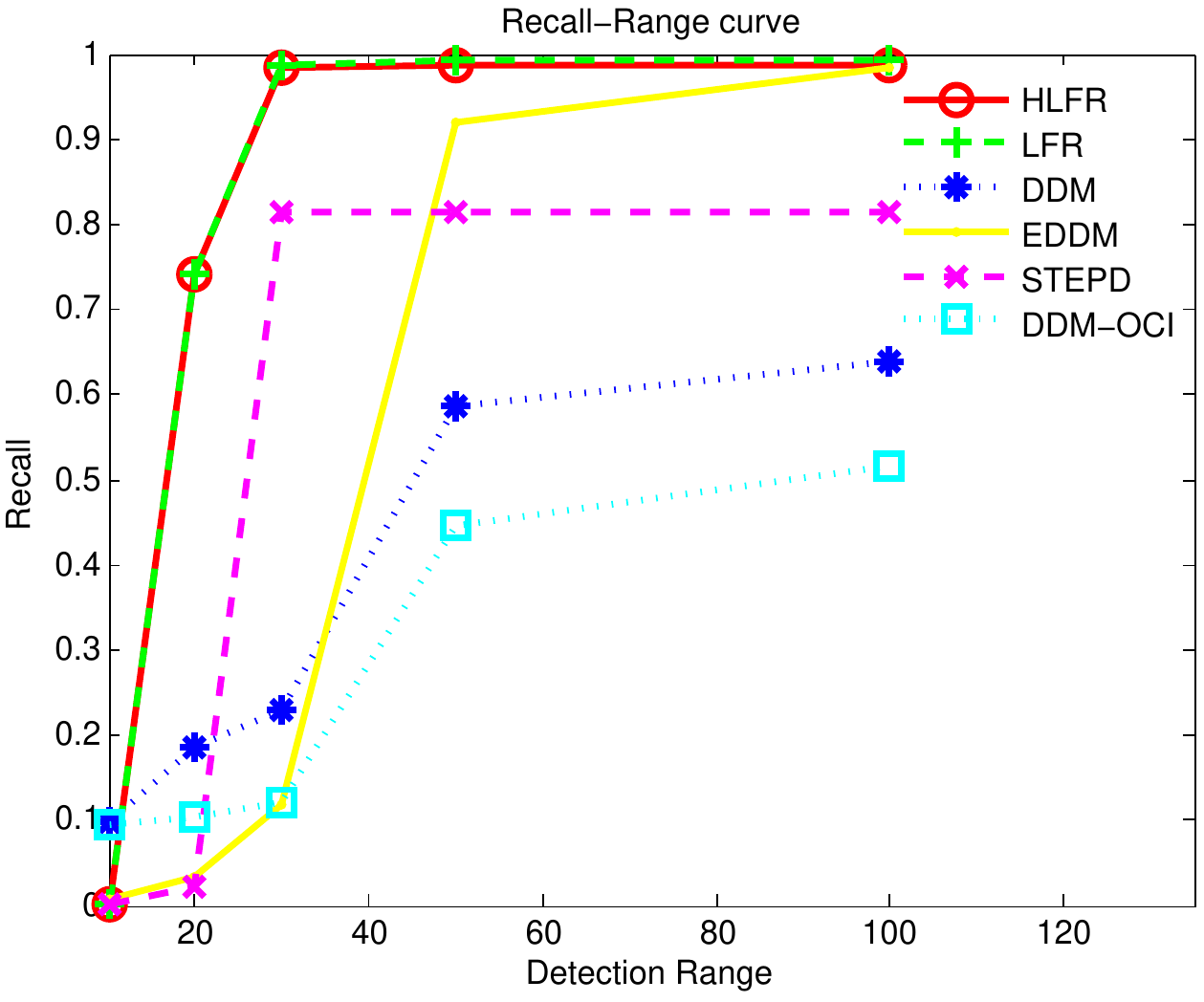}}\\
\end{tabular}
\caption{The $\mathbf{Precision}$ and $\mathbf{Recall}$ values of different methods over SEA, Checkerboard, Rotating hyperplane and USENET1 datasets. In each figure, the X-axis represents the predefined (largest allowable) detection delay, and the Y-axis denotes the corresponding metric values. For a specific delay range, a higher $\mathbf{Precision}$ or $\mathbf{Recall}$ value suggests better performance.\vspace{-0.5cm}}
\label{fig:precision_recall_comparison}
\end{figure*}

Quantitative comparison are performed as well. We define a True Positive ($TP$) as a detection within a fixed delay range after a concept drift occurred, a False Negative ($FN$) as missing a detection within the delay range, and a False Positive ($FP$) as a detection outside this range or an extra detection in the range. For each detector, its detection quality is then evaluated by the $\mathbf{Recall}=TP/(TP+FN)$ and the $\mathbf{Precision}=TP/(TP+FP)$ values.
The $\mathbf{Precision}$ and $\mathbf{Recall}$ values with respect to a predefined (largest allowable) detection delay are demonstrated in Fig. \ref{fig:precision_recall_comparison}.
At the first glance, HLFR, LFR and STEPD can always achieve higher $\mathbf{Precision}$ or $\mathbf{Recall}$ values across different ranges. If we look deeper, the $\mathbf{Precision}$ values is significantly improved with HLFR while the $\mathbf{Recall}$ values of HLFR and LFR are similar (except for Rotating hyperplane dataset).
This result corroborates our Type-I and Type-II error analysis in section~\ref{sec:error_analysis}: Layer-II test aims to confirm or deny the validity of layer-I detection results, thus it cannot compensate for the errors of missing a detection made by Layer-I test. In other words, the Type-I error of HLFR should be smaller than that of LFR theoretically, whereas the Type-II error of HLFR is lower bounded by LFR.
In fact, the relatively lower $\mathbf{Recall}$ of HLFR (compared to LFR) suggests that the used Layer-II test is a little conservative, i.e., it has a small probability to reject true positive detection triggered by Layer-I test (i.e., LFR). On the other hand, it seems that STEPD has much higher $\mathbf{Recall}$ values on SEA and Rotating hyperplane datasets. However, the result is meaningless. This is because STEPD triggers significantly more false alarms (as seen in the fifth row of Fig. \ref{fig:detection_comparison}(a) and Fig. \ref{fig:detection_comparison}(c)), such that its $\mathbf{Precision}$ values on these two datasets are consistently smaller than $0.15$. Table \ref{Tab:detection_delay} summarized the detection delays (ensemble average) for all competing algorithms. Out of the four datasets, our HLFR has the shortest (average) detection delay in three of them.

%The results of quantitative evaluations corroborate the qualitative observations.

\begin{table}[!hbpt]
\small
\setlength{\tabcolsep}{2pt}
\begin{center}
\caption{Average detection delay for all competing algorithms. The best performance in each dataset is highlighted in bold.}\label{Tab:detection_delay}
\begin{tabular}{cccccc}\hline
Algorithms & SEA & Checkerboard & Hyperplane & USENST1 \\ \hline
STEPD &  463 & 57 &  140 & 19 \\
DDM-OCI &  844 & 58 &  198 & 26 \\
EDDM &  939 & 93 &  166 & 36 \\
DDM &  1209 & 69 &  125 & 26 \\
LFR &  \textbf{458} & 56 & 127 & 17 \\
HLFR &  482  & \textbf{55} & \textbf{120} & \textbf{17} \\\hline
\end{tabular}
\end{center}
\end{table}

\subsection{Concept Drift Adaptation with A-HLFR} \label{section5.4}
In this section, we perform two case studies using representative real-world concept drift datasets from email filtering and weather prediction domain respectively, aiming to validate the rationale of HLFR on concept drift detection as well as the potency of A-HLFR on concept drift adaptation. Performance is compared to DDM, EDDM, STEPD as well as LFR. Note that, the results of DDM-OCI are omitted as it is hard to detect ``reasonable" concept drift points in the selected data. \phantom{\ref{section_adaptive}.}

The spam filtering dataset \cite{katakis2010tracking}, consisting of $9324$ instances and $500$ attributes, is used herein. This data represents email messages from the Spam Assassin Collection\footnote{http://spamassassin.apache.org/} and contains contains natural concept drifts \cite{katakis2010tracking,katakis2006dynamic}. The spam ratio is approximately $20\%$. Besides, the weather dataset \cite{ditzler2013incremental,elwell2011incremental}, a subset of the National Oceanic and Atmospheric Administration (NOAA) data\footnote{ftp://ftp.ncdc.noaa.gov/pub/data/gsod}, consisting of daily observations recorded in Offutt Air Force Base in Bellevue, Nebraska, is also used in the study. This data is collected and recorded over $50$ years, containing not only short-term seasonal changes, but also (possibly) long-term climate trend. Daily measurements include temperature, pressure, wind speed, visibility, and a variety of features. The task is to predict whether it is going to rain from these features. Minority class cardinality varied between $10\%$ and $30\%$ throughout these $50$ years.

%The Electricity Pricing data set (elec2) was first introduced in \cite{harries1999splice} and has become a benchmarking for concept drift adaptation \cite{zliobaite2013good,bifet2013pitfalls,vzliobaite2015evaluation}. This data set provides time and demand fluctuations in the price of electricity in New South Wales, Australia. We use the day, period, New South Wales (NSW) electricity demand, VIC (Victoria) electricity demand and the scheduled electricity transfer as features. The task is to predict whether the NSW price will be higher or lower than VICs in a $24$-hour period. Instances with missing features were removed. Besides, same to \cite{ditzler2013incremental}, since elec2 dataset does not contain class imbalance, one of the classes was under sampled to create an imbalance ratio of approximately $1:10$ (minority data are $9.1$ percent of total data size). Thus, there are total three different streaming data under exploration. A soft-margin SVM classifier is generated from the training dataset and evaluated (also adapted) on the test dataset.

~\\
\noindent
\textbf{On Parameter Tuning and Experimental Setting.} A common phenomenon for classification of real-world streaming data with concept drifts and temporal dependency is that ``\emph{the more random alarms fire the classifier, the better the accuracy} \cite{zliobaite2013good}". Thus, to provide a fair comparison, the parameters of all competing methods are tuned to detect similar number of concept drifts. Table \ref{Tab:spam_parameter} and Table \ref{Tab:weather_parameter} summarized the key parameters regarding significance levels (or thresholds) of different methods in two selected real-world datasets respectively. For spam data, an extensive search for appropriate partition of training and testing sets was performed based on two criteria. First, there is no strong autocorrelations in the classification error sequence on the training set. This is because once the errors are highly autocorrelated, it is very probably that the training data is no longer $i.i.d.$ or the training data spans different concepts. Second, the classifier trained on the training set can achieve promising classification accuracies on both minority and majority classes, i.e., sufficient number of training data is required. With these two considerations, the length of training set is set to $600$. As for the weather data, the training size is set to $120$ instances (days), approximately one season as suggested in \cite{ditzler2013incremental}.

\begin{table}[!hbpt]
\begin{center}
\caption{Parameter settings in spam email filtering.}\label{Tab:spam_parameter}
\begin{tabular}{cccccc}\hline
Algorithms & Parameter settings on significance levels (or thresholds) \\ \hline
STEPD &  $w=0.005$, $d=0.0003$ \\
EDDM &  $\alpha=0.95$, $\beta=0.90$ \\
DDM &  $\alpha=3$, $\beta=2.5$ \\
LFR &  $\delta_\star=0.01$, $\epsilon_\star=0.00001$ \\
HLFR &  $\delta_\star=0.01$, $\epsilon_\star=0.0001$, $\eta=0.01$ \\
A-HLFR &  $\delta_\star=0.01$, $\epsilon_\star=0.0001$, $\eta=0.01$ \\\hline
\end{tabular}
\end{center}
\end{table}

\begin{table}[!hbpt]
\begin{center}
\caption{Parameter settings in weather prediction.}\label{Tab:weather_parameter}
\begin{tabular}{cccccc}\hline
Algorithms & Parameter settings on significance levels (or thresholds) \\ \hline
STEPD &  $w=0.05$, $d=0.003$ \\
EDDM &  $\alpha=0.95$, $\beta=0.90$ \\
DDM &  $\alpha=2$, $\beta=1.5$ \\
LFR &  $\delta_\star=0.01$, $\epsilon_\star=0.0001$ \\
HLFR &  $\delta_\star=0.01$, $\epsilon_\star=0.00001$, $\eta=0.3$ \\
A-HLFR &  $\delta_\star=0.01$, $\epsilon_\star=0.00001$, $\eta=0.3$ \\\hline
\end{tabular}
\end{center}
\end{table}

\begin{figure}
\centering
\begin{tabular}{ccc}
\includegraphics[height=4.5cm,width=9.0cm]{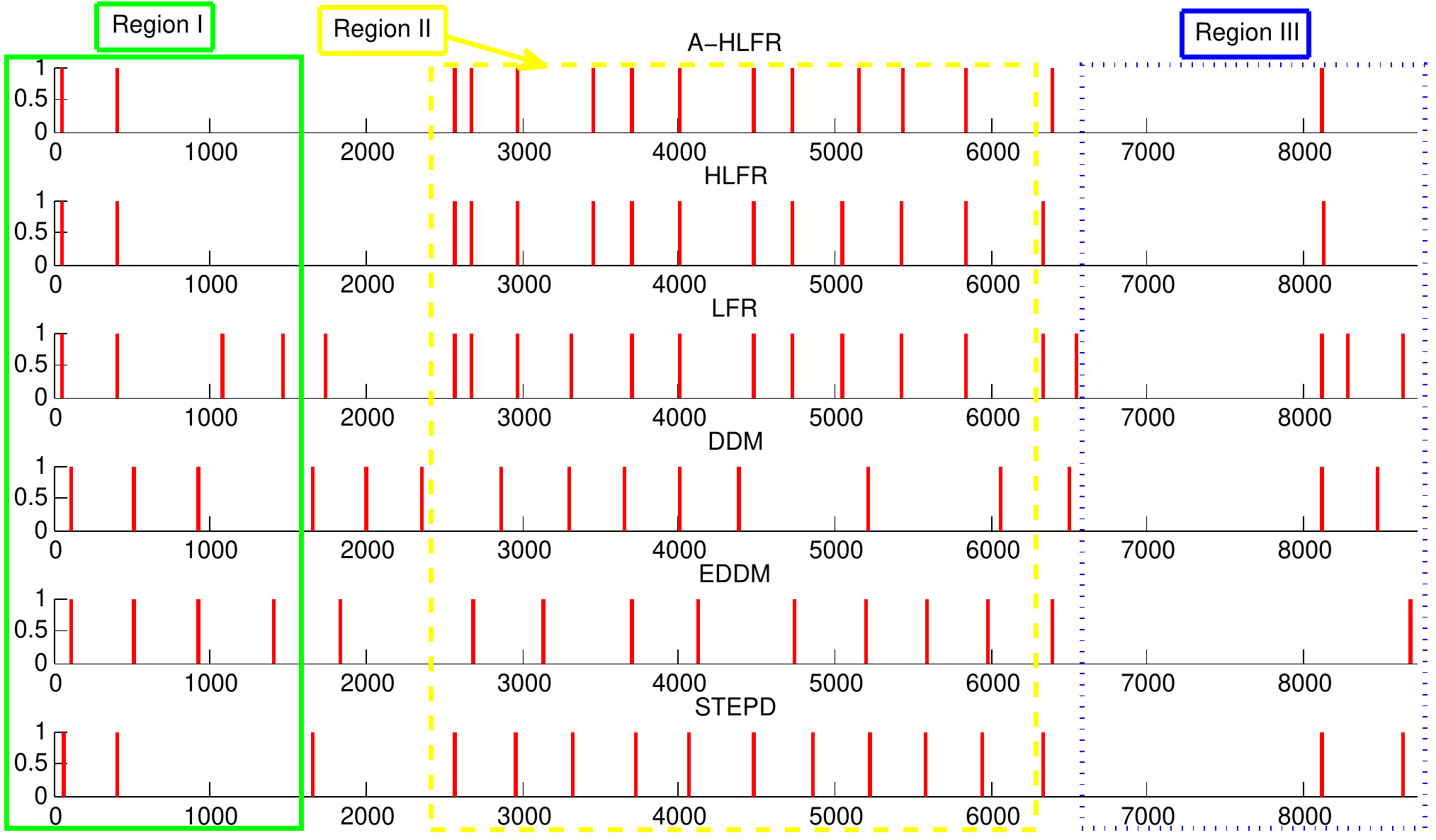} & \\
\end{tabular}
\caption{Concept drift detection results on the spam dataset.\vspace{-0.5cm}}
\label{fig:detection}
\end{figure}

~\\
\noindent
\textbf{Case study on spam dataset.} We first evaluate the performances of different methods on the spam dataset. %Before evaluating detection results, we recommend interested readers to refer to \cite{katakis2010tracking}, in which the expectation maximization (EM) clustering and the $k$-means algorithms have been applied to the conceptual vectors of spam filtering dataset.
%According to the authors of \cite{katakis2010tracking}, there are three dominating concepts distributed in different time periods and these concept drifts occurred approximately in the neighbors of time point $200$ in Region I, time point $8000$ in Region III, the ending location (time point $1800$) of Region I as well as the start and ending locations (time points $2300$ and $6200$) of Region II. Besides, there are many abrupt drifts in Region II. A possible reason for these abrupt and frequent drifts may be batches of outliers or noisy messages.
According to the authors of \cite{katakis2010tracking}, there are three dominating concepts distributed in different time periods and these concept drifts occurred approximately in the neighbors of time instants $200$ and $1800$ in Region I, time instants $2300$ and $6200$ in Region II, and time instant $8000$ in Region III. Besides, there are many abrupt drifts in Region II. A possible reason for these abrupt and frequent drifts may be batches of outliers or noisy messages. According to the concept drift detection results shown in Fig. \ref{fig:detection}, A-HLFR and HLFR best match these descriptions, except that they both miss a potential drift around time instant $1800$. By contrast, although other methods are able to detect this point, they have many other limitations: 1) LFR triggers some false positive detections as well; and 2) DDM or EDDM, not only misses obvious drift points, but also feeds back unconvincing drift locations in Region I or Region III.

\begin{figure}
\centering
\begin{tabular}{ccc}
\subfigure[KPS prequential representation on the spam dataset] {\includegraphics[height=4.5cm,width=9.0cm]{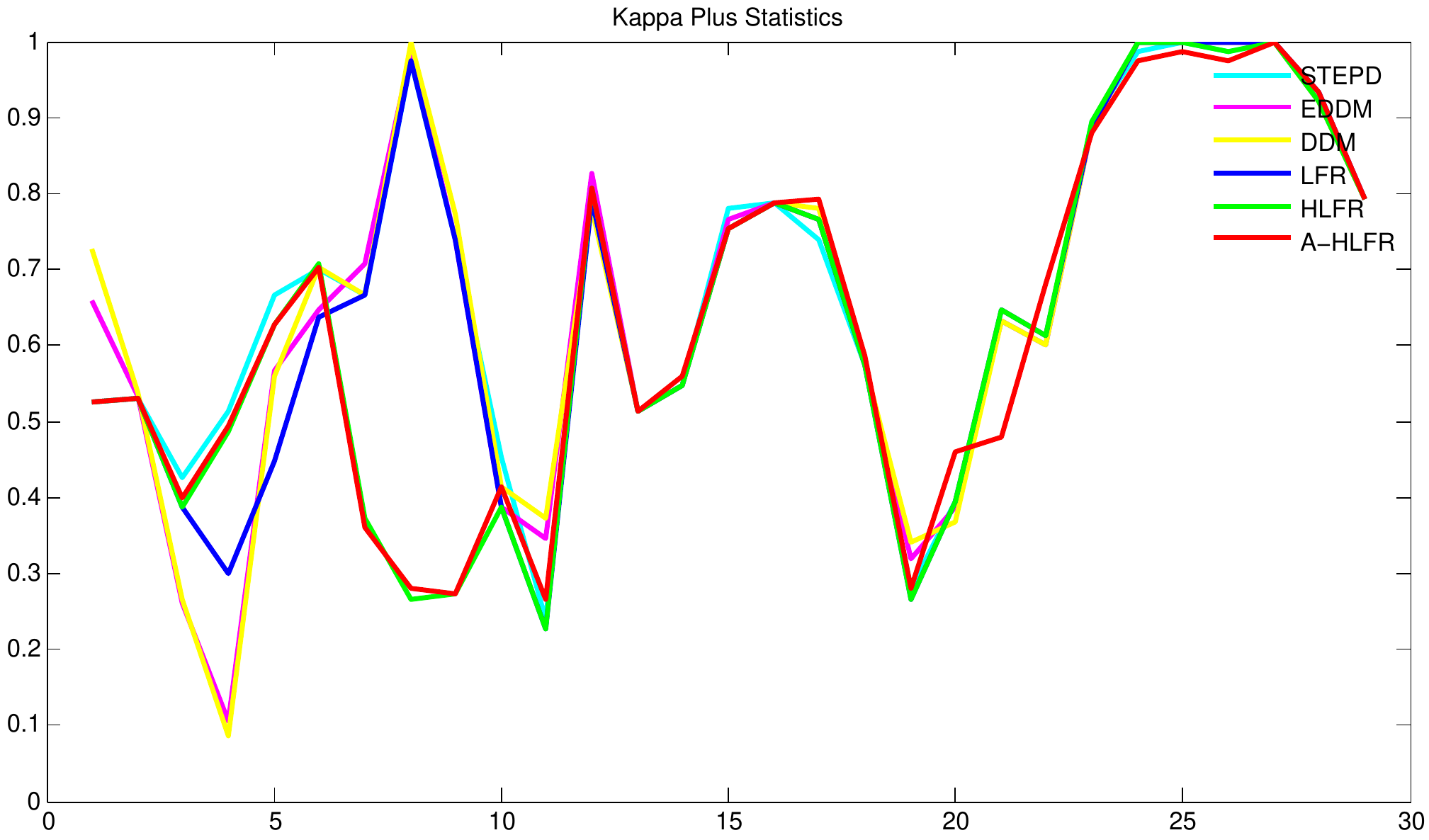}} & \\
\subfigure[KPS prequential representation on the weather dataset] {\includegraphics[height=4.5cm,width=9.0cm]{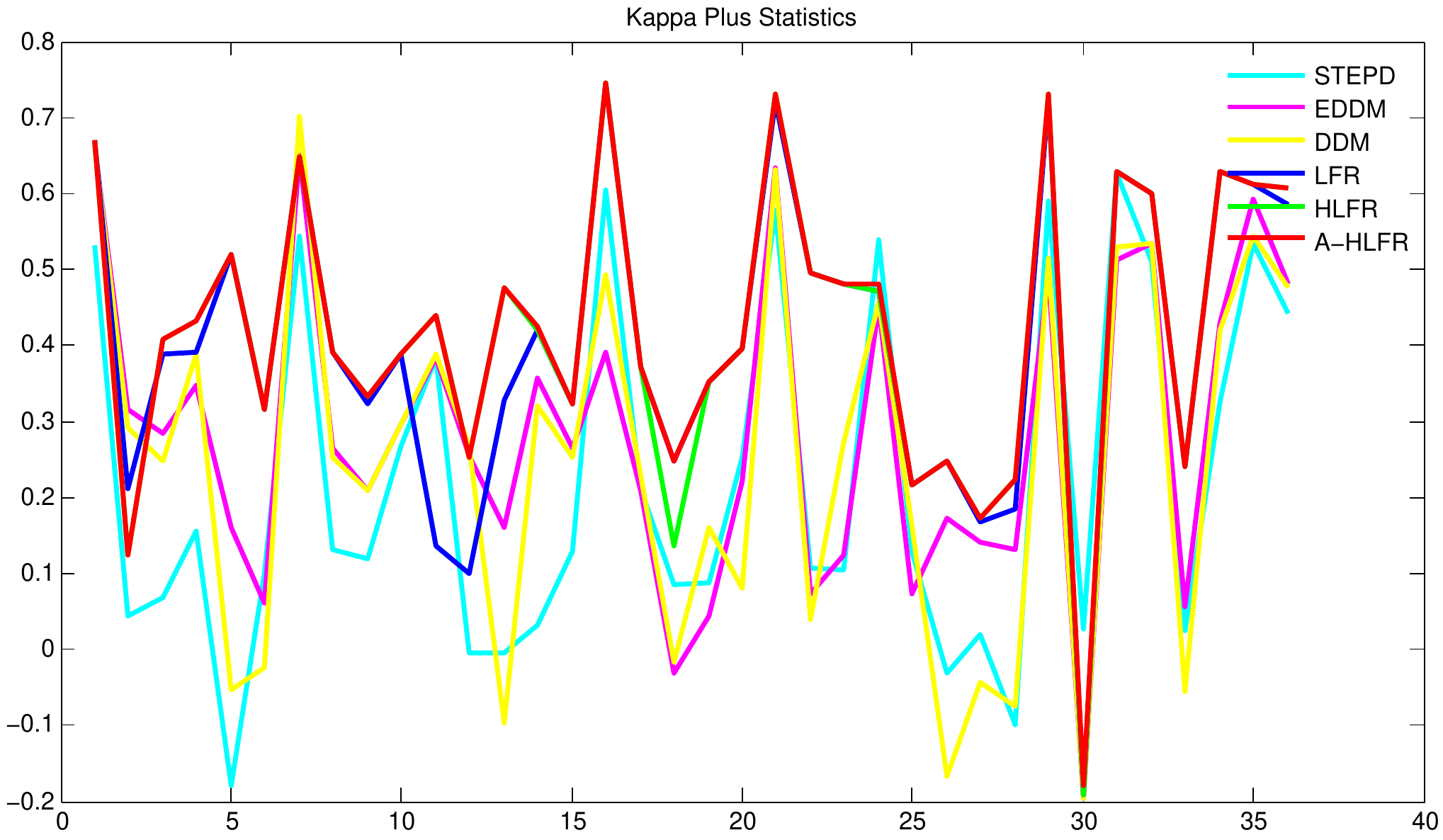}} \\
\end{tabular}
\caption{Kappa Plus Statistic (KPS) prequential representations.\vspace{-0.5cm}}
\label{fig:kappa}
\end{figure}

%To further bridge the connection between our detection results and the descriptions in \cite{katakis2010tracking},
We then applied a recently proposed measurement - Kappa Plus Statistic (KPS) \cite{vzliobaite2015evaluation} - to access experimental results. KPS, defined as $\kappa^{+}=\frac{p_0-p'_e}{1-p'_e}$, aims to evaluate a data stream classifiers performance, taking into account the temporal dependence and effectiveness of classifier adaptation. $p_0$ is the classifier's prequential accuracy \cite{gama2013evaluating} and $p'_e$ is the accuracy of No-Change classifier\footnote{The No-Change classifier is defined as a classifier that predicts the same label as previously observed, $i.e.$, $\hat{y}_t=y_{t-1}$ for any observation $\mathbf{X}_t$ \cite{vzliobaite2015evaluation}.}. We partition the training set into approximately $30$ consecutive time periods. The KPS prequential representation over these periods is shown in Fig. \ref{fig:kappa}(a). As can be seen, the HLFR and A-HLFR adaptations are most effective in periods $1$-$5$, but suffer from a sudden drop in periods $6$-$10$. These observations corroborate the detection results shown in Fig.~\ref{fig:detection}: HLFR and A-HLFR can accurately detect the first drift point without any false positives in Region I, but they both missed a target in Region II. On the other hand, there is almost no performance difference between the classifier update in A-HLFR and HLFR.
%This is perhaps because the temporal relatedness between consecutive concepts is weak or the concept changes gradually (or slowly) such that naively transferring previous knowledge to current domain cannot prompt the classification performance.}

\begin{figure*}[!htbp]
\centering
\begin{tabular}{ccc}
\subfigure[] {\includegraphics[height=3.5cm,width=.45\textwidth]{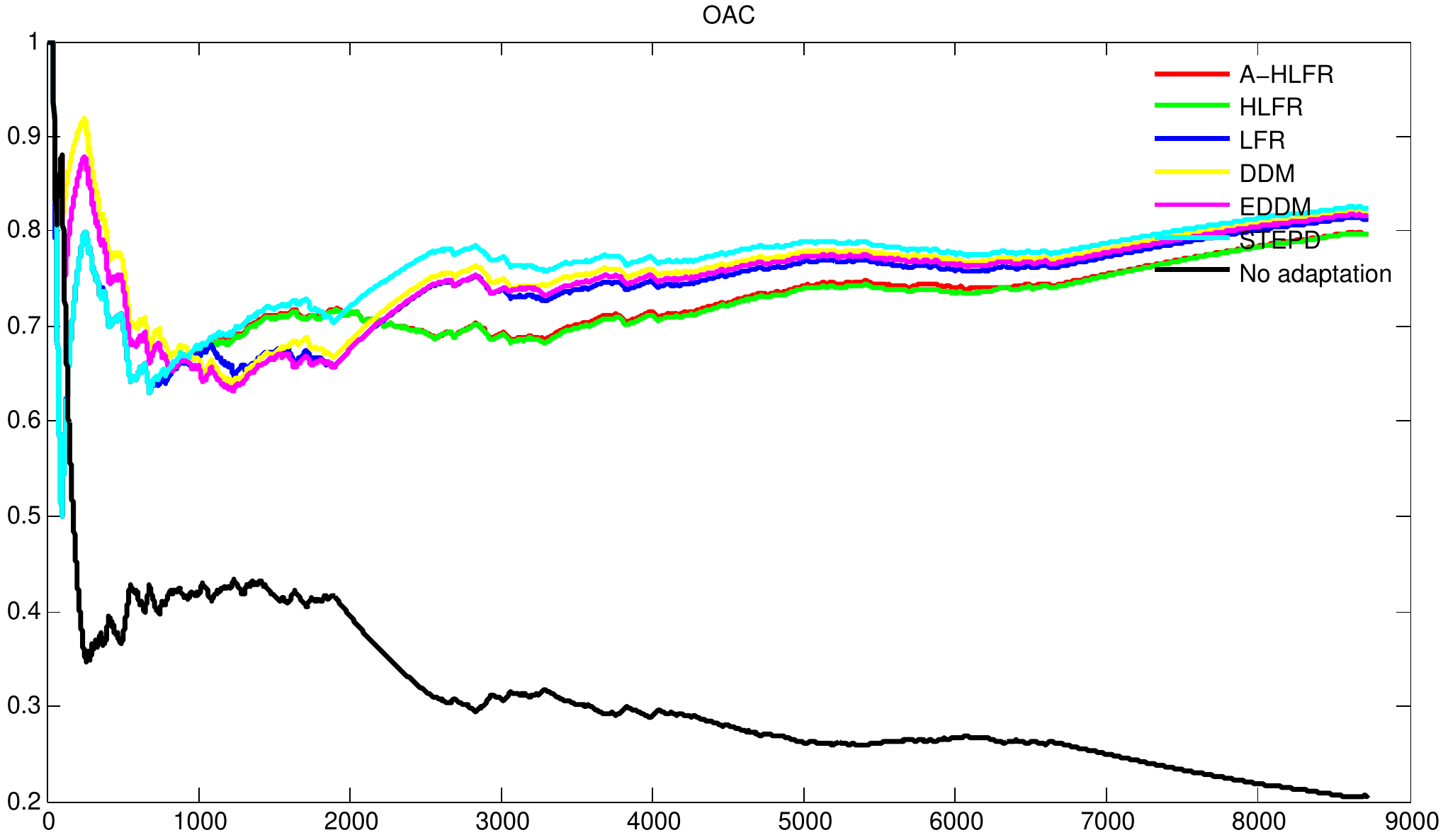}}
\subfigure[] {\includegraphics[height=3.5cm,width=.45\textwidth]{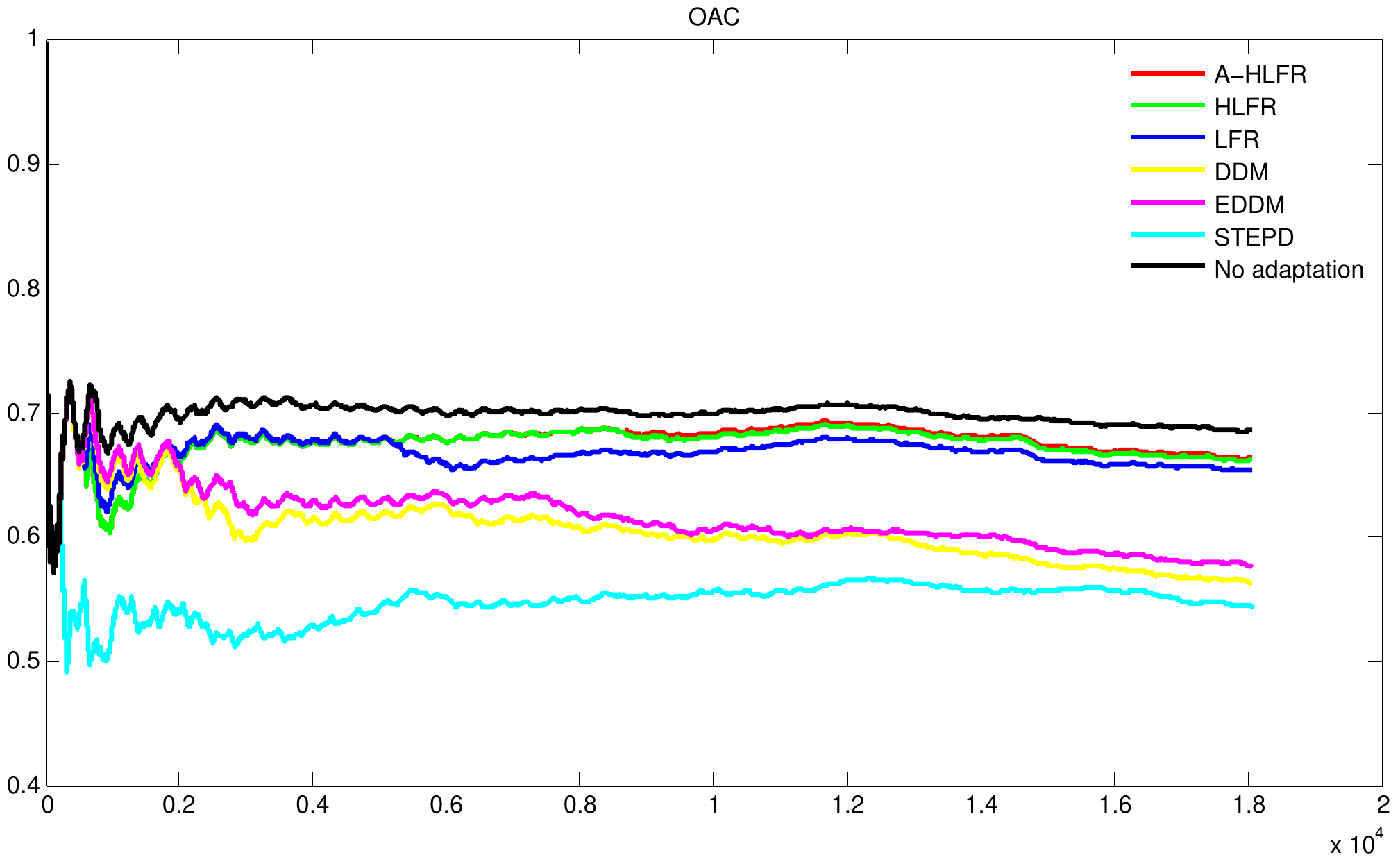}}\\
\subfigure[] {\includegraphics[height=3.5cm,width=.45\textwidth]{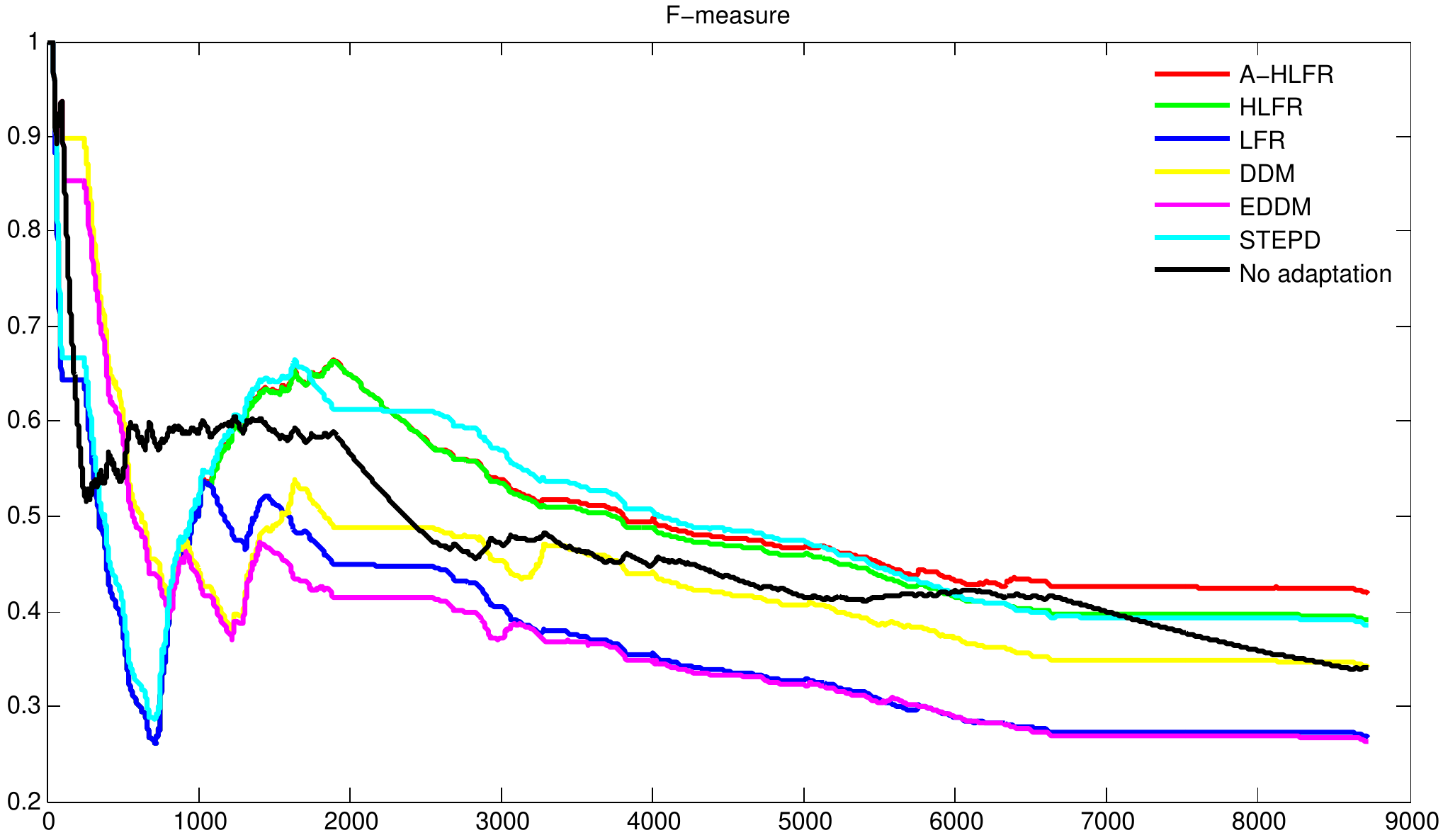}}
\subfigure[] {\includegraphics[height=3.5cm,width=.45\textwidth]{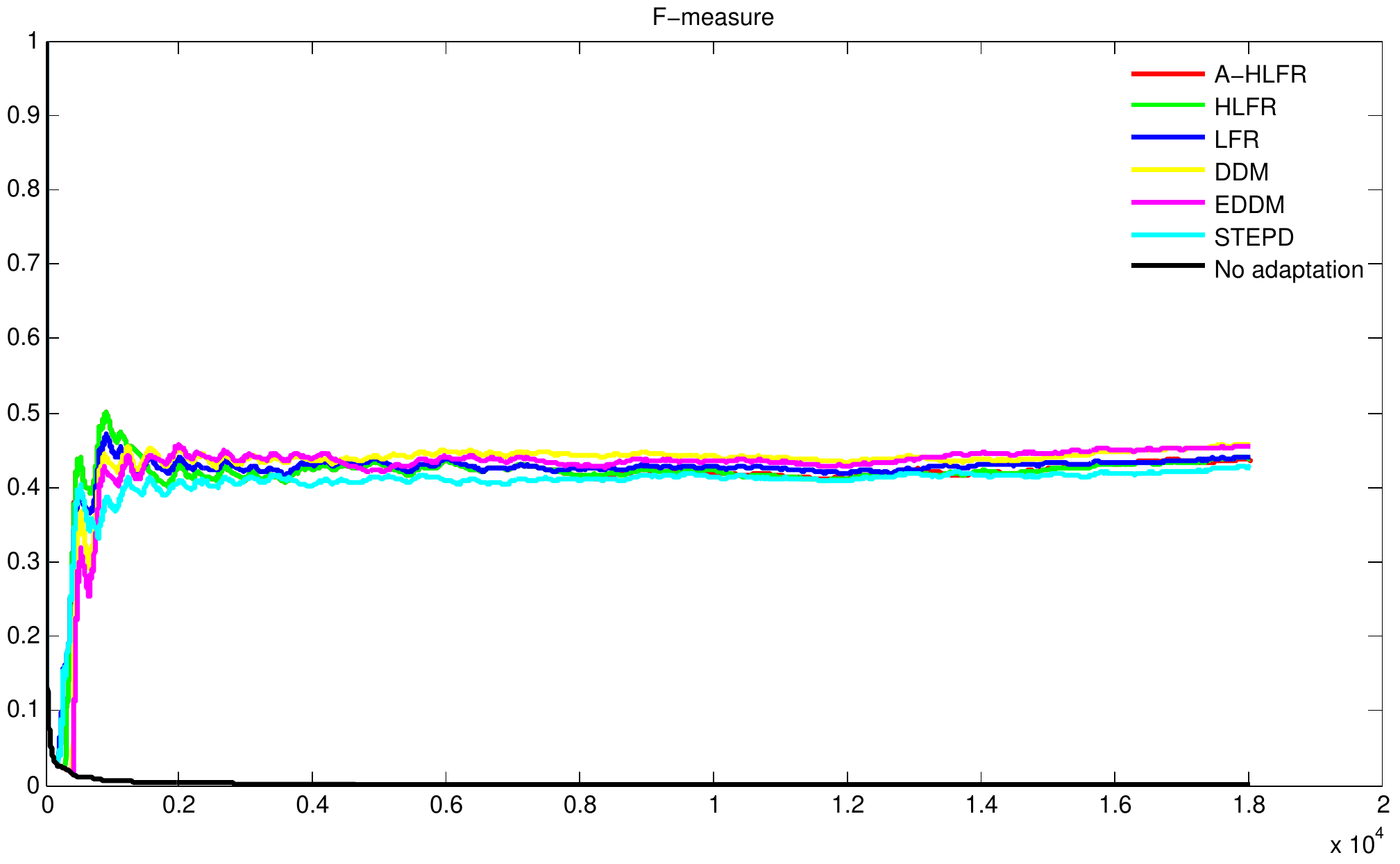}}\\
\subfigure[] {\includegraphics[height=3.5cm,width=.45\textwidth]{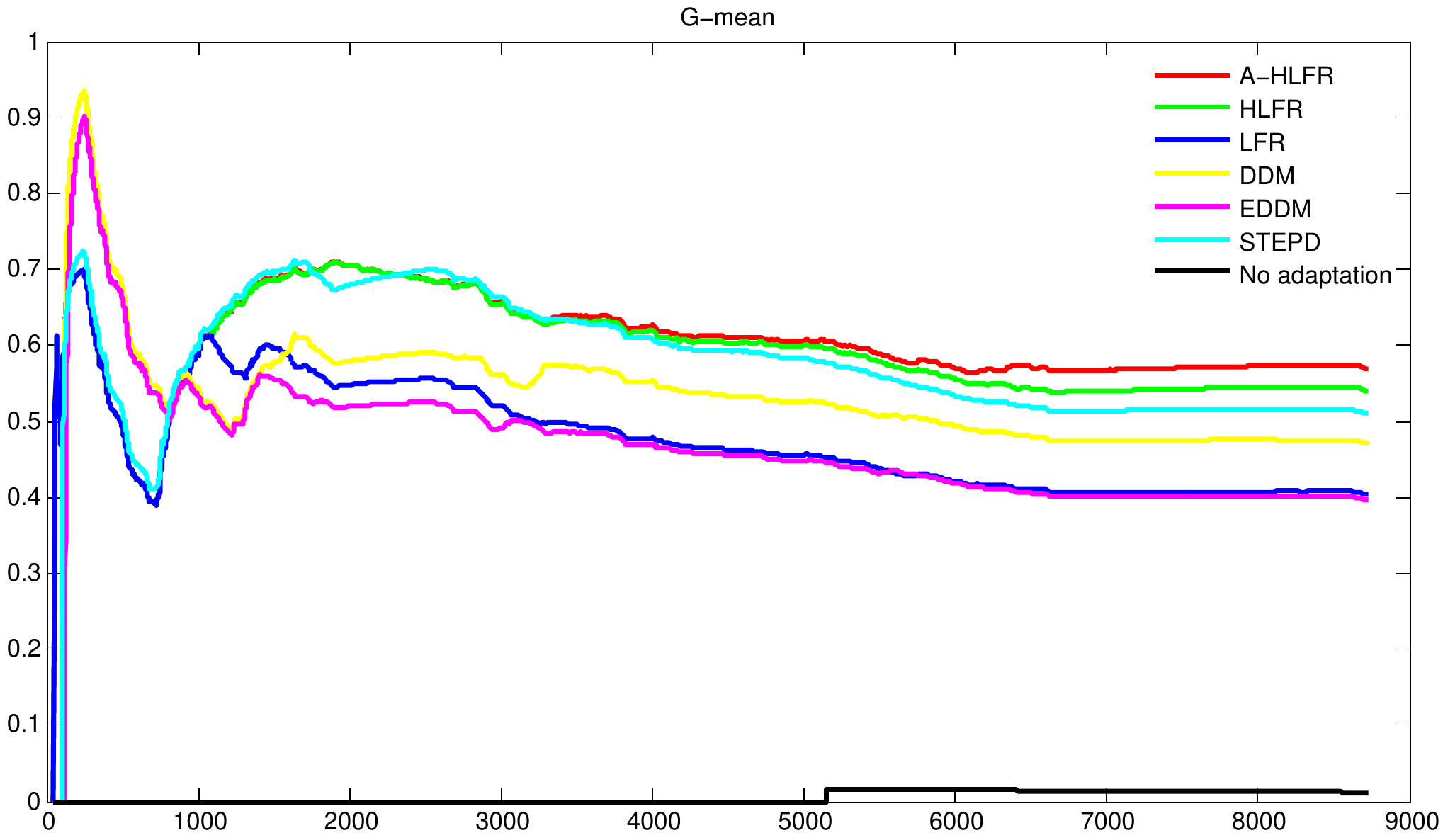}}
\subfigure[] {\includegraphics[height=3.5cm,width=.45\textwidth]{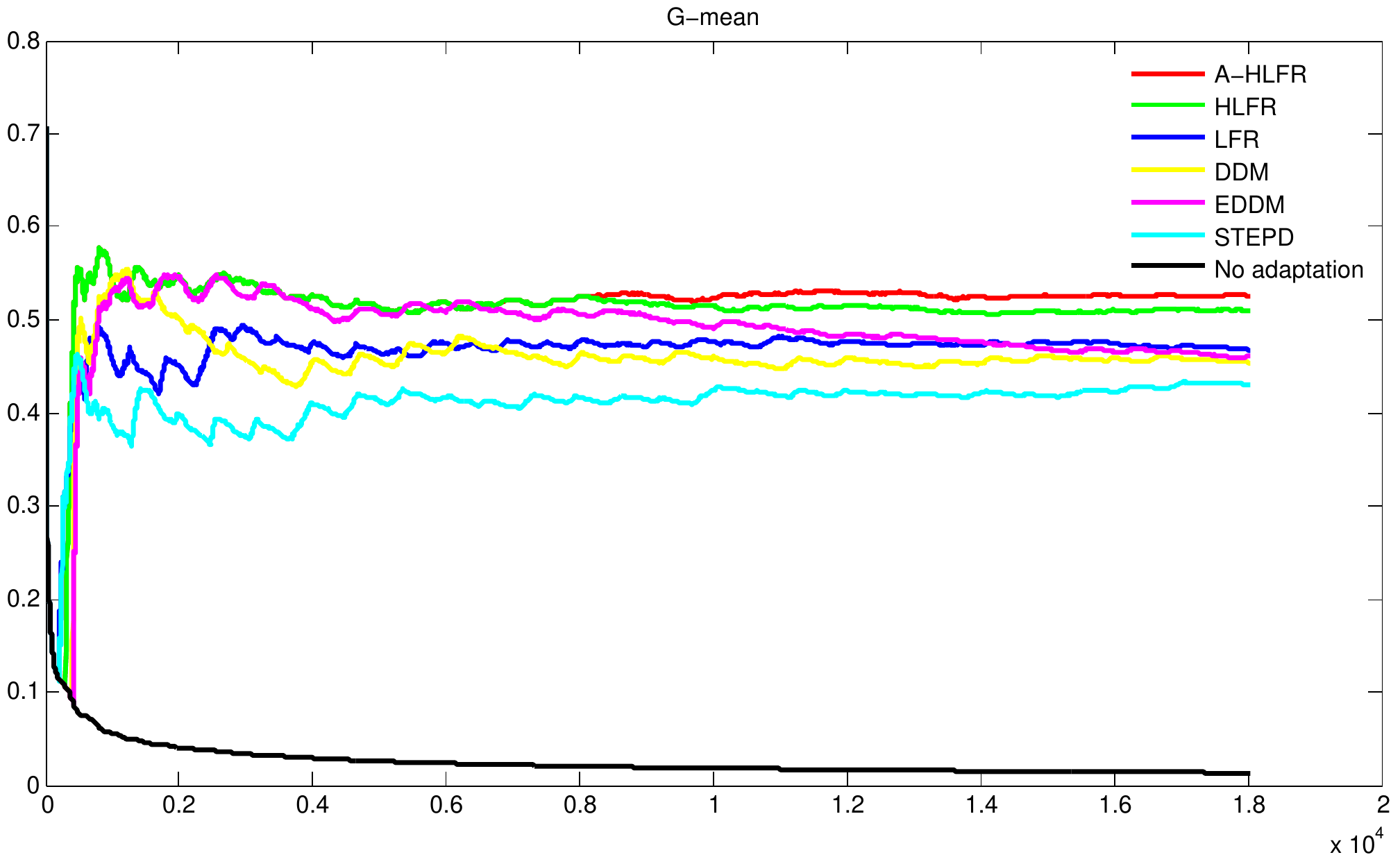}}
\end{tabular}
\caption{The time series representations of different metrics for all competing algorithms. (a) and (b): the OAC representations for spam data and weather data, respectively. (c) and (d): the F-measure representations for spam data and weather data, respectively. (e) and (f): the G-mean representations for spam data and weather data, respectively.}
\label{fig:classification}
\end{figure*}

We further employ several different quantitative measurements to have a thorough evaluation on streaming classification performance. The first measurement is the most commonly used overall accuracy (OAC). Although OAC is an important metric, it is inadequate for imbalanced data. Therefore, we include the F-measure\footnote{$\text{F-measure}=2\times(\frac{\mathbf{Precision}\times \mathbf{Recall}}{\mathbf{Precision}+\mathbf{Recall}})$.}~\cite{Rijsbergen1979Information} and the G-mean\footnote{$\text{G-mean}=\sqrt{\mathbf{Acc^+}\times \mathbf{Acc^-}}$, where $\mathbf{Acc^+}$ and $\mathbf{Acc^-}$ denote true positive rate and true negative rate respectively.}~\cite{kubat1997learning} metrics. All metrics are calculated in each time instant, creating a time series representation that ensembles learning curves. Fig. \ref{fig:classification}(a)-(c) plot the time series representations of OAC, F-measure and G-mean for all competing methods. As can be seen, A-HLFR and HLFR typically provide a significant improvement in F-measure and G-mean while maintaining good OAC when compared to their DDM, EDDM and LFR counterparts, with A-HLFR performs slightly better than HLFR. STEPD seems to demonstrate the best overall classification performance on the spam dataset. However, A-HLFR and HLFR provide more accurate (or rational) concept drift detections which best match with cluster assignments results in \cite{katakis2010tracking}.
%We also average each measurements over the entire testing set to form a single value that represents how well an algorithm performs on a particular data set. The averages ($i.e.$, OAC, F-measure, G-mean) are then ranked to make comparisons among algorithms. The ranks can range from (1) to (6), where (1) is the best and (6) is the worst performing algorithm. Table \ref{Tab:spam_ranking} summarized the quantitative evaluation results.

%\begin{table}[!hbpt]
%\begin{center}
%\caption{Elec2 data classification performance and ranking summary}\label{Tab:elec2_ranking}
%\begin{tabular}{cccccc}\hline
%Algorithms & OAC & F-measure & G-mean & Average Rank \\ \hline
%A-HLFR &  482(502)  & \textbf{55}(37) & \textbf{120}(85) & \textbf{17}(7) \\
%HLFR &  \textbf{458}(486) & 56(36) & 127(112) & 17(7) \\
%LFR &  \textbf{458}(486) & 56(36) & 127(112) & 17(7) \\
%DDM &  1209(450) & 69(56) &  125(128) & 26(17) \\
%EDDM &  939(550) & 93(50) &  166(121) & 36(8) \\
%STEPD &  463(423) & 57(58) &  140(118) & 19(7) \\\hline
%\end{tabular}
%\end{center}
%\end{table}

~\\
\noindent
\textbf{Case study on weather dataset.}
We then evaluate the performances of different methods on the weather dataset. Because the ground-truth drift point location is not available, we only demonstrate the concept drift adaptation comparison results. Fig. \ref{fig:kappa}(b) plots the KPS prequential representations. As can be seen, A-HLFR performs (or updates) best in majority of time segments. Fig. \ref{fig:classification}(d)-(f) plot the corresponding OAC, F-measure and G-mean time series representations for all competing algorithms. Although the \emph{no adaptation} (i.e., using the initial trained classifier for prequential classification without any classifier update) enjoys an overwhelming advantage in OAC compared to DDM, EDDM, LFR, STEPD, it is however invalid as the corresponding F-measure and G-mean tend to be zero as time evolves. This suggests that if \emph{no adaptation} is adopted, the initial classifier gradually identifies the remaining data as belonging to the majority class, i.e., no rain days, which is not realistic. A-HLFR and HLFR achieves close OAC values to the non-adaptive classifier, however, shows significant improvements on F-measure and G-mean. Again, A-HLFR performs slightly better than HLFR.

From these two real applications, we can summarize some key observations:\\
1) The given data has severe concept drifts, as the classification performance of \emph{no adaptation} deteriorates dramatically.\\
2) The adaptive training will not affect the performance of concept drift detection, as the concept drift detection results given by HLFR and A-HLFR are almost the same (see Fig.~\ref{fig:detection}). This argument is further empirically validated and elucidated in the next subsection.\\
3) A-HLFR and HLFR consistently produce the best overall performance in terms of OAC, F-measure, G-mean and the rationality of drift detected. For real data, A-HLFR only performs slightly better than HLFR. This is because the temporal relatedness between consecutive concepts in real-world data is weak or the concept changes gradually and slowly such that simply transferring previous knowledge to current domain (or concept) cannot prompt the generalization capacity of new classifier significantly. Therefore, adaptive training has great potency, but it deserves more investigations and future improvements. \\
4) There is still plenty of room for performance improvement on incremental learning under concept drifts in nonstationary environment, as the OAC, F-measure and G-mean values are far from optimal. In fact, even with the state-of-the-art methods which only focus on automatically adapting classifier behavior (or parameters) to stay up-to-date with the streaming data dynamics, the OAC can only reach to approximately $90\%$ in \cite{katakis2010tracking,katakis2006dynamic} for spam data and $80\%$ in \cite{ditzler2013incremental,elwell2011incremental} for weather data, let alone the relatively lower F-measure and G-mean values.\\
5) The ensemble of classifiers seems to be a promising direction for future work. However, most of the existing ensemble learning based methods (e.g., \cite{ditzler2013incremental,elwell2011incremental}) are developed for batch-incremental data~\cite{read2012batch}, which is not suitable for a fully online setting, where the sample is provided one by one in a sequential manner \cite{gama2014survey}.
%5) There is still a gap between ....\\

\subsection{Benefits of adaptive learning} \label{section5.5}
In this section, we demonstrate, via the application of concept drift adaptation on USENET1 and Checkerboard datasets, that the superiority of adaptive SVM for concept drift adaptation is not limited to the HLFR framework. To this end, we consider the algorithm performance of integrating adaptive SVM into DDM, EDDM, DDM-OCI, STEPD as well as LFR framework. We term this combinations A-DDM, A-EDDM, A-DDM-OCI, A-STEPD and A-LFR, respectively.

\begin{figure*}[!htbp]
\centering
\begin{tabular}{ccc}
\subfigure[Precision over Checkerboard dataset (Group I)] {\includegraphics[width=.25\textwidth]{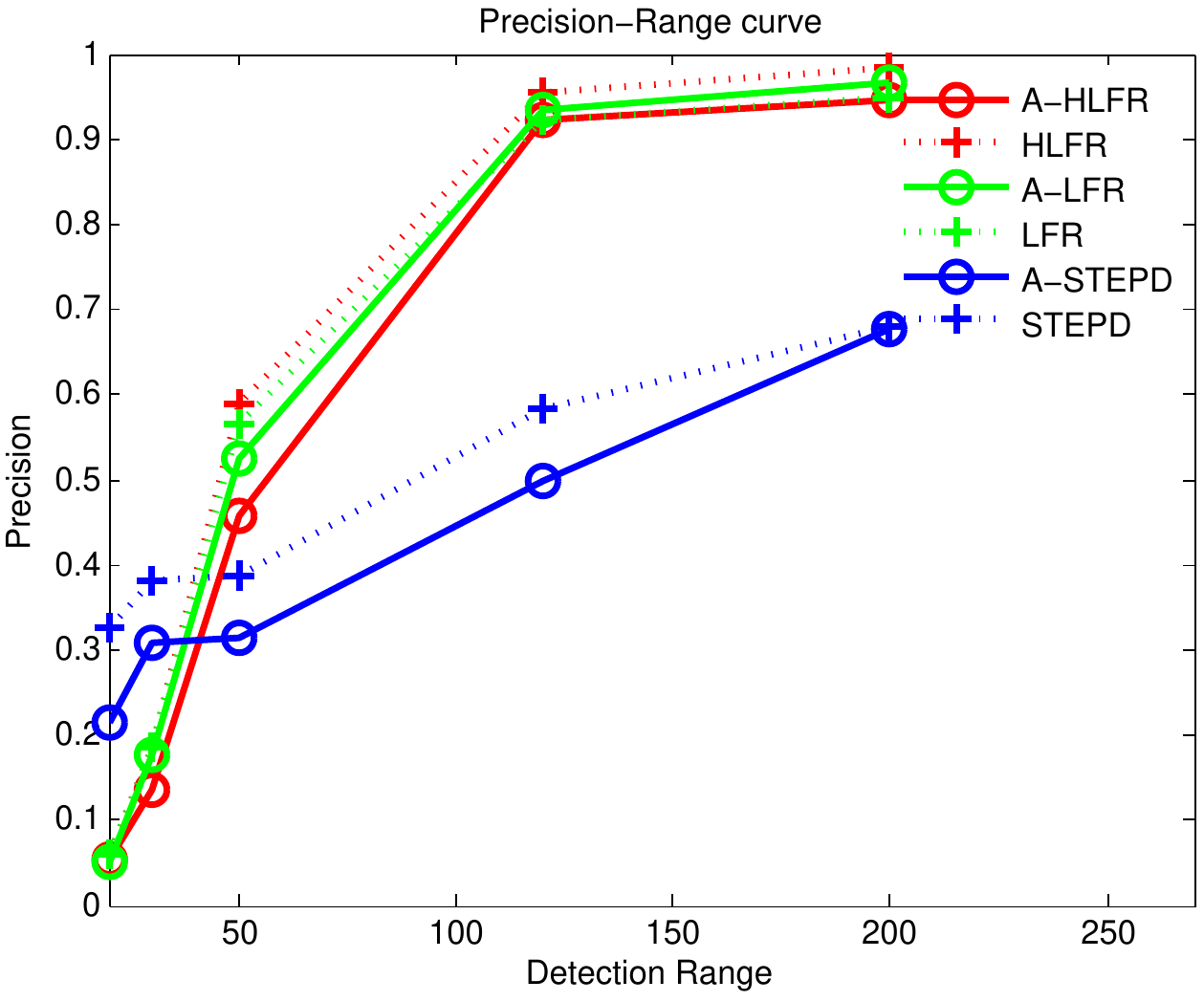}}
\subfigure[Precision over Checkerboard dataset (Group II)] {\includegraphics[width=.25\textwidth]{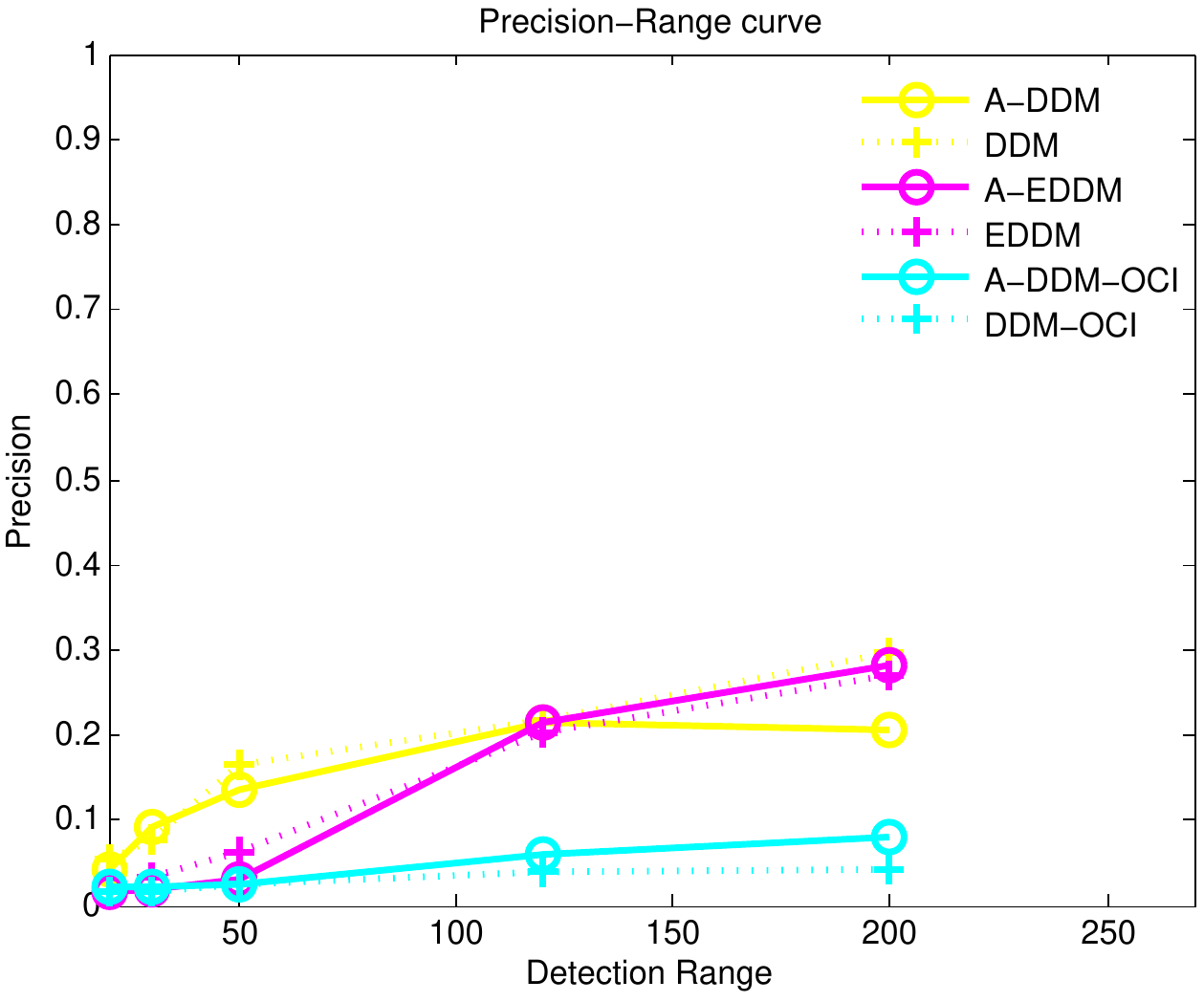}}
\subfigure[Precision over USENET1 dataset (Group I)] {\includegraphics[width=.25\textwidth]{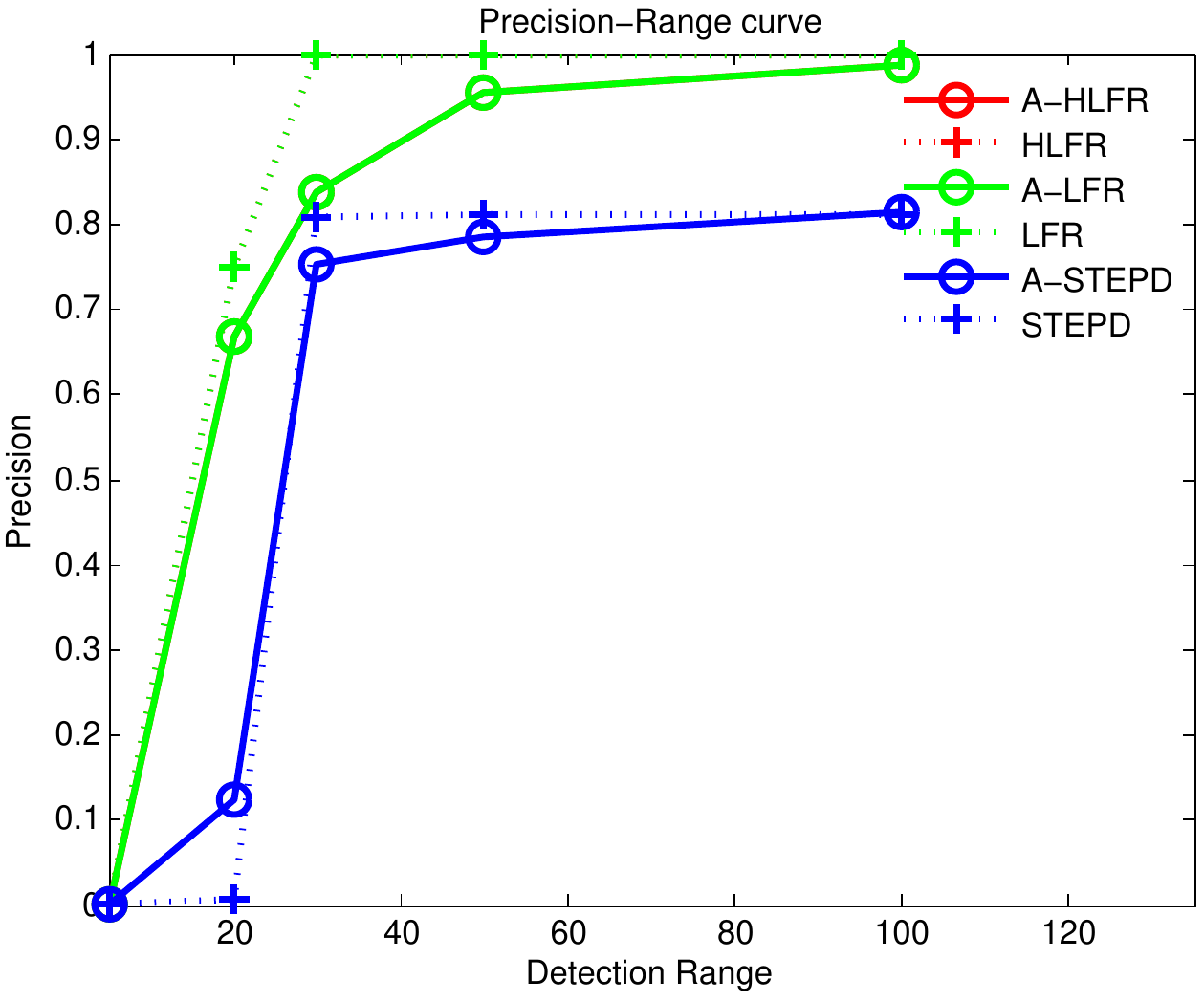}}
\subfigure[Precision over USENET1 dataset (Group II)] {\includegraphics[width=.25\textwidth]{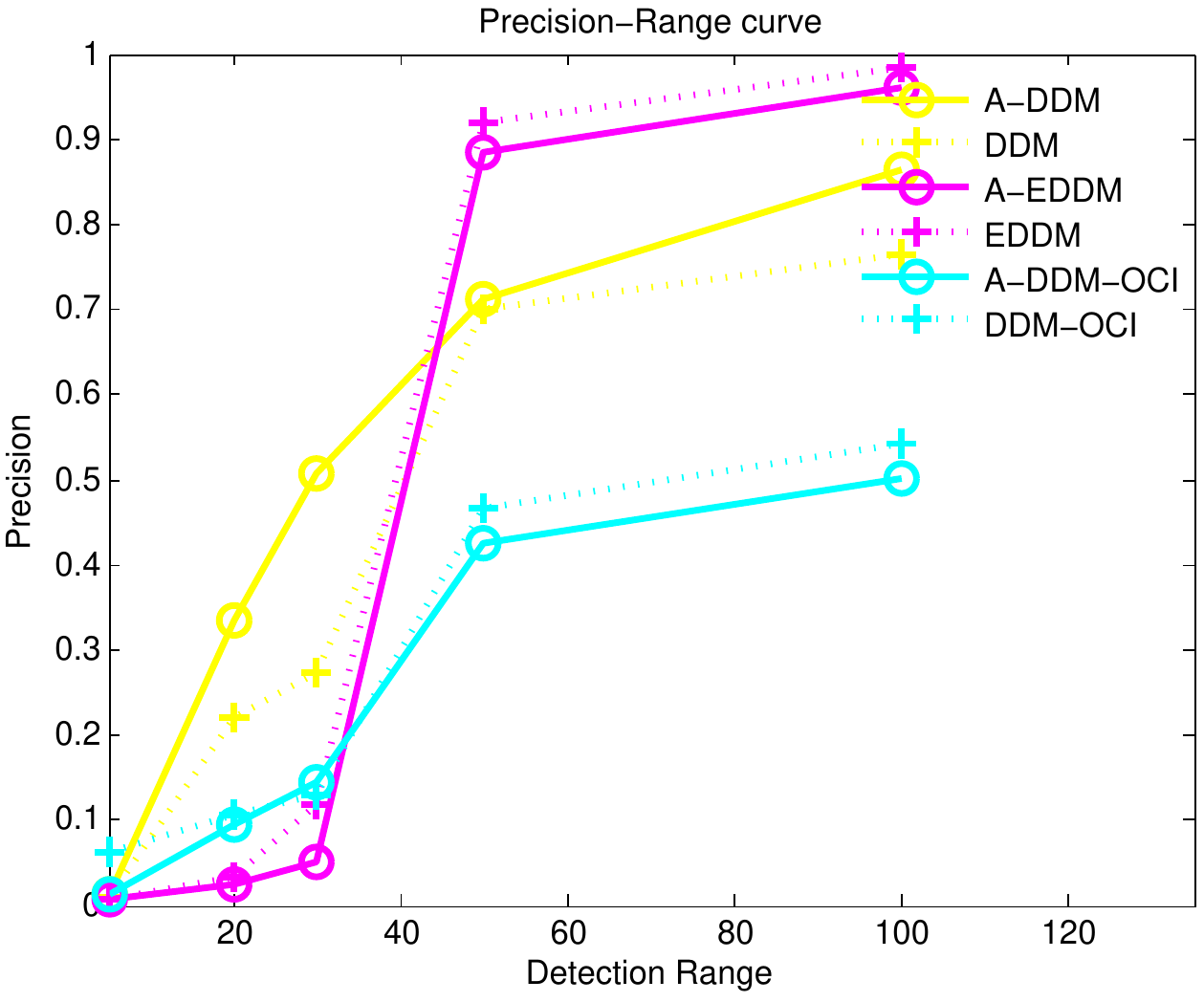}} & \\
\subfigure[Recall over Checkerboard dataset (Group I)] {\includegraphics[width=.25\textwidth]{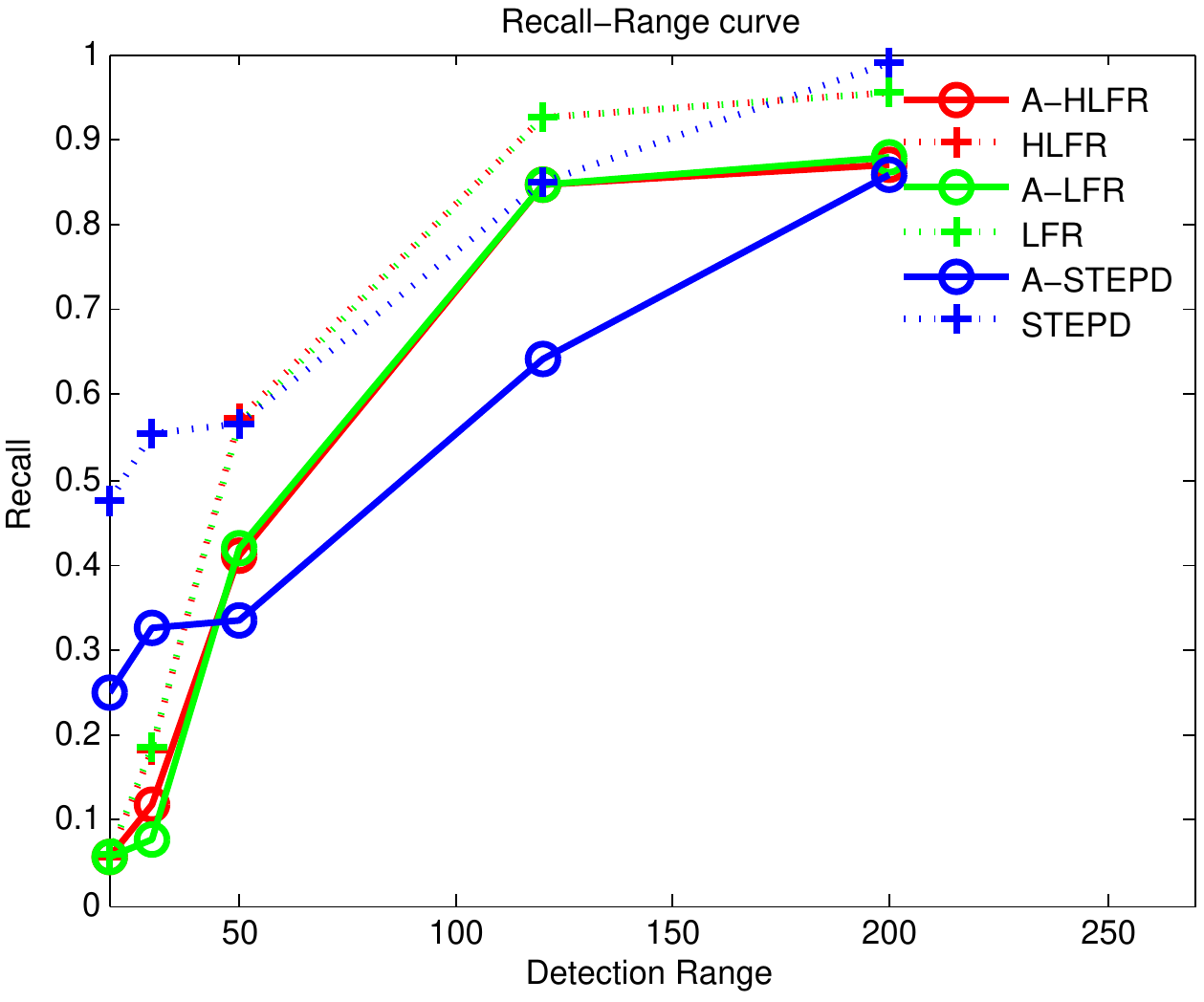}}
\subfigure[Recall over Checkerboard dataset (Group II)] {\includegraphics[width=.25\textwidth]{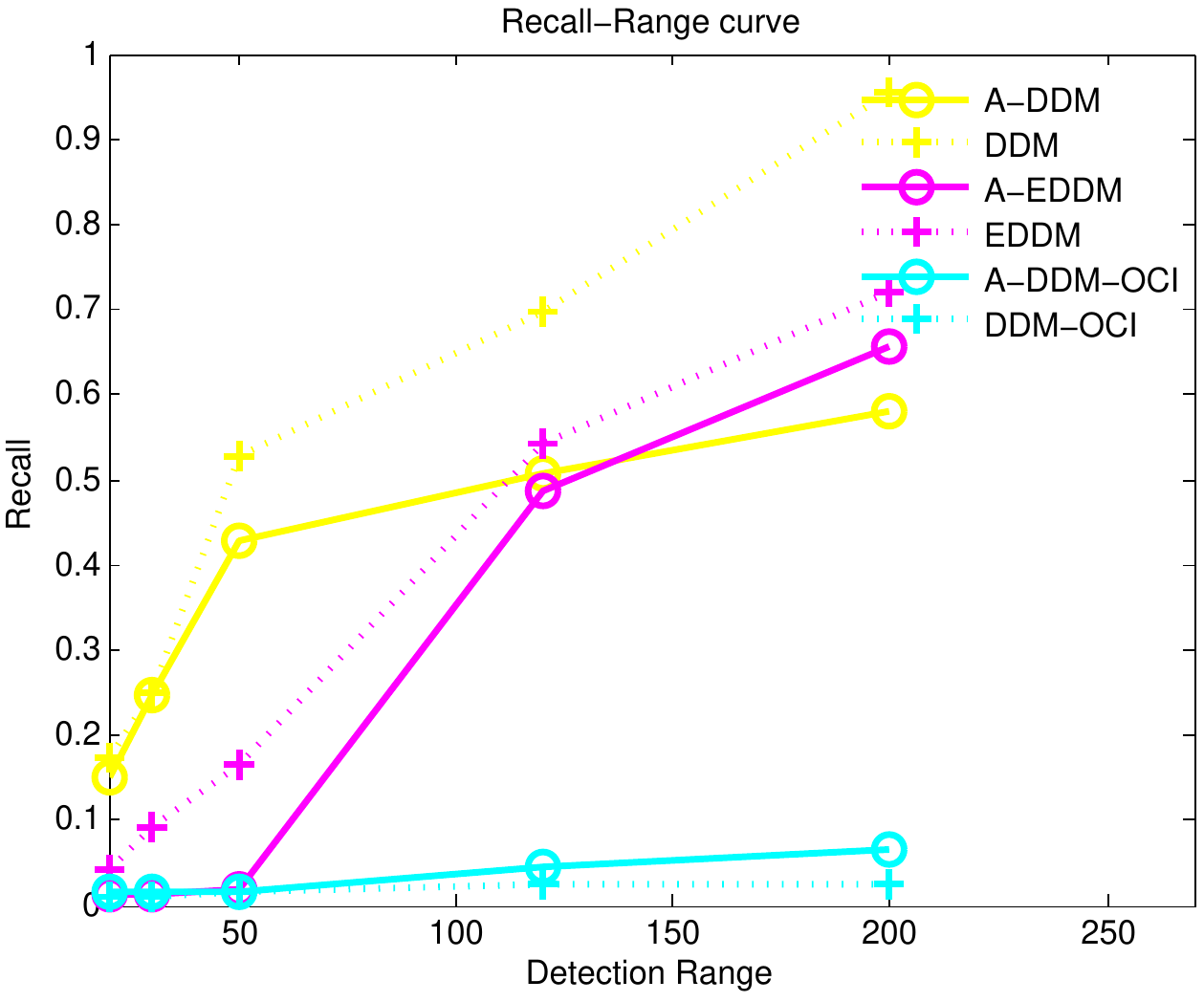}}
\subfigure[Recall over USENET1 dataset (Group I)] {\includegraphics[width=.25\textwidth]{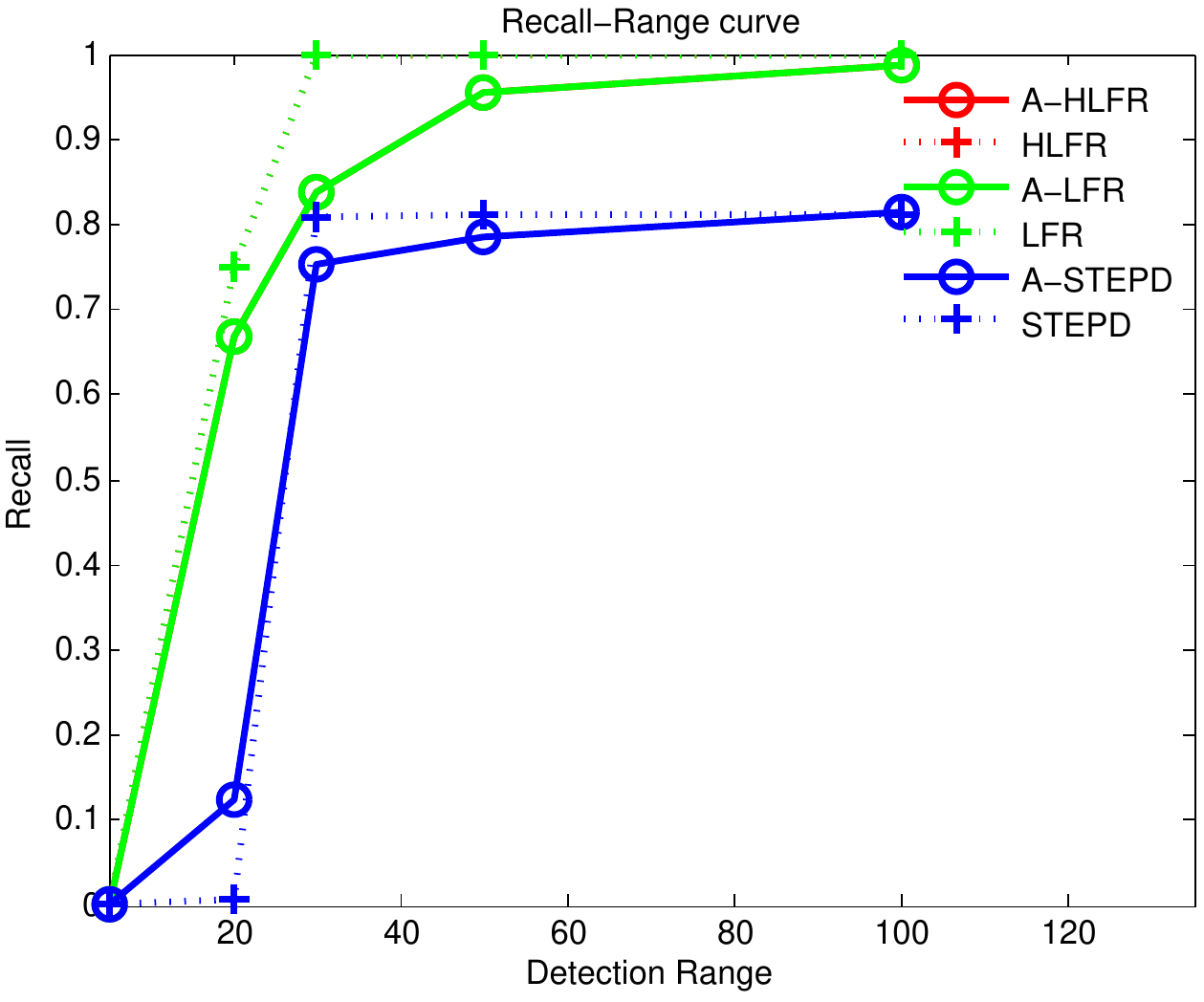}}
\subfigure[Recall over USENET1 dataset (Group II)] {\includegraphics[width=.25\textwidth]{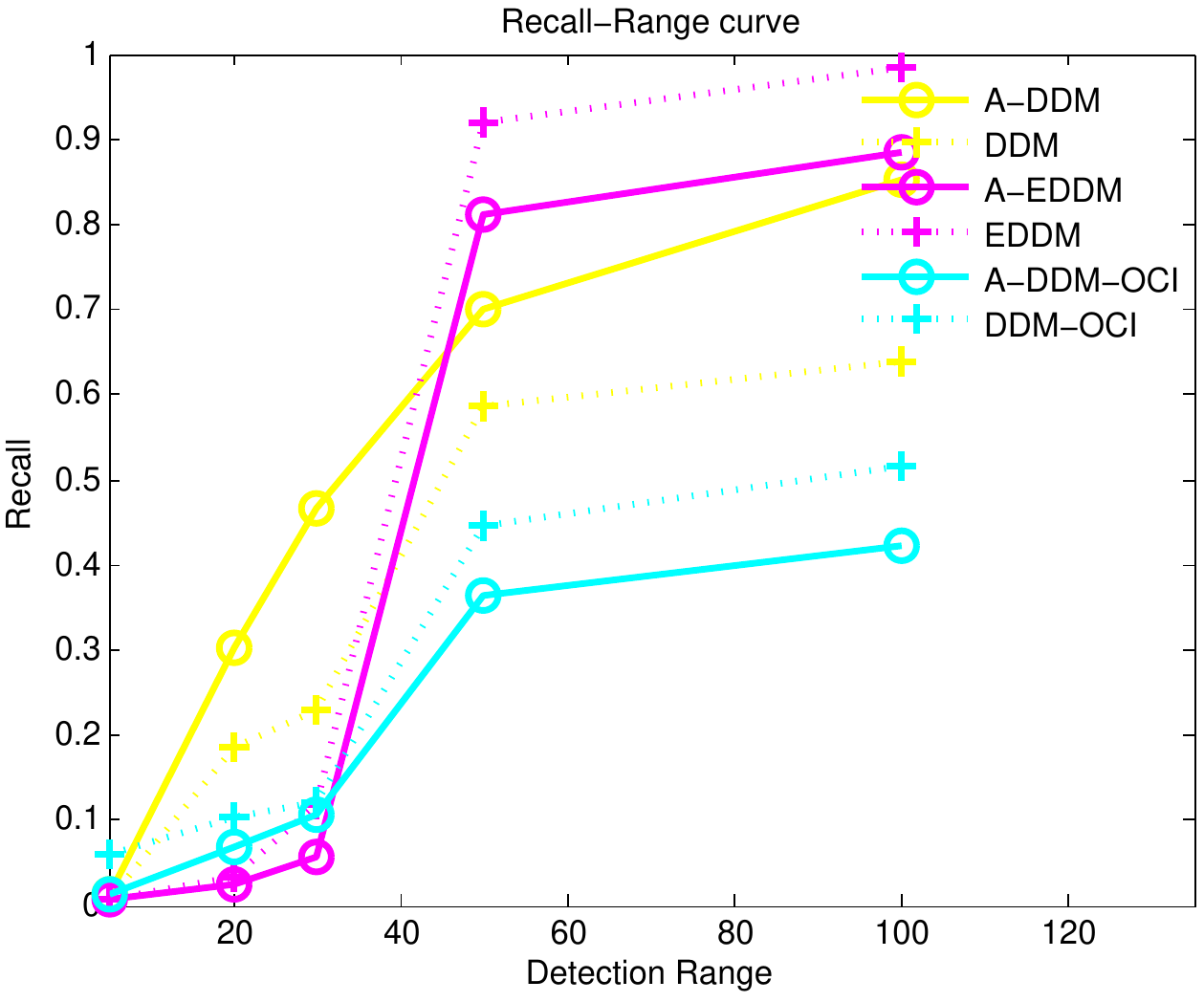}}\\
\end{tabular}
\caption{Summary of $\mathbf{Precision}$ and $\mathbf{Recall}$ over Checkerboard and USENET1 datasets for all competing algorithms and their adaptive versions. The X-axis in each figure represents the pre-defined detection delay range, whereas the Y-axis denotes the corresponding $\mathbf{Precision}$ and $\mathbf{Recall}$ values. For a specific delay range, a higher $\mathbf{Precision}$ or $\mathbf{Recall}$ value suggests better performance.\vspace{-0.5cm}}
\label{fig:precision_recall_comparison_update}
\end{figure*}

In Fig. \ref{fig:precision_recall_comparison_update}, we plotted the \textbf{Precision} and \textbf{Recall} curves of HLFR, LFR, DDM, EDDM, DDM-OCI, STEPD, A-HLFR, A-LFR, A-DDM, A-EDDM, A-DDM-OCI and A-STEPD on USENET1 and Checkerboard, respectively. For better visualization, we separate all the competing algorithms into two groups, group I includes HLFR, A-HLFR, LFR, A-LFR, STEPD and A-STEPD as they always perform better than their counterparts, while group II contains DDM, A-DDM, EDDM, A-EDDM, DDM-OCI and A-DDM-OCI. In each subfigure, the dashed line represents the baseline algorithm without adaptive training (e.g., HLFR), while the solid line denotes its adaptive version (e.g., A-HLFR). Meanwhile, for each baseline algorithm, its adaptive version is marked with the same color for comparison purpose. Obviously, the adaptive training will not affect the performance of concept drift detection\footnote{Admittedly, there is performance gap for DDM or STEPD, the difference is, however, data-dependent. For example, DDM seems to be better than A-DDM in Checkerboard dataset, but this advantage does not hold in USENET1.}. This is because the drift is determined by keeping track of ``significant" changes of classification performance, rather than the specific performance measurement itself.

\begin{figure*}[!htbp]
\centering
\begin{tabular}{ccc}
\subfigure[] {\includegraphics[width=.5\textwidth]{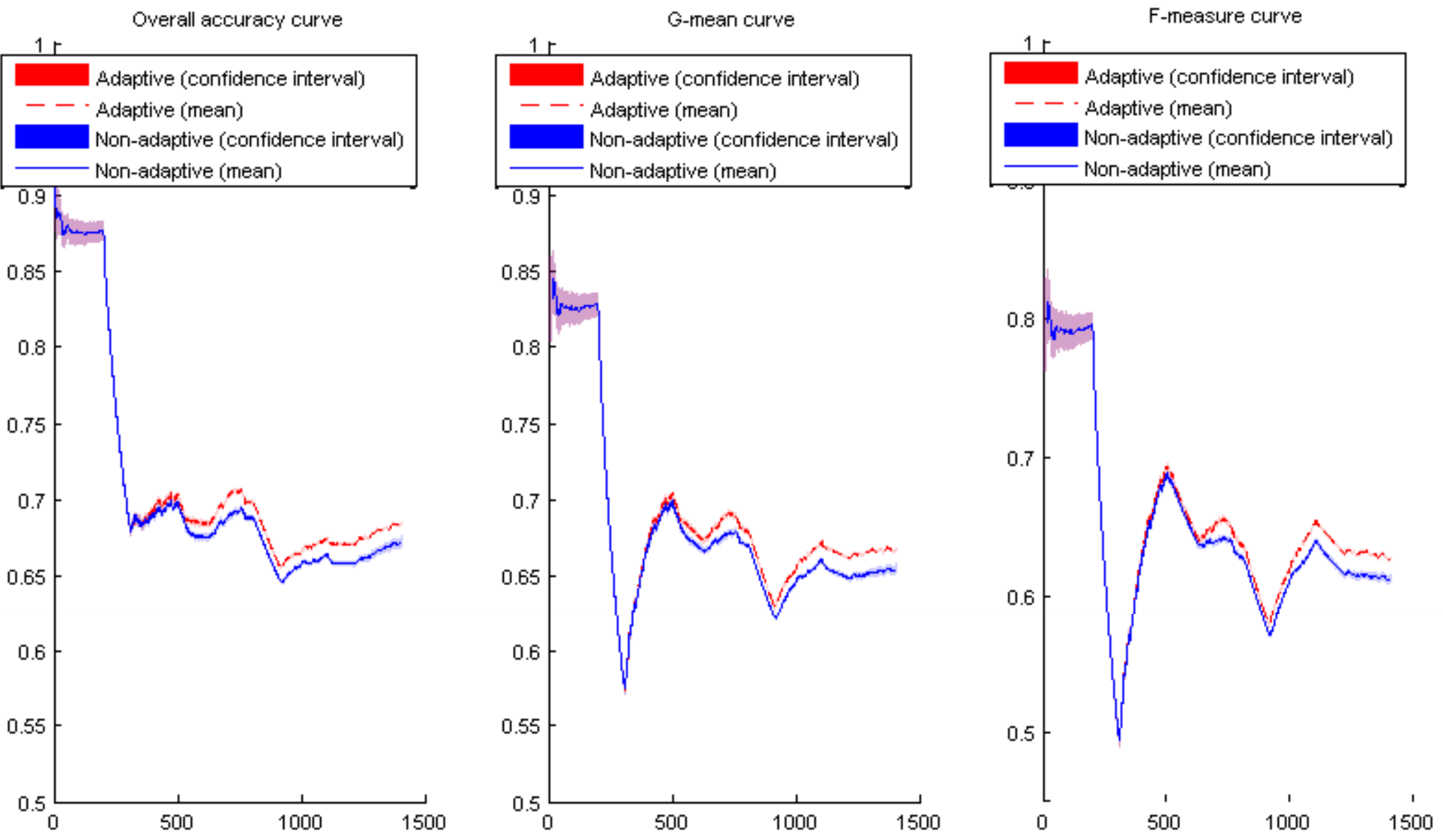}}
\subfigure[] {\includegraphics[width=.5\textwidth]{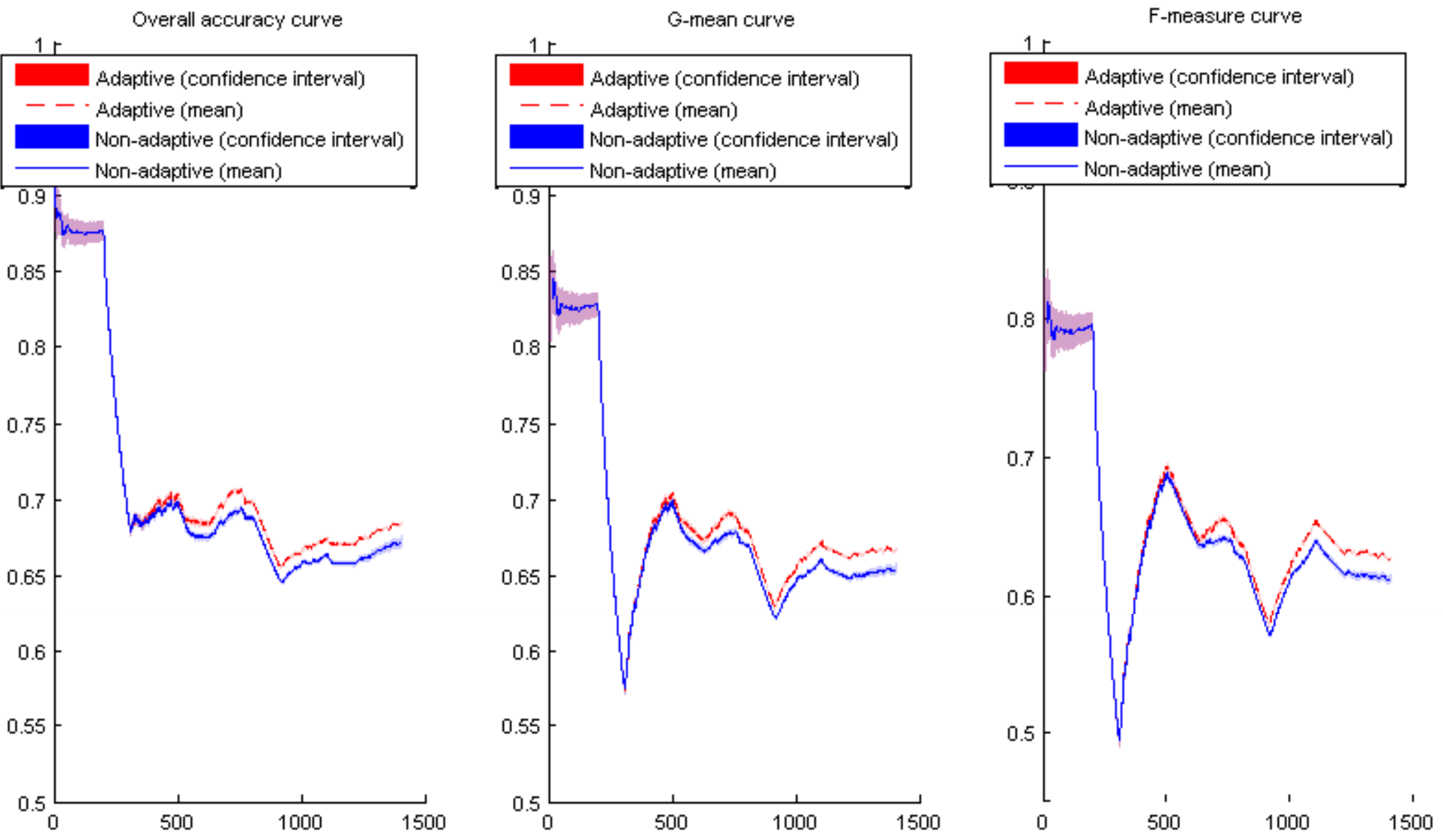}} & \\
\subfigure[] {\includegraphics[width=.5\textwidth]{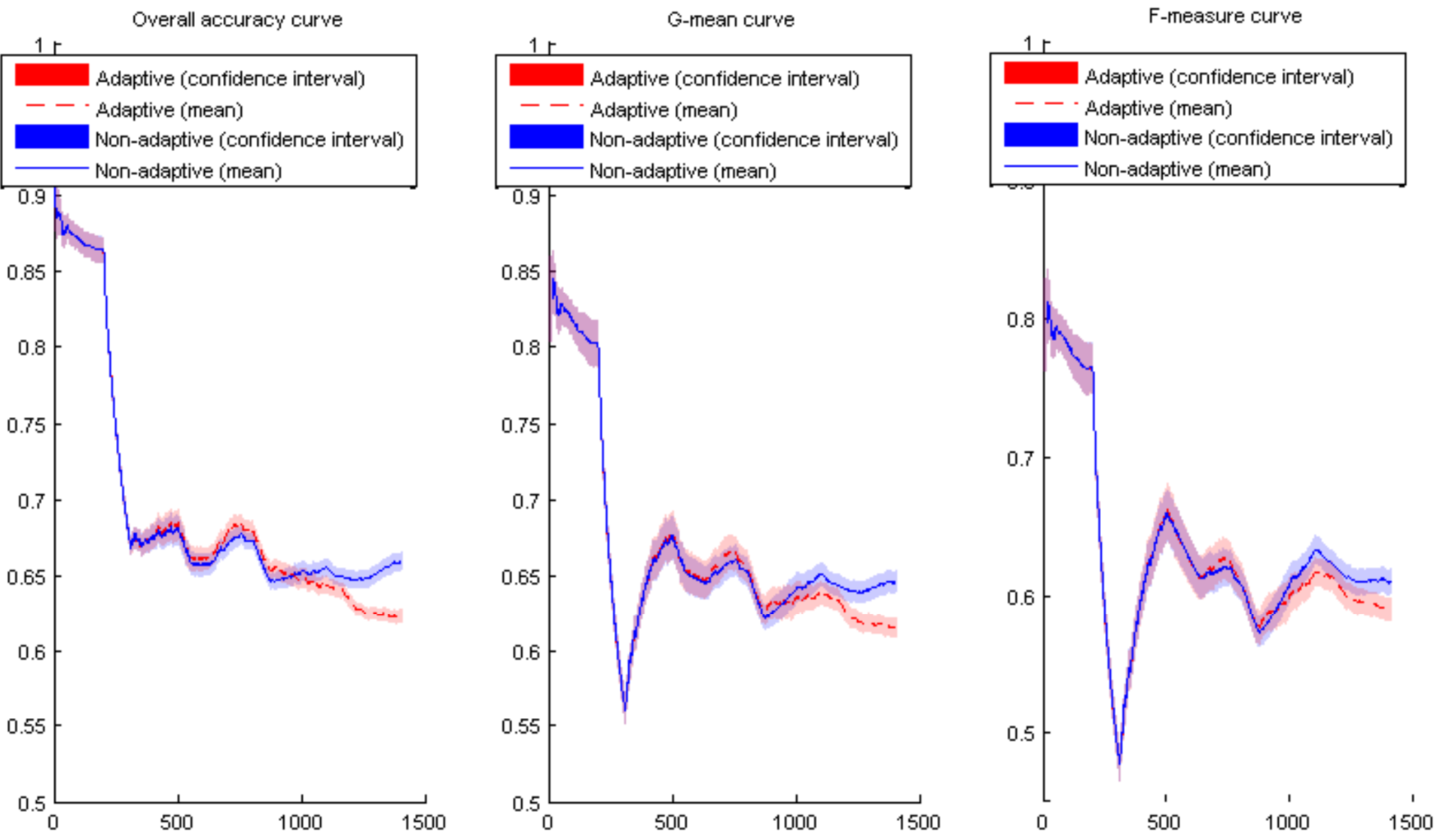}}
\subfigure[] {\includegraphics[width=.5\textwidth]{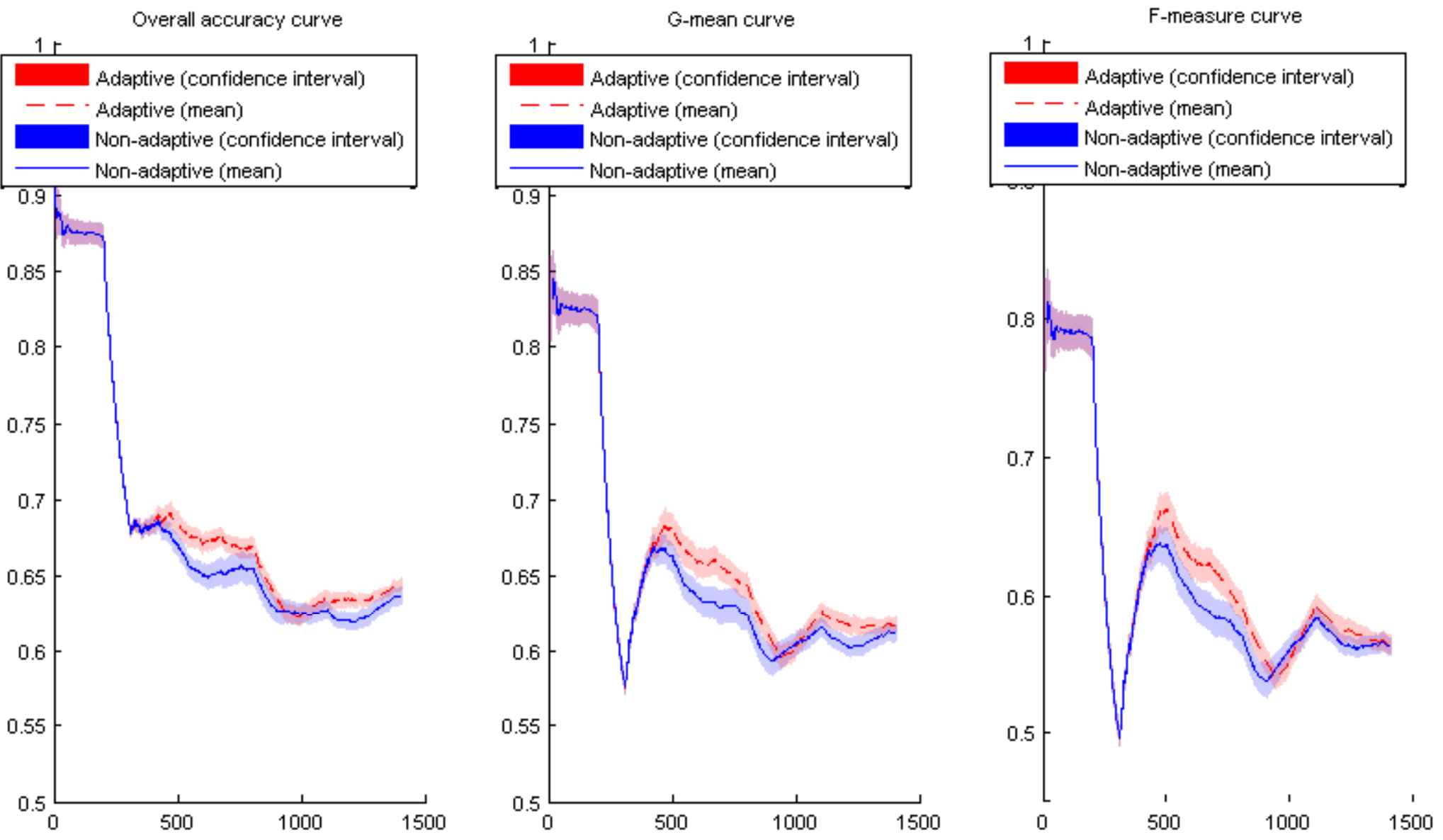}} & \\
\subfigure[] {\includegraphics[width=.5\textwidth]{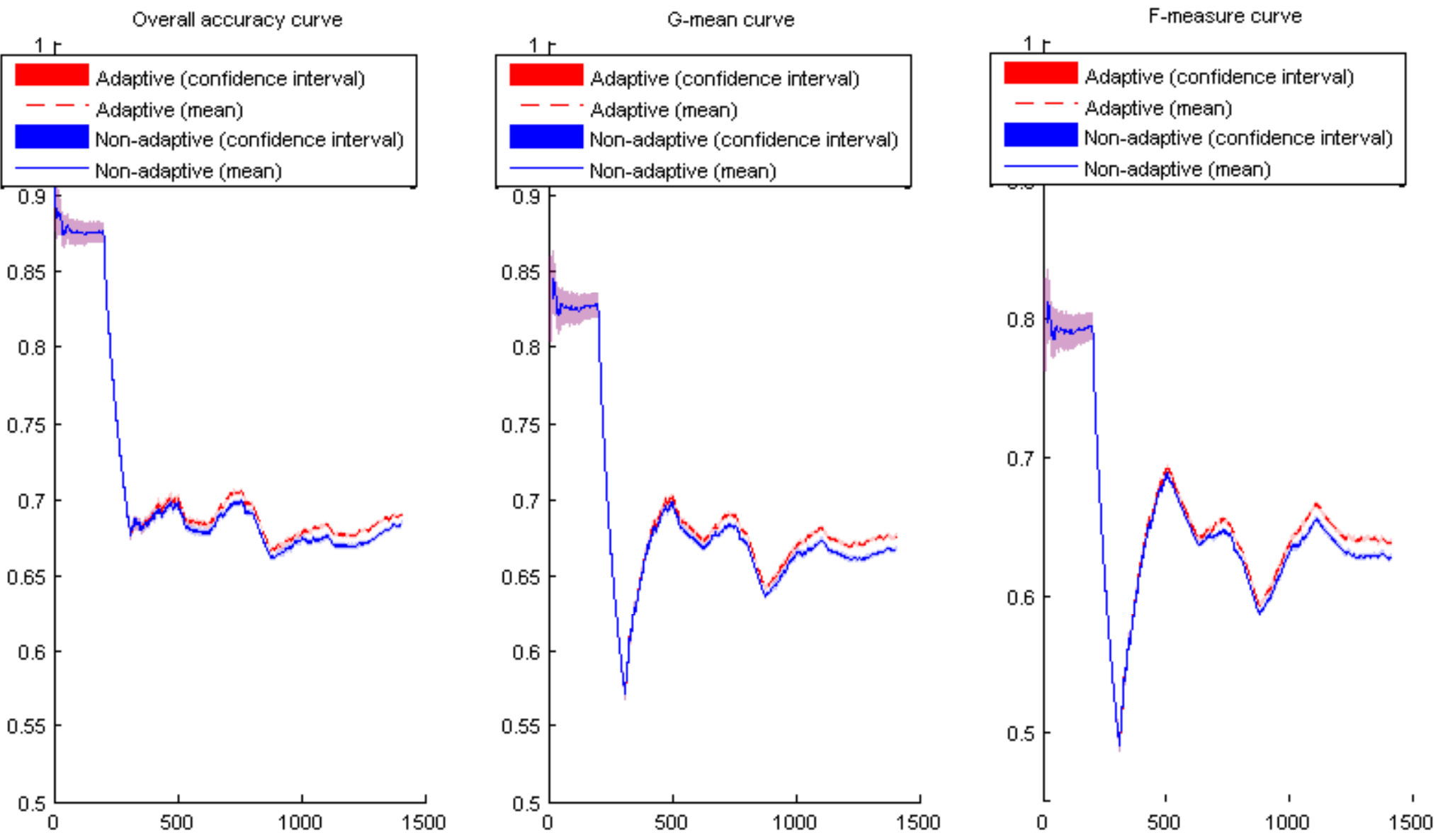}}
\subfigure[] {\includegraphics[width=.5\textwidth]{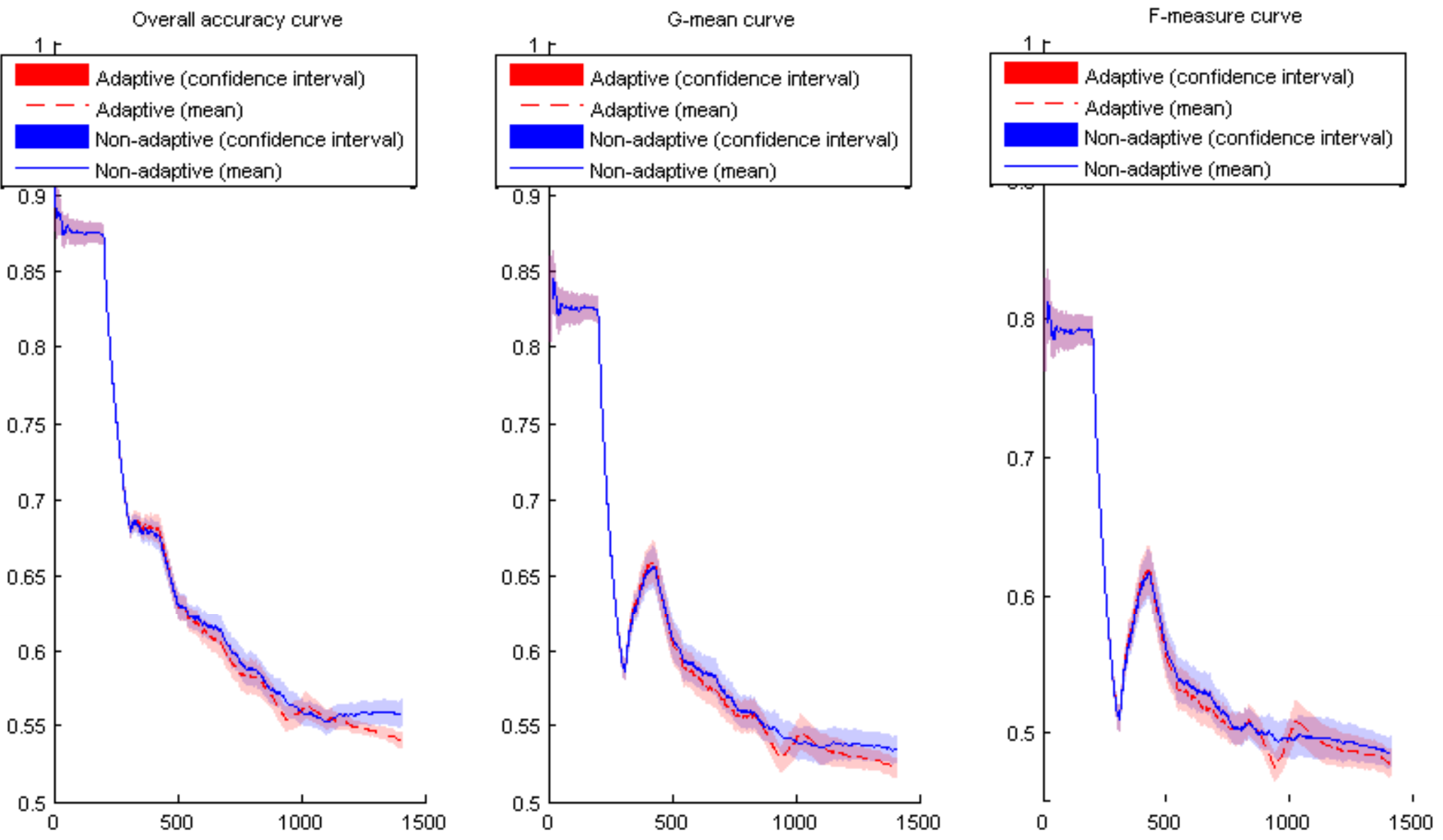}} \\
\end{tabular}
\caption{The time series representations of different metrics (OAC, F-measure, G-mean) on USENET1 dataset for (a) A-HLFR, HLFR; (b) A-LFR, LFR; (c) A-DDM, DDM; (d) A-EDDM, EDDM; (e) A-STEPD, STEPD; and (f) A-DDM-OCI, DDM-OCI. The red dashed line denotes mean values for adaptive learning methods, the red shading envelop represents $95\%$ confidence interval. The blue solid line denotes mean values for non-adaptive learning methods, the blue shading envelop represents $95\%$ confidence interval.}
\label{fig:usenet1_adaptive_learning}
\end{figure*}

\begin{figure*}[!htbp]
\centering
\begin{tabular}{ccc}
\subfigure[] {\includegraphics[width=.5\textwidth]{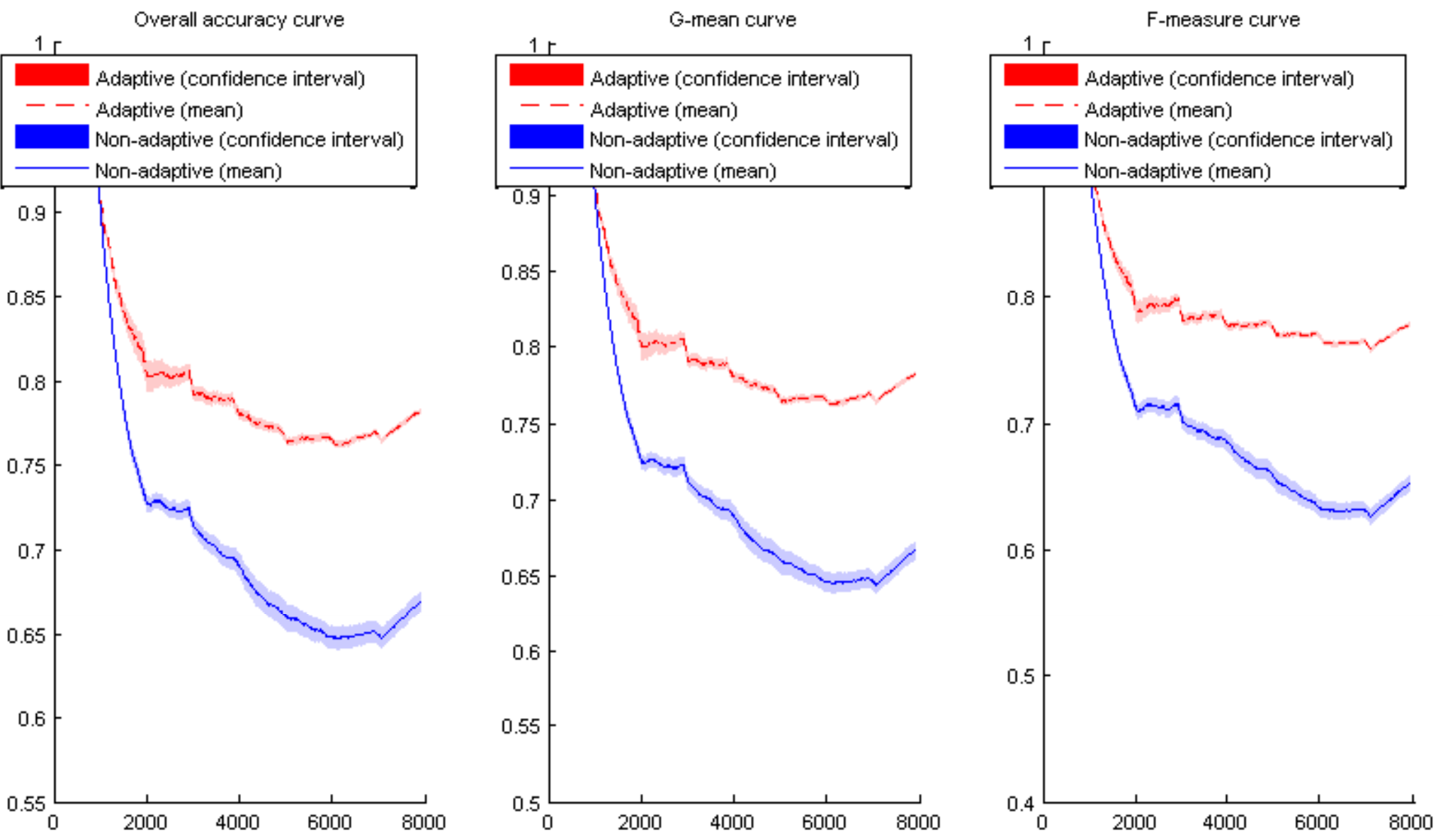}}
\subfigure[] {\includegraphics[width=.5\textwidth]{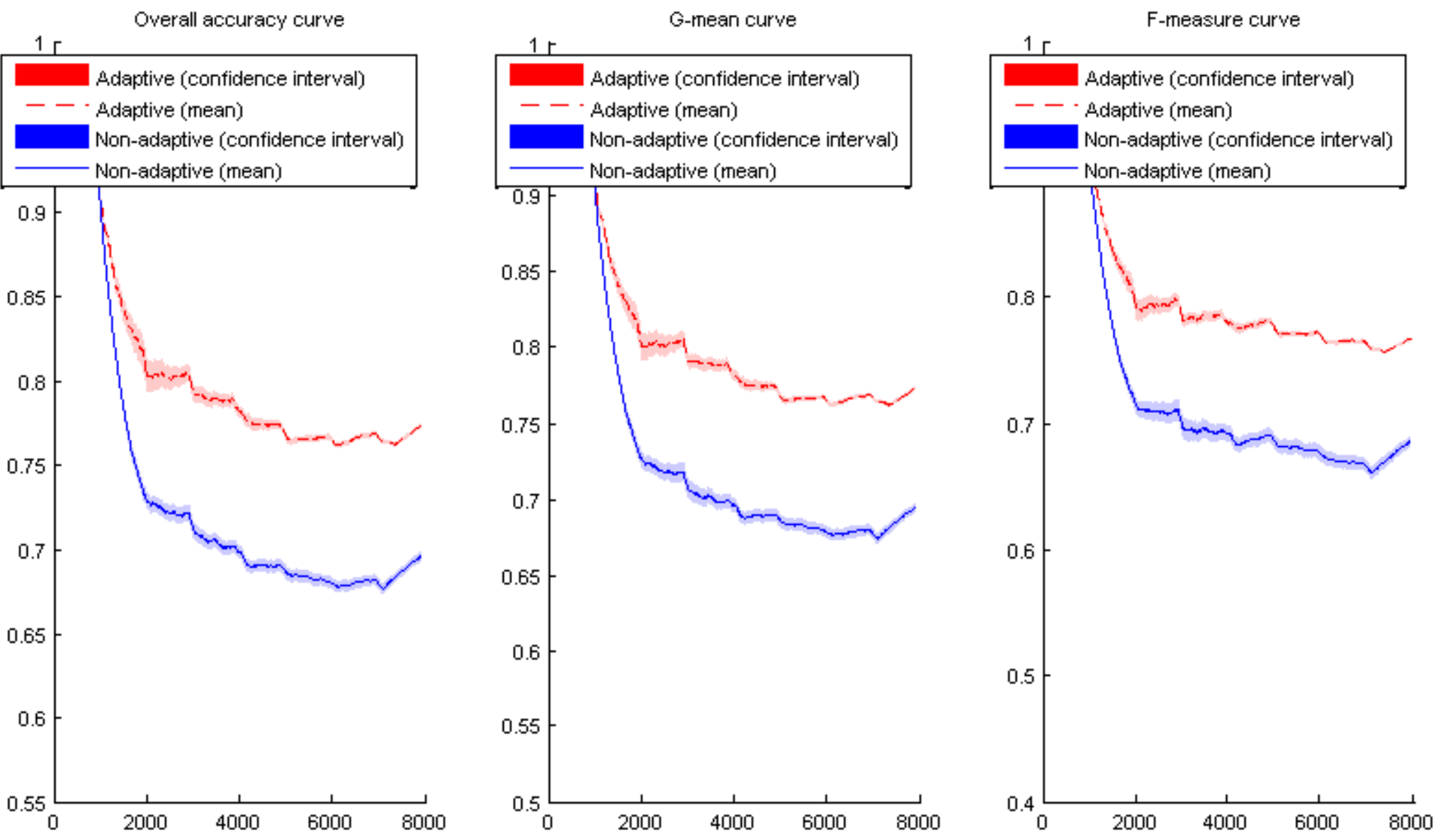}} & \\
\subfigure[] {\includegraphics[width=.5\textwidth]{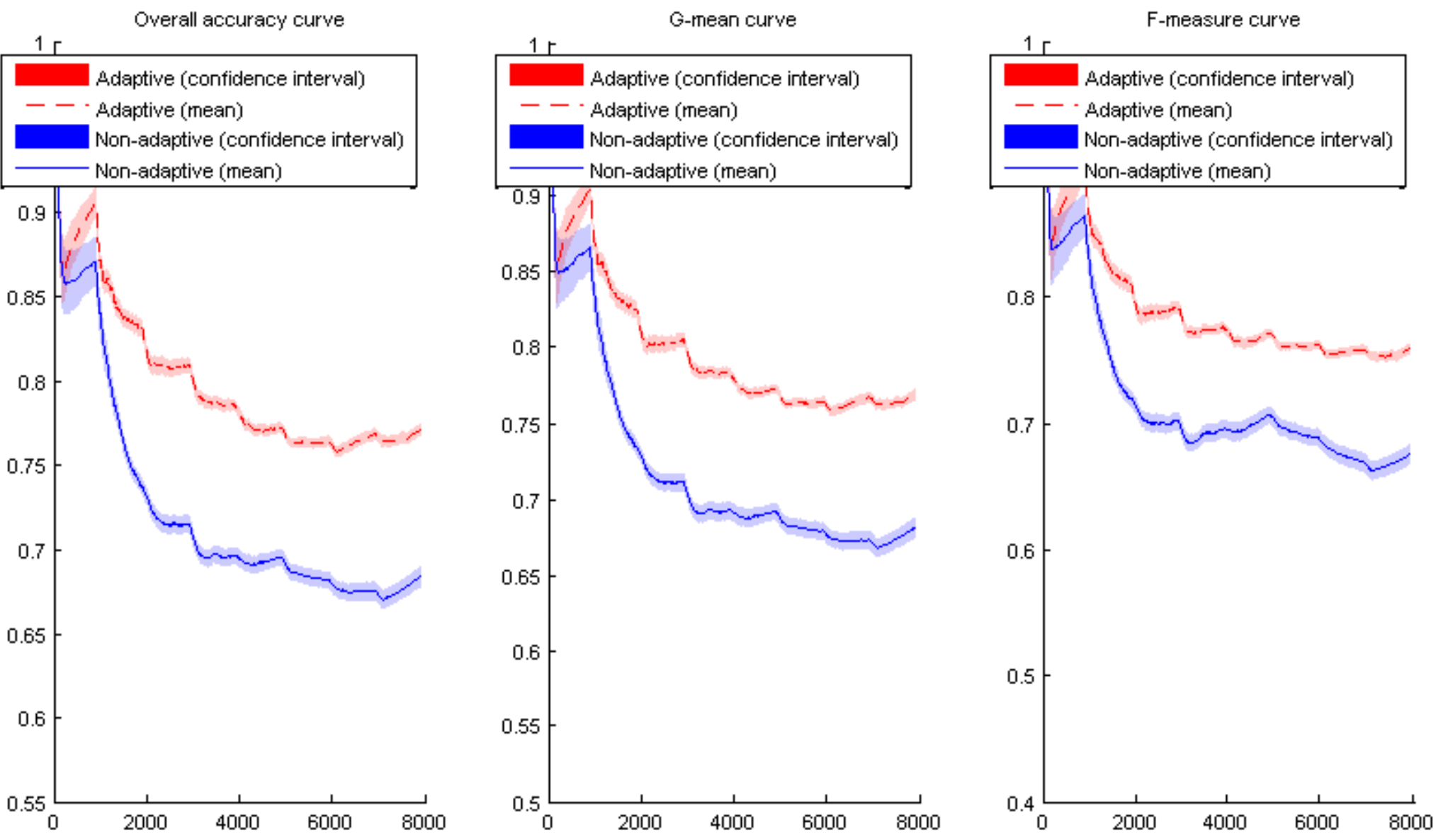}}
\subfigure[] {\includegraphics[width=.5\textwidth]{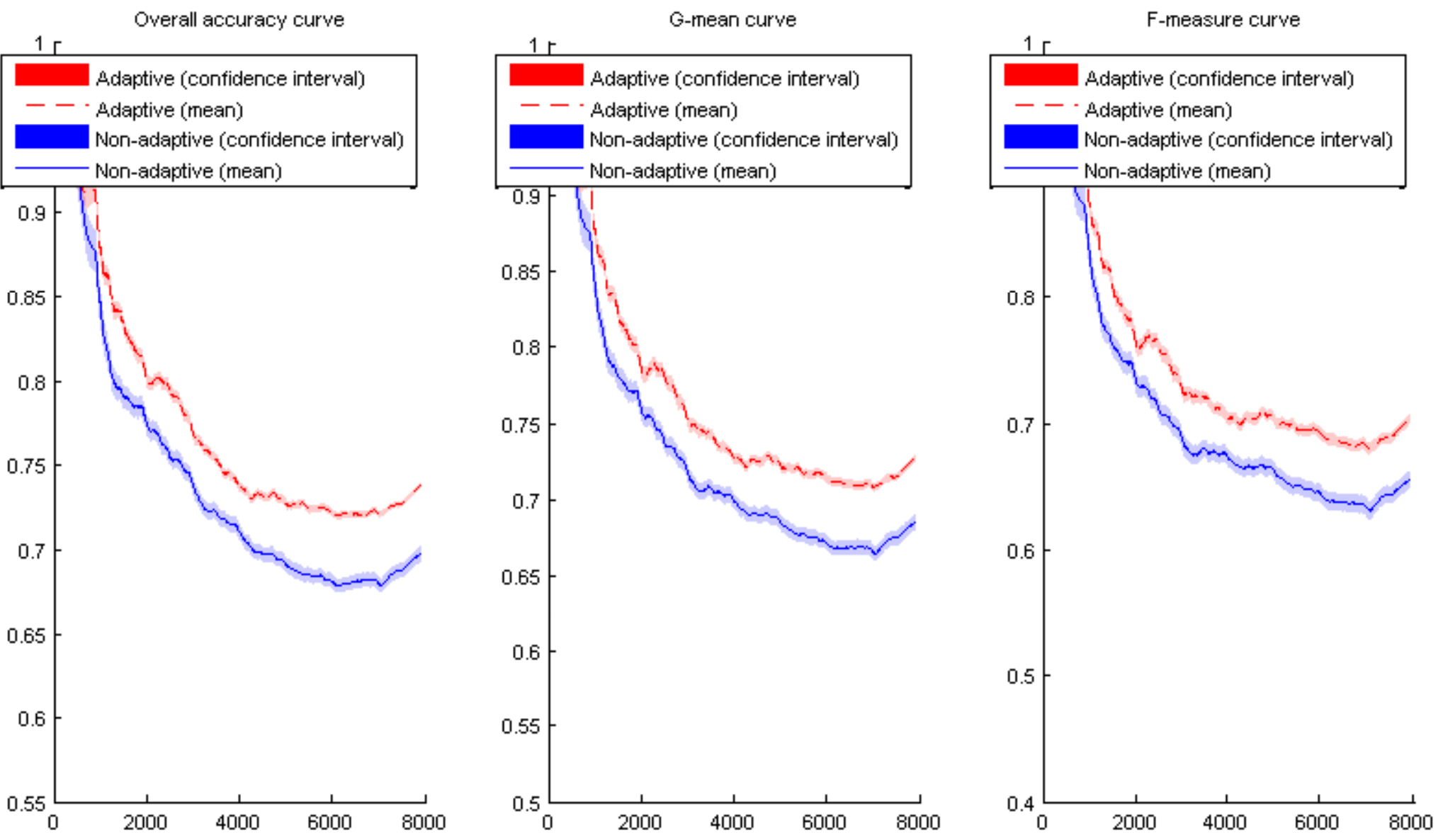}} & \\
\subfigure[] {\includegraphics[width=.5\textwidth]{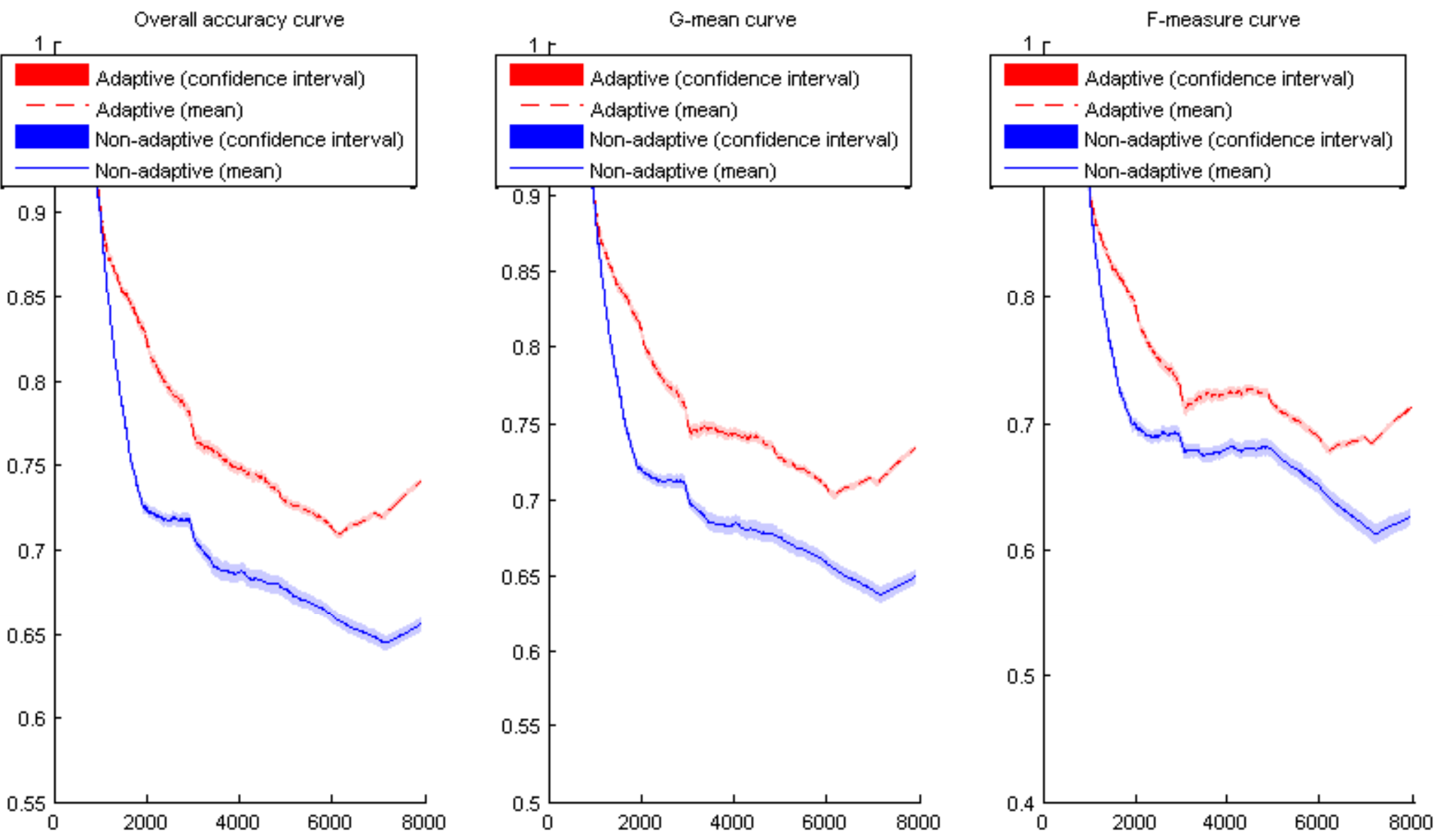}}
\subfigure[] {\includegraphics[width=.5\textwidth]{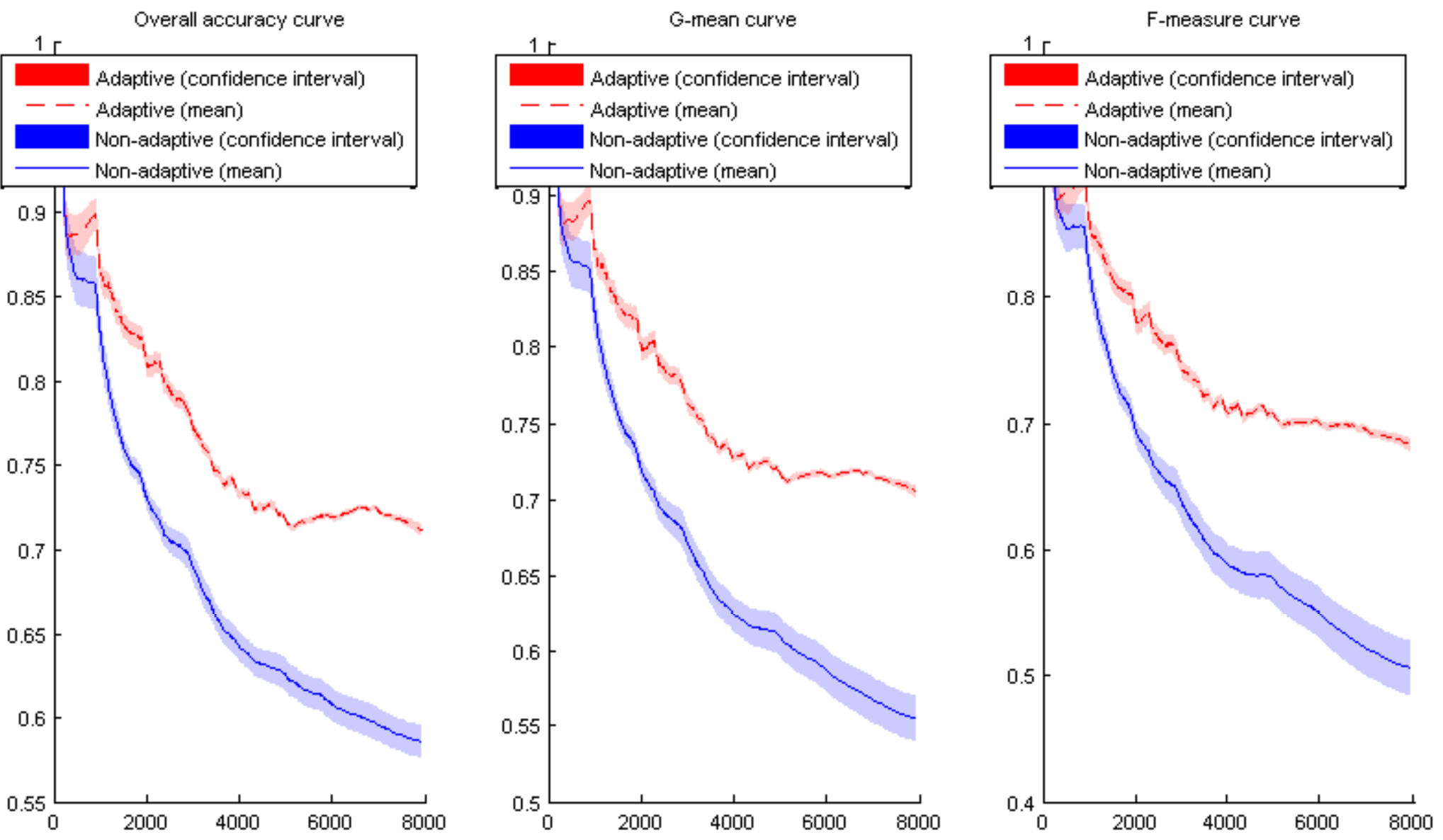}} \\
\end{tabular}
\caption{The time series representations of different metrics (OAC, F-measure, G-mean) on Checkerboard dataset for (a) A-HLFR, HLFR; (b) A-LFR, LFR; (c) A-DDM, DDM; (d) A-EDDM, EDDM; (e) A-STEPD, STEPD; and (f) A-DDM-OCI, DDM-OCI. The red dashed line denotes mean values for adaptive learning methods, the red shading envelop represents $95\%$ confidence interval. The blue solid line denotes mean values for non-adaptive learning methods, the blue shading envelop represents $95\%$ confidence interval.}
\label{fig:checkerboard_adaptive_learning}
\end{figure*}

In Fig. \ref{fig:usenet1_adaptive_learning} and Fig. \ref{fig:checkerboard_adaptive_learning}, we plotted the time series representations of OAC, F-measure and G-mean on these two datasets over $100$ Monte-carlo simulations. The shading enveloping each curve in the figures represents $95\%$ percent confidence interval. In each sub-figure, the red dashed (or blue solid) line represents mean values for drift detection algorithm with (or without) adaptive training scheme, while the red (or blue) shading envelop represents the corresponding confidence intervals. For almost all the competing algorithms their corresponding adaptive versions achieve much better classification results than the non-adaptive counterparts. This performance boost begins from the first concept drift adaptation and grows gradually with  increasing number of adaptations. As seen, A-HLFR and A-LFR achieves more compelling learning performance compared with A-DDM, A-EDDM, A-DDM-OCI and A-STEPD\footnote{The comparable performance of A-DDM on Checkerboard dataset results from more times of adaptations, which is however unreasonable as the adaptation alarms are false alarms}. This also coincides with the quantitative analysis results of concept drift detection shown in Fig. \ref{fig:precision_recall_comparison_update}. These results empirically validate the potential and superiority of using adaptive classifier techniques for concept drift adaptation, instead of the re-training strategy adopted in previous work. It is also worth noting that the adaptive classifier is not limited to soft-margin SVM. In fact, adaptive logistic regression \cite{anagnostopoulos2009temporally}, adaptive single-layer perceptron \cite{pavlidis2011lambda} and adaptive decision tree \cite{alippi2007just} frameworks all have been developed in recent years with the advance of statistical machine learning. We leave investigations of concept drift adaptation using other adaptive classifiers as future work.

\subsection{On the computational complexity analysis of concept drift detection} \label{section5.6}
Having demonstrated the benefits and effectiveness of the HHT framework, this section discusses the computational complexity of the aforementioned concept drift detectors, particularly the additional computation cost incurred by incorporating the Layer-II test. In fact, DDM, EDDM, DDM-OCI, STEPD and LFR have a constant time complexity ($\mathcal{O}(1)$) at each time point, as all of them follow a single-layer-based hypothesis testing framework that monitors one or four error-related statistics \cite{wang2015concept}. The computational complexity for generating bound tables used by LFR or HLFR to determine the corresponding warning and detection bounds with respect to different rate values $P_\star$ is $\mathcal{O}(M)$, where $M$ is the number of Monte-Carlo simulations used. However, since the bound tables can be computed offline, the time complexity for looking up the bound table values once $\hat{P}_\star$ is given (see line $18$ and $19$ of Algorithm \ref{FourRatesAlg}) remains $\mathcal{O}(1)$.
Due to the introduction of Layer-II test, HLFR is more computational expensive than other single-layer-based methods. This is because HLFR requires training $P$ classifiers ($1000$ in this work) for validating the occurrence of a potential concept drift time point\footnote{HLFR has the same computational complexity with LFR if the Layer-I test does not reject the null hypothesis at the tested time point.}. Suppose the computational complexity of training a new classifier is $\mathcal{O}(K)$, the total computational complexity of HLFR at a suspected time point is $\mathcal{O}(KP)\gg\mathcal{O}(1)$.

Despite this limitation, the HHT framework introduces a new perspective to the field of concept drift detection, especially considering its overwhelming advantages on detection precision and delay of detection. Finally, it should be noted that the $P$ permutations in Layer-II test can be run in parallel, as the classifier trained are independent across different permutations.

\section{Conclusions} \label{conclusions}

This paper proposed a novel concept drift detector, namely Hierarchical Linear Four Rates (HLFR), under the hierarchical hypothesis testing (HHT) framework.
Unlike previous work, HLFR is able to detect all possible variants of concept drifts regardless of data characteristics, it is also independent of the underlying classifier. Using Adaptive SVM as its base classifier, HLFR can be easily extended to a concept drift-agnostic framework, i.e., A-HLFR. The performance of HLFR and A-HLFR in detecting and adapting to concept drifts are compared to state-of-the-art methods using both simulated and real-world datasets that span the gamut of concept drift types (recurrent or irregular, gradual or abrupt, etc.) and data distributions (balanced or imbalanced labels). Experimental results corroborate our theoretically analysis on Type-I and Type-II errors of HLFR and also demonstrate that our methods can significantly outperform our competitors in terms of earliest detection of concept drift, highest detection precision as well as powerful adaptability across different concepts. Two real examples on email filtering and weather prediction are finally presented to illustrate effectiveness and great potential of our methods.

In the future, we will extend HLFR and A-HLFR to multi-class classification scenario. One possible solution is to use the one-vs-all strategy to convert the $N$-class classification problem into $N$ binary-class classification problems. Since the four rates associated with each binary-class classification are still geometrically weighted sum of Bernoulli random variables, HLFR and A-HLFR might be able to be applied straightforwardly. Additionally, we are also interested in investigating the performance of more sensitive metrics, from an information theoretic learning (ITL) perspective~\cite{principe2010information}, to monitor the streaming environment. Finally, we will continue on designing more power tests under HHT framework for industrial-level noisy data.

% http://www.ctan.org/tex-archive/macros/latex/contrib/supported/IEEEtran/bibtex
\bibliographystyle{IEEEabrv}
\bibliography{T_system_concept}

%\end{document}

\end{document}